\pdfoutput=1

\documentclass[10pt,twocolumn,letterpaper]{article}

\usepackage[pagenumbers]{cvpr}

\renewcommand{\paragraph}[1]{\par\addvspace{.5em}\noindent\textbf{#1}}

\usepackage{multirow}
\usepackage{colortbl}
\definecolor{oursrow}{gray}{0.92}
\usepackage{xspace}
\usepackage{algorithm}
\usepackage{algpseudocode}

\newcommand{\method}{DecisionAD\xspace}
\newcommand{\Pyes}{P(\text{yes})}
\newcommand{\nat}{\mathcal{N}}
\newcommand{\win}{\mathcal{W}}
\newcommand{\zA}{z_A}
\newcommand{\zR}{z_R}
\newcommand{\zs}{z_s}
\newcommand{\zL}{z_L}
\newcommand{\Zt}{\tilde{Z}}
\newcommand{\ta}{t_{\mathrm{a}}}
\newcommand{\tn}{t_{\mathrm{n}}}
\DeclareMathOperator{\NN}{NN}

\newcommand{\fitbox}[1]{\resizebox{\ifdim\width>\linewidth\linewidth\else\width\fi}{!}{#1}}
\usepackage{placeins}

\definecolor{cvprblue}{rgb}{0.21,0.49,0.74}
\usepackage[pagebackref,breaklinks,colorlinks,allcolors=cvprblue]{hyperref}

\def\confName{CVPR}
\def\confYear{2027}
\newcommand{\SuppSec}{Supp.~Sec.}\newcommand{\SuppSecs}{Supp.~Secs.}\newcommand{\SuppSection}{Supp.~Sec.}\newcommand{\SuppSections}{Supp.~Secs.}
\AtBeginDocument{%
  \crefname{appendix}{\SuppSec}{\SuppSecs}\Crefname{appendix}{\SuppSection}{\SuppSections}%
  \crefname{subappendix}{\SuppSec}{\SuppSecs}\Crefname{subappendix}{\SuppSection}{\SuppSections}%
  \crefname{subsubappendix}{\SuppSec}{\SuppSecs}\Crefname{subsubappendix}{\SuppSection}{\SuppSections}}

\title{\method: Training-Free Anomaly Detection\\with Consistent Answers and Visual Evidence}

\author{Woojun Lee \qquad Jungi Lee \qquad Pilsung Kang\thanks{Corresponding author.}\\
Department of Industrial Engineering, Seoul National University\\
{\tt\small \{woojun\_lee, jungi\_lee, pilsung\_kang\}@snu.ac.kr}
}

\begin{document}
\setcounter{footnote}{1}%
\maketitle
\begin{abstract}
Decision models answer typed questions with a probability for each answer instead of free-form text.
Several of them now accept images, which allows an image to be inspected by asking whether it shows a defect, without defect data or task-specific training.
On industrial images, the best of seven commercial decision models achieves a higher image-level AUROC than zero-shot anomaly detection (ZSAD) methods trained on auxiliary defect data.
However, three challenges remain for reliable deployment.
(i) Their answers are inconsistent, as the probability of a defect differs by up to 0.38 on average between asking whether an image is anomalous and asking whether it is normal.
(ii) An answer shows neither what it is based on nor where the defect is.
(iii) Every answer requires a large vision-language model behind an API, at fewer than two images per second under sequential requests.
We present DecisionAD, a training-free anomaly detector with the decision-model interface.
It answers consistently, shows its evidence, and runs on site.
(i) It reads only the polarity of the question and therefore yields the same probability of a defect for both questions whenever the polarity is parsed.
(ii) It computes each answer with an evidence map, which marks the patches that support the answer and those rare within the image.
(iii) It uses a frozen vision encoder and its paired text encoder, at 25 images per second on one GPU.
On five industrial datasets, DecisionAD achieves the highest mean image-level AUROC (88.2) among 12 reproduced ZSAD methods and the seven decision models.
It processes images about 15$\times$ faster than these models under sequential requests.
\end{abstract}

\section{Introduction}
\label{sec:intro}

\begin{figure}[t]
\centering
\includegraphics[width=\linewidth]{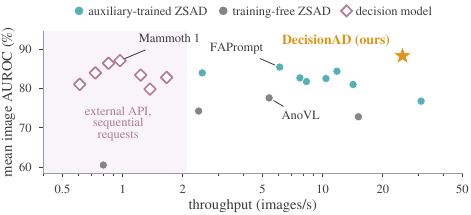}
\caption{\textbf{Image AUROC versus throughput.}
Mean image AUROC over the five industrial datasets. Decision models use sequential requests. Hollow markers have no evidence map.}
\label{fig:decision}
\end{figure}

Visual inspection needs an answer and the evidence behind it.
An inspector asks ``Is there a visible defect?''
When a part is rejected, the inspector checks where the defect is.
Existing methods provide only one of the two.

\paragraph{Zero-shot anomaly detection.}
Zero-shot anomaly detection (ZSAD) methods~\cite{jeong2023winclip,zhou2024anomalyclip} provide the evidence.
They detect defects without training images of the target and return an anomaly map and an image score.
However, their text is a fixed prompt and not a question to answer.
Most of them also require prior knowledge about the defects or the product.
Auxiliary-trained methods learn prompts or adapters from an auxiliary defect dataset with pixel masks~\cite{zhou2024anomalyclip,qu2025bayespfl,ma2025aaclip,hou2026visualad}.
Training-free methods skip this training, but most of them need the product category in the prompt, as in ``a photo of a damaged bottle''~\cite{jeong2023winclip,deng2023anovl,li2024fade}.

\paragraph{Decision models.}
\emph{Decision models}~\cite{openrouter2026decisions}, recently offered as commercial APIs, provide the answer.
A user declares the possible answers to each question in advance, such as yes/no, one of several options, or a graded score.
The model then returns a probability for each answer instead of free-form text.
Although these models were first built for text and structured data~\cite{ibrahim2026decision}, several now accept images~\cite{openrouter2026decisions,typesafe2026jev}.
Such a model takes the inspector's question and requires neither defect data nor task-specific training.

\begin{figure}[t]
\centering
\includegraphics[width=0.8\linewidth]{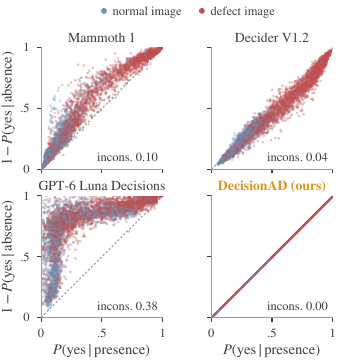}
\caption{\textbf{Answers to a question and its negation.}
One dot per image of four industrial datasets.
Consistent answers lie on the diagonal, and the number is the mean absolute difference of the two answers, the polarity inconsistency (\cref{sec:pilot}).}
\label{fig:polarity}
\end{figure}

\paragraph{Promise and challenges.}
We therefore ask whether decision models can answer the inspection question reliably.
To our knowledge, we present the first evaluation of commercial decision models for anomaly detection (\cref{sec:pilot}).
We send the images of five industrial datasets to seven models, once with a question that asks whether a defect is present and once with its negation.
The models rank images accurately.
The best one achieves a mean industrial image AUROC of 87.1, while the eight auxiliary-trained ZSAD methods reach 76.8--85.4 (\cref{fig:decision}).
However, three challenges remain for reliable deployment.
\begin{enumerate*}[label=(\roman*)]
\item \emph{Inconsistency under negation.} The probability of a defect differs by up to 0.38 on average between the question and its negation, which ranking metrics do not reveal (\cref{fig:polarity}).
\item \emph{Missing evidence.} An answer shows neither what it is based on nor where the defect is.
\item \emph{Low throughput.} Sequential requests yield fewer than two images per second, and every image must leave the site.
\end{enumerate*}
They rank images accurately, but their answers are inconsistent and lack evidence, and their throughput is low.

\begin{figure}[t]
\centering
\includegraphics[width=\linewidth]{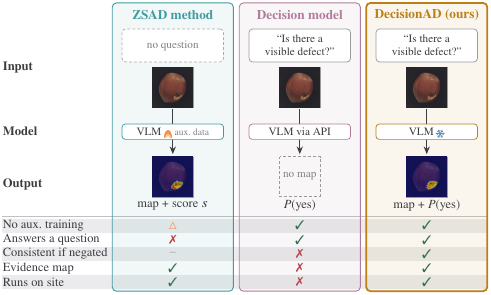}
\caption{\textbf{ZSAD methods, decision models, and \method.}
All three use a vision-language model (VLM). ZSAD methods return a map but take no question, and decision models answer a question but return no map. \method returns both. The flame marks training on auxiliary defect data, $\triangle$ marks that only some methods train, and the snowflake marks a frozen model.}
\label{fig:concept}
\end{figure}

\paragraph{Our approach.}
We present \method, which provides consistent answers with evidence at high throughput (\cref{fig:concept}).
It keeps the decision-model interface, taking a question as input and returning $\Pyes$, the probability of the answer yes.
Among the answer types of this interface, it supports yes/no questions about a defect.
The answer is derived from evidence in a frozen vision encoder and its paired text encoder, without training, auxiliary defect data, or category names.
\begin{enumerate*}[label=(\roman*)]
\item \emph{Consistency under negation.} \method takes from the question only its \emph{polarity}, \ie, whether it asks for the presence or the absence of a defect. All evidence is computed from the image. Because the polarity sets the sign of the answer, a question and its negation yield the same probability of a defect whenever the polarity is parsed.
\item \emph{Evidence.} Each answer comes with an evidence map. The map combines two terms on a common scale given by unlabeled natural images. \emph{Answer attribution} is the gradient of an image-level \emph{answer score}, and \emph{self-rarity} marks patches without a similar patch in the same image. The pooled map and the answer score together determine the answer.
\item \emph{Throughput.} \method runs on site at 25 images per second on one GPU, about 15$\times$ faster than sequential requests. It has 1.2B parameters, against 27B for the decision models of known size.
\end{enumerate*}
On five industrial datasets, its mean image AUROC of 88.2 exceeds the best decision model (87.1) and ZSAD method (85.4).

\paragraph{Contributions.}
\begin{itemize}
\item We evaluate seven decision models for anomaly detection and identify three challenges, namely inconsistency under negation, missing evidence, and low throughput.
\item We propose \method, a training-free method that answers a yes/no question about a defect with $\Pyes$ and an evidence map. It is consistent under negation whenever the polarity is parsed.
\item On five industrial datasets, \method achieves the highest mean image AUROC among 12 reproduced ZSAD methods and seven decision models. It is about 15$\times$ faster than these models under sequential requests.
\item With a few normal images, \method outperforms five few-shot baselines in image AUROC across all datasets and reference counts while remaining training-free.
\end{itemize}

\section{Related work}
\label{sec:related}

\paragraph{Zero-shot anomaly detection.}
APRIL-GAN~\cite{chen2023aprilgan} and AnomalyCLIP~\cite{zhou2024anomalyclip} established cross-dataset evaluation, training on one defect dataset and testing on another.
Follow-up work learns visual context prompts~\cite{qu2024vcpclip}, prompt distributions~\cite{qu2025bayespfl}, an anomaly-aware text space~\cite{ma2025aaclip}, fine-grained abnormality prompts~\cite{zhu2025faprompt}, patch-wise experts~\cite{park2026moeclip}, conditional prompt synthesis~\cite{chen2026cops}, or a foreground--background decomposition~\cite{hu2026fbclip}.
VisualAD~\cite{hou2026visualad} works without a text encoder.
Among training-free methods, WinCLIP~\cite{jeong2023winclip} builds anomaly maps from window-wise image--text similarity in CLIP~\cite{radford2021clip}.
AnoVL~\cite{deng2023anovl} and FADE~\cite{li2024fade} also build their maps from CLIP features, whereas SAA+~\cite{cao2023saa} composes GroundingDINO~\cite{liu2024groundingdino} and SAM~\cite{kirillov2023sam} through defect prompts. The three CLIP-based methods include the product category in the prompt.
\method needs neither auxiliary defect data nor category names, and it answers a question with a probability.

\paragraph{Comparison within and across images.}
Few-shot methods~\cite{roth2022patchcore,damm2025anomalydino,zhai2025foundad,li2024promptad,gao2025adaptclip,tao2025kagprompt} measure patch distances to normal reference images.
Without references, patches can be compared within a single texture image~\cite{aota2023zeroshot}.
MuSc~\cite{li2024musc} instead scores the unlabeled test images against each other and thus needs many test images at once.
Self-rarity compares patches inside one image and processes each image independently.

\paragraph{Decision models.}
Decision models~\cite{openrouter2026decisions} return a probability for each answer instead of free-form text.
Both the ranking and the consistency of these probabilities can therefore be measured.
Language models are sensitive to meaning-preserving changes of the prompt~\cite{sclar2024formatting}, and vision-language models often fail to handle negation~\cite{alhamoud2025negation}.
We measure how negating the question changes their answers.

\section{Decision models for anomaly detection: promise and challenges}
\label{sec:pilot}

\begin{figure*}[t]
  \centering
  \includegraphics[width=\textwidth]{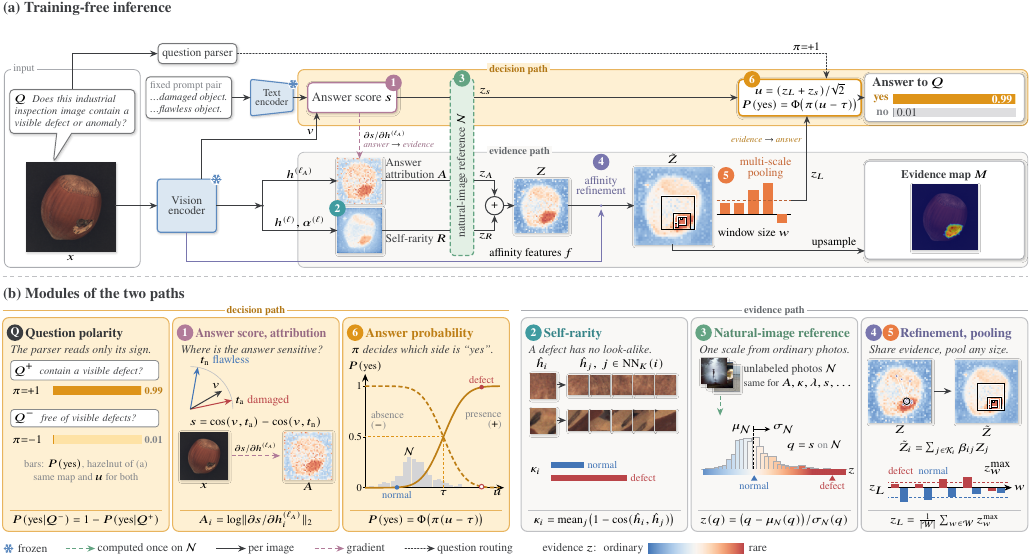}
  \caption{\textbf{Overview of \method.}
  (a)~One forward and one backward pass of a frozen encoder give $\Pyes$ for the question $Q$ and an evidence map. The map of the evidence path at the bottom is pooled into the answer of the decision path at the top.
  (b)~The modules of each path, numbered as in (a).}
  \label{fig:overview}
\end{figure*}

\paragraph{Setup.}
We evaluate seven commercial decision models served through the OpenRouter Decisions API~\cite{openrouter2026decisions}.
Each request sends one image and one question and receives $\Pyes$, the probability of the answer yes.
No labels, file names, or categories are sent.
The \emph{presence question} $Q^{+}$ is ``Does this industrial inspection image contain a visible defect or anomaly?''
It is asked on the five industrial datasets of \cref{sec:exp_setup}.
For the polarity study, the same images are sent again with the \emph{absence question} $Q^{-}$, ``Is this industrial inspection image free of any visible defect or anomaly?'' (all questions in \cref{sec:supp_decision}).
A consistent model yields the same probability of a defect for both questions, $P(\text{yes}\mid Q^{+})=1-P(\text{yes}\mid Q^{-})$.
We call their absolute difference the \emph{polarity inconsistency} of that image.

\paragraph{Promise.}
Decision models rank images accurately, and Mammoth~1 and Decider~V1.2 exceed all eight auxiliary-trained ZSAD methods in mean industrial image AUROC (\cref{tab:industrial,fig:decision}).
\paragraph{(i) Inconsistency under negation.}
The mean polarity inconsistency of the seven models is 0.04--0.38 (\cref{fig:polarity,tab:decision,tab:supp_gap}).
Ranking metrics do not reveal it, because the two questions rank the images similarly.
Their median Spearman correlation is 0.94, and their AUROC differs by at most 3\%p for 31 of 35 model--dataset pairs.
Instead, the probabilities shift.
In 26 of 35 pairs, the absence question yields a higher probability of a defect.
The answer for an image therefore depends on the wording of the question.
Other wordings change both the answer and the ranking.
Over six more question pairs, the mean inconsistency of a model reaches 0.49, and its image AUROC differs by up to 15.6\%p between the two questions of a pair (\cref{tab:supp_wording_pairs}).

\paragraph{(ii) Missing evidence.}
A decision model returns only $\Pyes$, without the location of the defect or the basis of the answer.
An inspector therefore cannot verify the answer.
Only Jev 1.14~\cite{typesafe2026jev} adds a bounding box, one per image with a confidence value, whether or not the target is present.
The answer is not based on this box, as the box is returned for any value of $\Pyes$.
The box also covers part of every normal image, and its AUPRO is 32--48 (\cref{sec:supp_box}).

\paragraph{(iii) Low throughput.}
Sequential requests yield 0.6--1.7 images per second, computed from the median request latency.
The fastest model is slower than 11 of the 12 ZSAD methods on one GPU (\cref{fig:decision}). Every request runs a large vision-language model behind an API, with 27B parameters for the models of known size (\cref{sec:supp_efficiency}). Concurrent requests raise the throughput, but every image still passes through such a model and must leave the site.

\section{Method}
\label{sec:method}

\method has the input and output of a decision model but builds its answer from evidence in a frozen encoder.
It follows five principles.
First, it takes from the question only its polarity, which sets the sign of the answer.
Second, a fixed prompt pair yields the answer score of the whole image.
Third, the evidence combines two maps. Answer attribution shows where this score is sensitive, and self-rarity shows what is uncommon within the image.
Fourth, the answer score and both kinds of evidence share one scale defined by unlabeled natural images, which removes the need for a fusion weight per dataset.
Fifth, the pooled evidence map and the answer score together determine the answer. Answer and evidence thus come from the same computation.
\Cref{fig:overview} shows the pipeline, and \cref{sec:supp_impl} the settings.

\subsection{Problem setting}
\label{sec:setting}
The input is an image $x$ and a yes/no question $Q$ that asks whether a defect is present or absent.
The output is an answer probability $\Pyes=P(\text{yes}\mid Q,x)\in[0,1]$ and an evidence map $M\in\mathbb{R}^{H\times W}$.
For a presence question $Q^{+}$, $P(\text{yes}\mid Q^{+},x)=P(\text{defect}\mid x)$, and for an absence question $Q^{-}$, $P(\text{yes}\mid Q^{-},x)=1-P(\text{defect}\mid x)$.
\method satisfies three conditions.
\begin{enumerate*}[label=(C\arabic*)]
\item No weights are trained.
\item No normal/anomalous labels or defect masks are used, either at inference or for computing the reference statistics.
\item Images are processed independently, without statistics of the evaluation set.
\end{enumerate*}
The only external data is a set $\nat$ of 480 unlabeled natural images from COCO~\cite{lin2014coco}.

\subsection{Question polarity}
\label{sec:parser}
A rule-based parser reads the polarity $\pi=b\,(-1)^{n}$ without an LLM.
The base sign $b$ is $+1$ for a defect cue such as ``defect'' and $-1$ for a quality cue such as ``intact'', and $n$ is the number of negators such as ``not'' or ``without''.
A question without a cue is declined, and its polarity is then declared with the question.
The sign $\pi$ enters only the final answer (\cref{eq:answer}).
Everything before it is computed from the image and a fixed prompt pair (\cref{sec:answer}).
The question itself is not fed to the text encoder (\cref{sec:encoder}).
This encoder does not handle negation~\cite{alhamoud2025negation}, and an absence question would therefore invert its ranking (\cref{tab:question}).
\Cref{sec:supp_parser} evaluates the parser on 104 questions of six types.

\subsection{Frozen encoder and a single pass}
\label{sec:encoder}
The vision encoder is C-RADIOv4-SO400M~\cite{nvidia2026cradiov4} with two of its heads.
RADIO~\cite{ranzinger2024amradio,heinrich2025radiov25} distills several teachers into one backbone and provides heads that map its features into each teacher's output space.
The SigLIP2 head~\cite{tschannen2025siglip2} yields a summary vector $v$, and the SAM3 head~\cite{carion2025sam3} yields patch features $f$ for segmentation.
The SigLIP2 text encoder receives only the fixed prompt pair, and its embeddings $\ta$ and $\tn$ are compared with the summary vector $v$ (\cref{sec:answer}).
The image is resized to a 592\,px square, so that its $37{\times}37$ patch grid matches that of the CLIP-based baselines at 518\,px.
A single pass yields the intermediate tokens $h^{(\ell)}$ and attention weights $\alpha^{(\ell)}$ of each layer, together with $f$ and $v$.
All maps are computed on the patch grid.
Self-rarity is computed on the layers $\mathcal{L}$ in the middle third of the depth, and answer attribution on one layer $\ell_A$ among them.

\subsection{Answer score and answer attribution}
\label{sec:answer}
\paragraph{Answer score $s$.}
Each sentence of the prompt pair combines a template from a set $\mathcal{T}$ with an object-agnostic state phrase.
The state phrases come in two sets, $\mathcal{S}_{\mathrm{n}}$ for the normal side, such as ``flawless object'', and $\mathcal{S}_{\mathrm{a}}$ for the anomalous side, such as ``damaged object''.
The text embeddings of all sentences on each side are averaged,
\begin{equation}
t_c \propto \sum_{\psi\in\mathcal{T}}\sum_{\omega\in\mathcal{S}_c}\hat\phi\big(\psi(\omega)\big),\qquad c\in\{\mathrm{n},\mathrm{a}\},
\label{eq:prompt}
\end{equation}
where $\hat\phi$ is the $\ell_2$-normalized SigLIP2 text embedding and $\propto$ denotes re-normalization of the sum to unit length.
The prompts do not contain a product name, and the model therefore receives no category label.
The answer score measures which side the summary vector $v$ is closer to,
\begin{equation}
s(x)=\cos(v,\ta)-\cos(v,\tn).
\label{eq:s}
\end{equation}
Set sizes and example sentences are given in \cref{tab:impl}.

\paragraph{Answer attribution $A$.}
We differentiate $s$ with respect to the output tokens $h^{(\ell_A)}_i$ of layer $\ell_A$ and take the log of the gradient norm,
\begin{equation}
A_i=\log\big\|\partial s/\partial h^{(\ell_A)}_i\big\|_2 .
\label{eq:A}
\end{equation}
Because the computation graph is cut at the output of layer $\ell_A$, the backward pass only traverses the layers above it and the SigLIP2 head.
A large $A_i$ means that the answer score is sensitive to patch $i$. We use this sensitivity as a component of the map, not as a causal explanation.
As in gradient-based saliency~\cite{simonyan2014saliency} and unlike Grad-CAM~\cite{selvaraju2017gradcam}, we use only the magnitude of the token gradient, without channel weights.
We take its log because the magnitudes span several orders across patches (\cref{sec:supp_refstats}).
As in anomaly detection with mid-level features~\cite{roth2022patchcore,guo2025dinomaly}, we use a middle layer, as deeper tokens blur locations (\cref{sec:exp_ablation}).

\subsection{Self-rarity}
\label{sec:rarity}
The surface and texture of a normal product repeat within an image, whereas a defect rarely has a similar region in the same image~\cite{aota2023zeroshot}.
Self-rarity measures this with two statistics per patch $i$ and layer $\ell\in\mathcal{L}$. The first is feature rarity,
\begin{equation}
\kappa^{(\ell)}_i=\frac{1}{K}\sum_{j\in\NN_K(i)}\big(1-\cos(\hat h^{(\ell)}_i,\hat h^{(\ell)}_j)\big),
\label{eq:knn}
\end{equation}
where $\hat h$ is the normalized token and $\NN_K(i)$ is the set of the $K$ patches in the same image that are most similar to $\hat h^{(\ell)}_i$.
The patch itself and its immediate neighbors are excluded from the candidates, since neighboring patches resemble each other even on a defect.
The second is attention locality,
\begin{equation}
\lambda^{(\ell)}_i=-\frac{\sum_j \alpha^{(\ell)}_{ij}\,\|c_i-c_j\|}{\sum_j \alpha^{(\ell)}_{ij}},
\label{eq:att}
\end{equation}
where $\alpha^{(\ell)}_{ij}$ is the attention of one head, the sum runs over patch tokens $j$ only, and $c_i$ is the grid coordinate of patch $i$. $\lambda^{(\ell)}_i$ is averaged over the heads.
It is large when the attention of a patch stays nearby.
As a defect patch has no similar patch elsewhere, its attention tends to stay on itself and its neighbors.
Each map, for every layer and both statistics, is standardized on $\nat$ (\cref{sec:reference}).
Their average $R$ is standardized again to give $\zR=z(R)$.
Computed without text, self-rarity complements answer attribution.

\subsection{Natural-image reference}
\label{sec:reference}
Adding different kinds of evidence requires a common scale~\cite{bergmann2020uninformed}. \method therefore standardizes every quantity $q$ on $\nat$ as
\begin{equation}
z(q)=(q-\mu_q)/\sigma_q.
\label{eq:z}
\end{equation}
This applies to the rarity statistics, $A$, $s$, and the window statistics of \cref{sec:pooling}. For a map, the mean $\mu_q$ and the standard deviation $\sigma_q$ are taken over all patches of $\nat$, and for an image-level quantity, over its images.
Both are computed once and used unchanged for every dataset, which satisfies (C3).
The set $\nat$ carries no labels, and its images need neither be ``normal'' nor come from the target domain. They fix the units for comparing different kinds of evidence.
A unit is the typical strength of that evidence in natural images.

\begin{table*}[t]
\centering
\caption{\textbf{Comparison on industrial datasets} (category macro, \%).
Train is the data used to train for anomaly detection, where -- means none and AD means auxiliary defect datasets.
In all tables, AUC denotes AUROC and PRO denotes AUPRO.
The decision models return no map.
FPS is measured on one GPU for the local methods and under sequential requests for the decision models.
Best in bold, second underlined.}
\label{tab:industrial}
\fitbox{\scriptsize
\setlength{\tabcolsep}{1.5pt}
\begin{tabular}{l c c cccc cccc cccc cccc cccc cccc r}
\toprule
\multicolumn{1}{l}{\multirow{3}{*}[-1\dimexpr\aboverulesep+\belowrulesep\relax]{Method}} & \multirow{3}{*}[-1\dimexpr\aboverulesep+\belowrulesep\relax]{Venue} & \multirow{3}{*}[-1\dimexpr\aboverulesep+\belowrulesep\relax]{Train} & \multicolumn{4}{c}{MVTec AD} & \multicolumn{4}{c}{VisA} & \multicolumn{4}{c}{MPDD} & \multicolumn{4}{c}{BTAD} & \multicolumn{4}{c}{Real-IAD} & \multicolumn{4}{c}{Mean} & \multirow{3}{*}[-1\dimexpr\aboverulesep+\belowrulesep\relax]{FPS} \\
\cmidrule(lr){4-7} \cmidrule(lr){8-11} \cmidrule(lr){12-15} \cmidrule(lr){16-19} \cmidrule(lr){20-23} \cmidrule(lr){24-27}
 &  &  & \multicolumn{2}{c}{Image} & \multicolumn{2}{c}{Pixel} & \multicolumn{2}{c}{Image} & \multicolumn{2}{c}{Pixel} & \multicolumn{2}{c}{Image} & \multicolumn{2}{c}{Pixel} & \multicolumn{2}{c}{Image} & \multicolumn{2}{c}{Pixel} & \multicolumn{2}{c}{Image} & \multicolumn{2}{c}{Pixel} & \multicolumn{2}{c}{Image} & \multicolumn{2}{c}{Pixel} &  \\
\cmidrule(lr){4-5} \cmidrule(lr){6-7} \cmidrule(lr){8-9} \cmidrule(lr){10-11} \cmidrule(lr){12-13} \cmidrule(lr){14-15} \cmidrule(lr){16-17} \cmidrule(lr){18-19} \cmidrule(lr){20-21} \cmidrule(lr){22-23} \cmidrule(lr){24-25} \cmidrule(lr){26-27}
 &  &  & AUC & AP & AUC & PRO & AUC & AP & AUC & PRO & AUC & AP & AUC & PRO & AUC & AP & AUC & PRO & AUC & AP & AUC & PRO & AUC & AP & AUC & PRO &  \\
\midrule
WinCLIP~\cite{jeong2023winclip} & CVPR'23 & -- & 90.4 & 95.6 & 82.3 & 62.2 & 75.5 & 78.7 & 73.0 & 51.3 & 61.5 & 69.2 & 71.5 & 41.9 & 69.6 & 70.9 & 72.6 & 26.6 & 67.1 & 63.0 & 84.2 & 55.6 & 72.8 & 75.5 & 76.7 & 47.5 & 15.1 \\
AnoVL~\cite{deng2023anovl} & arXiv'23 & -- & 92.5 & 96.8 & 86.5 & 70.9 & 79.2 & 81.8 & 85.1 & 60.2 & 68.6 & 71.8 & 60.0 & 33.0 & 80.7 & 72.1 & 75.2 & 41.8 & 67.0 & 63.2 & 80.8 & 48.6 & 77.6 & 77.1 & 77.5 & 50.9 & 5.4 \\
FADE~\cite{li2024fade} & BMVC'24 & -- & 90.1 & 95.5 & 89.2 & 84.2 & 74.7 & 78.5 & 91.0 & 77.6 & 62.8 & 70.3 & 95.2 & 85.7 & 76.1 & 72.5 & 96.4 & 76.7 & 67.7 & 63.8 & 93.0 & 82.5 & 74.3 & 76.1 & 93.0 & 81.3 & 2.4 \\
SAA+~\cite{cao2023saa} & TCYB'25 & -- & 62.4 & 81.1 & 73.9 & 55.0 & 71.0 & 77.2 & 76.4 & 58.3 & 56.3 & 67.5 & 81.9 & 66.5 & 57.0 & 76.4 & 59.3 & 35.9 & 56.1 & 52.9 & 75.6 & 48.4 & 60.6 & 71.1 & 73.4 & 52.8 & 0.8 \\
\midrule
APRIL-GAN~\cite{chen2023aprilgan} & arXiv'23 & AD & 86.0 & 93.4 & 86.7 & 84.6 & 77.5 & 80.8 & 94.2 & 86.9 & 76.5 & 82.6 & 94.3 & 84.7 & 75.5 & 70.7 & 89.3 & 69.0 & 68.4 & 64.5 & 95.2 & 86.6 & 76.8 & 78.4 & 91.9 & 82.4 & \textbf{31.1} \\
AnomalyCLIP~\cite{zhou2024anomalyclip} & ICLR'24 & AD & 91.6 & 96.3 & 91.1 & 85.3 & 82.0 & 85.3 & 95.5 & 86.6 & 77.5 & 82.5 & 96.5 & 89.0 & 88.8 & 90.2 & 94.3 & 74.4 & 68.9 & 63.5 & 95.0 & 85.0 & 81.8 & 83.6 & 94.5 & 84.1 & 8.3 \\
VCP-CLIP~\cite{qu2024vcpclip} & ECCV'24 & AD & 89.8 & 95.5 & \textbf{92.1} & 87.7 & 83.8 & 87.6 & \underline{95.8} & 90.0 & 69.3 & 72.9 & 96.8 & 90.0 & 89.7 & 94.0 & 93.8 & 75.0 & 72.4 & 69.9 & 96.4 & 87.7 & 81.0 & 84.0 & \underline{95.0} & 86.1 & 14.2 \\
Bayes-PFL~\cite{qu2025bayespfl} & CVPR'25 & AD & 92.2 & 96.2 & \underline{91.9} & \underline{88.1} & 86.9 & \underline{89.3} & \textbf{95.9} & 88.8 & 76.9 & 78.7 & \textbf{97.1} & \underline{90.8} & 91.2 & 90.6 & 92.2 & 77.8 & 72.6 & 68.9 & 96.0 & 85.8 & 84.0 & 84.7 & 94.6 & 86.2 & 2.5 \\
AA-CLIP~\cite{ma2025aaclip} & CVPR'25 & AD & 89.8 & 95.1 & 91.3 & 87.6 & 78.1 & 81.8 & 94.7 & 81.8 & 74.3 & 79.2 & 96.3 & 87.0 & \underline{95.4} & \underline{98.0} & \textbf{97.1} & \underline{79.7} & 75.0 & 73.2 & 95.0 & 81.2 & 82.5 & 85.4 & 94.9 & 83.5 & 10.4 \\
FAPrompt~\cite{zhu2025faprompt} & ICCV'25 & AD & 91.8 & 95.6 & 89.4 & 85.5 & 85.0 & 86.7 & \underline{95.8} & 87.9 & 79.8 & \underline{83.0} & 96.3 & 88.3 & 93.0 & 96.2 & 96.4 & 78.5 & 77.4 & 74.8 & 94.7 & 84.8 & 85.4 & 87.3 & 94.5 & 85.0 & 6.1 \\
MoECLIP~\cite{park2026moeclip} & CVPR'26 & AD & \underline{93.0} & 97.0 & 91.5 & 86.2 & 82.8 & 84.8 & 94.8 & 83.9 & 68.3 & 73.9 & 96.8 & 87.8 & 90.3 & 95.2 & 96.4 & 76.7 & 79.0 & 76.8 & 94.5 & 79.7 & 82.7 & 85.6 & 94.8 & 82.8 & 7.7 \\
VisualAD~\cite{hou2026visualad} & CVPR'26 & AD & 92.2 & 96.7 & 90.8 & 87.9 & 84.6 & 87.6 & 95.7 & \underline{90.2} & 71.5 & 79.5 & \textbf{97.1} & \textbf{92.5} & 95.0 & \underline{98.0} & 91.1 & 79.2 & 78.5 & 76.2 & \underline{96.6} & \underline{91.8} & 84.4 & 87.6 & 94.3 & \underline{88.3} & 11.8 \\
\midrule
Jev 1.14 & API & -- & 92.5 & 96.2 & -- & -- & 85.3 & 87.5 & -- & -- & 62.8 & 71.9 & -- & -- & 92.3 & 88.8 & -- & -- & \underline{81.1} & 79.0 & -- & -- & 82.8 & 84.7 & -- & -- & 1.7 \\
Decider V1.2 & API & -- & 92.1 & 96.9 & -- & -- & \underline{87.5} & \textbf{90.8} & -- & -- & 79.9 & \underline{83.0} & -- & -- & 91.5 & 91.7 & -- & -- & 80.9 & \underline{79.7} & -- & -- & 86.4 & 88.4 & -- & -- & 0.8 \\
GPT-6 Luna Decisions & API & -- & 87.2 & 94.6 & -- & -- & 80.2 & 83.2 & -- & -- & 76.2 & 77.9 & -- & -- & 79.0 & 71.5 & -- & -- & 76.6 & 74.4 & -- & -- & 79.8 & 80.3 & -- & -- & 1.4 \\
Clef & API & -- & 91.5 & 96.4 & -- & -- & 78.9 & 84.1 & -- & -- & 77.3 & 82.3 & -- & -- & 81.0 & 83.6 & -- & -- & 76.7 & 75.4 & -- & -- & 81.1 & 84.4 & -- & -- & 0.6 \\
Mammoth 1 & API & -- & \underline{93.0} & \underline{97.2} & -- & -- & \textbf{88.2} & \textbf{90.8} & -- & -- & \underline{80.4} & 82.0 & -- & -- & 92.7 & 95.3 & -- & -- & 81.0 & \textbf{79.8} & -- & -- & \underline{87.1} & \underline{89.0} & -- & -- & 1.0 \\
Drex Max & API & -- & 90.7 & 96.3 & -- & -- & 85.6 & 88.7 & -- & -- & 73.6 & 78.8 & -- & -- & 89.9 & 90.0 & -- & -- & 79.9 & 78.3 & -- & -- & 83.9 & 86.4 & -- & -- & 0.7 \\
Span-02 & API & -- & 90.4 & 95.5 & -- & -- & 82.5 & 85.9 & -- & -- & 77.7 & 82.3 & -- & -- & 88.3 & 88.4 & -- & -- & 78.0 & 77.0 & -- & -- & 83.4 & 85.8 & -- & -- & 1.2 \\
\midrule
\textbf{DecisionAD} & -- & -- & \textbf{96.0} & \textbf{97.9} & 91.6 & \textbf{90.4} & \underline{87.5} & 89.2 & 95.1 & \textbf{90.5} & \textbf{80.5} & \textbf{83.6} & \underline{97.0} & \underline{90.8} & \textbf{96.0} & \textbf{99.1} & \underline{96.8} & \textbf{85.6} & \textbf{81.3} & 79.4 & \textbf{96.7} & \textbf{92.8} & \textbf{88.2} & \textbf{89.8} & \textbf{95.4} & \textbf{90.0} & \underline{25.1} \\
\bottomrule
\end{tabular}
}
\end{table*}

\subsection{Evidence fusion and affinity refinement}
\label{sec:refine}
The two standardized maps, $\zA=z(A)$ and $\zR$, are added with equal weights,
\begin{equation}
Z=(\zA+\zR)/\sqrt{2}.
\label{eq:Z}
\end{equation}
The factor $\sqrt{2}$ keeps unit variance for independent maps.
Because both maps share a scale, they are added without a fusion weight.
Computed per patch, $Z$ has coarse boundaries and uneven values inside a defect.
Patches with similar SAM3-head features $f$ therefore share their values, as in affinity-based refinement~\cite{wang2020seam}.
SAM features have served as such affinities for language-aligned features~\cite{shi2025trident}, and the SAM head of RADIO has refined open-vocabulary segmentation~\cite{alama2026radseg}.
The refined map is
\begin{equation}
\begin{gathered}
\Zt_i=\sum_{j\in\mathcal{K}_i}\beta_{ij}Z_j,\\
\beta_{ij}=\frac{\exp(\cos(f_i,f_j)/\gamma)}{\sum_{j'\in\mathcal{K}_i}\exp(\cos(f_i,f_{j'})/\gamma)},
\end{gathered}
\label{eq:refine}
\end{equation}
where $\mathcal{K}_i$ holds a fixed number of patches whose features are most similar to $f_i$, with $i$ itself among them, and $\gamma$ is a temperature.
As patches on the same part or defect have similar $f$, the values spread evenly within each region.
The evidence map $M$ is $\Zt$ bilinearly upsampled to the input size.

\subsection{Multi-scale window pooling and the answer}
\label{sec:pooling}
As defect sizes are unknown, we slide square windows of several sizes $w\in\win$ over the map, take the maximum window mean~\cite{zavrtanik2021draem}, and standardize it on $\nat$ for each $w$,
\begin{equation}
\begin{gathered}
z^{\max}_w=z\Big(\max_{p}\frac{1}{w^2}\sum_{i\in\Omega_w(p)}\Zt_i\Big),\\
\zL=\frac{1}{|\win|}\sum_{w\in\win}z^{\max}_w ,
\end{gathered}
\label{eq:pool}
\end{equation}
where $\Omega_w(p)$ is the $w{\times}w$ window at $p$. Small windows respond to small defects and large windows to extended defects. Because larger windows yield smaller maxima, each $w$ is standardized before averaging.
The answer statistic and probability are
\begin{equation}
u=(\zL+\zs)/\sqrt{2},\qquad \Pyes=\Phi\big(\pi\cdot(u-\tau)\big),
\label{eq:answer}
\end{equation}
where $\zs=z(s)$, $\Phi$ is the standard normal CDF, $\pi$ is the polarity of \cref{sec:parser}, and $\tau=\Phi^{-1}(0.95)$.
With this $\tau$, about 5\% of natural images receive $\Pyes$ above 0.5 for a presence question, under a normal approximation.
The choice of $\tau$ affects neither the ranking metrics nor the consistency under negation.
For a presence question, $\Pyes$ thus increases with the rarity of the evidence relative to natural images.
It is not a calibrated probability of a defect, as calibration needs labeled data~\cite{guo2017calibration}.

\paragraph{Properties.}
\label{sec:properties}For a presence question $Q^{+}$ and an absence question $Q^{-}$ about the same image, $P(\text{yes}\mid Q^{+})+P(\text{yes}\mid Q^{-})=\Phi(u-\tau)+\Phi(\tau-u)=1$.
The two answers are therefore consistent whenever the parser reads the polarity correctly.
This consistency follows from the construction of the answer and not from an understanding of the question.
Moreover, $\Pyes$ for a presence question increases with $\zs$ and with every window statistic. The strongest region of the map therefore raises the probability of the answer yes.

\section{Experiments}
\label{sec:exp}

\subsection{Experimental setup}
\label{sec:exp_setup}

\paragraph{Datasets.}
The industrial datasets are MVTec AD~\cite{bergmann2019mvtec}, VisA~\cite{zou2022visa}, MPDD~\cite{jezek2021mpdd}, BTAD~\cite{mishra2021btad}, and the large-scale Real-IAD~\cite{wang2024realiad}.
Nine medical datasets are evaluated separately in \cref{sec:supp_medical}.
We use only the official test sets, without any training images of the targets (\cref{sec:supp_impl}).

\paragraph{Metrics.}
For ranking and localization, we report four category-macro metrics, namely image AUROC, image AP, pixel AUROC, and AUPRO@0.30~\cite{bergmann2020uninformed}, written AUPRO. F1-max and pixel AP are in \cref{sec:supp_full}.
For consistency, we report the polarity inconsistency of \cref{sec:pilot}.

\paragraph{Baselines and implementation.}
We reproduce 12 methods from official code under one evaluator, namely four training-free methods (WinCLIP, AnoVL, FADE, SAA+) and eight auxiliary-trained methods (\cref{sec:supp_baselines}).
All settings of \method are shared across the five datasets.
All local methods are timed on the same GPU with the same protocol (\cref{sec:supp_timing}).

\subsection{Comparison with ZSAD methods and decision models}
\label{sec:exp_zsad}
\Cref{tab:industrial} reports the industrial results.
Among the ZSAD methods, \method achieves the highest image AUROC on all five datasets.
It exceeds the best auxiliary-trained method by 0.6--3.0\%p and the best training-free method by 3.5--15.3\%p, with the largest margin on BTAD.
It also outperforms the four training-free methods in every metric of \cref{tab:industrial}.
In AUPRO, which weights every defect region equally, \method is the best on four datasets and the second-best on MPDD.
In pixel AUROC, which large defects dominate, it is the highest on Real-IAD and within 0.8\%p of the best auxiliary-trained method on the others.
On average over the five datasets, \method ranks first in all four metrics.
At 25.1\,FPS, it is faster than 11 of the 12 ZSAD methods.
Against the seven decision models, \method also achieves the highest mean image AUROC (88.2 \vs 87.1 for Mammoth~1) and the highest mean image AP.
It ranks first in image AUROC on MVTec AD, MPDD, BTAD, and Real-IAD, and Mammoth~1 ranks first on VisA (88.2 \vs 87.5).
On the VisA categories with multiple objects per image, the maximum of the anomaly map lies on the defect in 85\% of the defect images for \method, compared with 57\% and 25\% on average for the auxiliary-trained and training-free methods (\cref{sec:supp_peak}).
Without self-rarity, this share decreases to 70\%.

\subsection{Comparison with decision models}
\label{sec:exp_decision}
We compare \method with the seven decision models along the three challenges of \cref{sec:pilot} (\cref{tab:decision}).

\begin{table}[t]
\centering
\caption{\textbf{Decision models and \method.}
Means over the five industrial datasets.
Image AUC is in \%, Incons.\ is the polarity inconsistency, and Cost is the API fee per 1k images.
FPS assumes sequential requests.
Best in bold, second underlined.}
\label{tab:decision}
\footnotesize
\setlength{\tabcolsep}{3pt}
\begin{tabular}{l c c r r c}
\toprule
\multicolumn{1}{l}{\multirow{2}{*}{Model}} & Image & \multirow{2}{*}{Incons.\,$\downarrow$} & \multirow{2}{*}{FPS\,$\uparrow$} & Cost & Evid. \\
 & AUC\,$\uparrow$ &  &  & (\$/1k) & map \\
\midrule
Jev 1.14 & 82.8 & 0.11 & \underline{1.7} & 0.042 & -- \\
Decider V1.2 & 86.4 & \underline{0.04} & 0.8 & 0.007 & -- \\
GPT-6 Luna Decisions & 79.8 & 0.38 & 1.4 & 0.042 & -- \\
Clef & 81.1 & 0.27 & 0.6 & 0.098 & -- \\
Mammoth 1 & \underline{87.1} & 0.10 & 1.0 & 0.043 & -- \\
Drex Max & 83.9 & 0.07 & 0.7 & 0.021 & -- \\
Span-02 & 83.4 & 0.10 & 1.2 & 0.004 & -- \\
\midrule
\textbf{DecisionAD} & \textbf{88.2} & \textbf{0.00} & \textbf{25.1} & -- & \checkmark \\
\bottomrule
\end{tabular}

\end{table}

\paragraph{(i) Consistency under negation.}
\method yields the same probability of a defect for a question and its negation whenever the parser reads their polarity (\cref{sec:properties}).
Its polarity inconsistency is zero, against 0.04--0.38 for the decision models (\cref{tab:decision}).
The ranking does not depend on the wording of the question, as ten other questions receive the same image AUROC (\cref{tab:question}).
A decision model also becomes consistent if it always receives one question of a pair and its answer is inverted for the other.
Its image AUROC then depends on which question it receives, by up to 15.6\%p (\cref{sec:pilot}).
For \method, a misread polarity inverts all answers (\cref{sec:supp_parser}).
As the question is fixed for a deployment, its polarity is confirmed once at setup, like the answer type of a decision API.

\paragraph{(ii) Evidence.}
Every answer of \method comes with a map, whose pixel metrics are the best on average among the ZSAD methods (\cref{tab:industrial}).
Under the box evaluation of \cref{sec:supp_box}, its AUPRO is 86--93, against 32--48 for the box of Jev 1.14.
Unlike this box, the map enters the answer.
$\Pyes$ is computed from its window statistics and increases with each (\cref{sec:properties}).
The decision models return no map.

\paragraph{(iii) Throughput.}
At 25.1\,FPS, \method is about 15$\times$ faster than the fastest decision model (1.66\,FPS) and keeps the images on site.

\begin{table}[t]
\centering
\caption{\textbf{Component ablation.} The first row is our setting (\%), and the other rows are changes from it in \%p, with red lower and blue higher. The answer score enters the map through $z_A$ and the answer statistic $u$ through $z_s$. A dot marks an unchanged map.}
\label{tab:ablation}
\fitbox{\scriptsize
\setlength{\tabcolsep}{2.2pt}
\begin{tabular}{l cccccc}
\toprule
\multicolumn{1}{l}{\multirow{3}{*}[-1\dimexpr\aboverulesep+\belowrulesep\relax]{Setting}} & \multicolumn{3}{c}{MVTec AD} & \multicolumn{3}{c}{VisA} \\
\cmidrule(lr){2-4} \cmidrule(lr){5-7}
 & Image & \multicolumn{2}{c}{Pixel} & Image & \multicolumn{2}{c}{Pixel} \\
\cmidrule(lr){2-2} \cmidrule(lr){3-4} \cmidrule(lr){5-5} \cmidrule(lr){6-7}
 & AUC & AP & PRO & AUC & AP & PRO \\
\midrule
\rowcolor{oursrow}\textbf{DecisionAD} & 96.0 & 45.8 & 90.4 & 87.5 & 23.9 & 90.5 \\
\midrule
\multicolumn{7}{l}{\textit{(a) Answer score removed from}} \\
\quad map and $u$ (image only) & \cellcolor{red!40}$-$8.7 & \cellcolor{red!31}$-$4.4 & \cellcolor{red!29}$-$4.0 & \cellcolor{red!40}$-$7.1 & \cellcolor{red!21}$-$2.7 & \cellcolor{red!21}$-$2.6 \\
\quad map ($Z{=}z_R$) & \cellcolor{red!14}$-$1.5 & \cellcolor{red!31}$-$4.4 & \cellcolor{red!29}$-$4.0 & \cellcolor{blue!12}$+$1.2 & \cellcolor{red!21}$-$2.7 & \cellcolor{red!21}$-$2.6 \\
\quad $u$ ($u{=}z_L$) & \cellcolor{red!16}$-$1.9 & $\cdot$ & $\cdot$ & \cellcolor{red!33}$-$4.6 & $\cdot$ & $\cdot$ \\
\midrule
\multicolumn{7}{l}{\textit{(b) Image side removed}} \\
\quad map from $u$ ($u{=}z_s$) & $-$0.3 & $\cdot$ & $\cdot$ & \cellcolor{red!14}$-$1.5 & $\cdot$ & $\cdot$ \\
\quad self-rarity ($Z{=}z_A$) & \cellcolor{red!9}$-$0.6 & \cellcolor{red!40}$-$7.1 & \cellcolor{red!20}$-$2.5 & \cellcolor{red!9}$-$0.6 & \cellcolor{red!27}$-$3.7 & \cellcolor{red!30}$-$4.1 \\
\quad refinement & $-$0.1 & \cellcolor{red!18}$-$2.2 & \cellcolor{red!8}$-$0.5 & \cellcolor{red!9}$-$0.7 & \cellcolor{red!14}$-$1.5 & 0.0 \\
\quad multi-scale pooling & \cellcolor{red!17}$-$2.0 & $\cdot$ & $\cdot$ & \cellcolor{red!20}$-$2.5 & $\cdot$ & $\cdot$ \\
\bottomrule
\end{tabular}
}
\end{table}

\subsection{Ablation}
\label{sec:exp_ablation}
\begin{figure}[t]
\centering
\includegraphics[width=\linewidth]{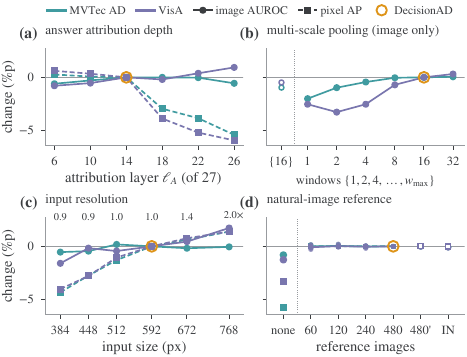}
\caption{\textbf{Design sweeps} on MVTec AD and VisA, as changes from our setting (\%p).
(a) Attribution layer. (b) Window sets. (c) Input size, with the time relative to our setting on top. (d) Natural-image reference, where ``none'' is per-image standardization, 480$'$ a disjoint COCO set, and IN Imagenette.}
\label{fig:ablation}
\end{figure}

\Cref{tab:ablation} ablates the components and \cref{fig:ablation} the design values, and \cref{sec:supp_ablation} varies the other choices.
\textbf{Answer score and attribution.} Removing the answer score and its attribution lowers the image AUROC by 8.7 and 7.1\%p on MVTec AD and VisA (\cref{tab:ablation}a). The attribution adds 4.4 and 2.7\%p of pixel AP. On VisA, the attribution lowers the image AUROC by 1.2\%p. The loss comes from pcb1 alone, whose image AUROC is 51.9 with the attribution and 67.7 without it. On this category, the attribution also marks components that are present on every board.

\textbf{Map and answer score.} Either alone yields a lower image AUROC, as the score reflects the whole image and the map the strongest local evidence.

\textbf{Image-side parts.} Self-rarity does not use text and marks structure that the prompts do not describe. Without it, pixel AP decreases by 7.1 and 3.7\%p, and without refinement by 2.2 and 1.5\%p. The smallest window alone, which gives the map maximum, lowers image AUROC by 2.0--2.5\%p, and larger windows have a marginal effect (\cref{fig:ablation}b).

\textbf{Design choices.} A deeper attribution layer lowers pixel AP by up to about 6\%p, and a shallower one lowers image AUROC (\cref{fig:ablation}a). The smallest input lowers pixel AP by about 4\%p, and the largest doubles the time (\cref{fig:ablation}c). Other reference images differ by at most 0.3\%p, but per-image standardization lowers pixel AP by 3--6\%p (\cref{fig:ablation}d).

\begin{figure}[t]
\centering
\includegraphics[width=\linewidth]{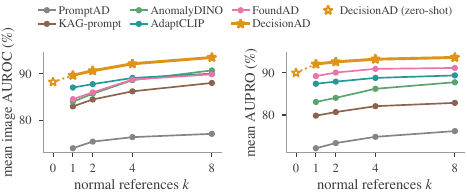}
\caption{\textbf{Few-shot extension.} Means over the five industrial datasets against the number of normal references $k$.}
\label{fig:fewshot}
\end{figure}

\subsection{Few-shot extension}
\label{sec:exp_fewshot}
Given $k$ normal images, \method changes two steps without any training (\cref{sec:supp_fewshot}).
Self-rarity is replaced by \emph{reference rarity}, the distance of each patch to its most similar reference patch.
The answer score is centered on its mean over the references.
All other steps are unchanged, including the natural-image scale.
We compare it with five few-shot baselines on the five industrial datasets.
It achieves the highest image AUROC on every dataset for all $k\in\{1,2,4,8\}$, and also the highest mean AUPRO (\cref{fig:fewshot}).
Without references, its mean image AUROC (88.2) exceeds that of all baselines with one or two references.
It is also the fastest method for all $k$.

\begin{figure}[t]
\centering
\includegraphics[width=\linewidth]{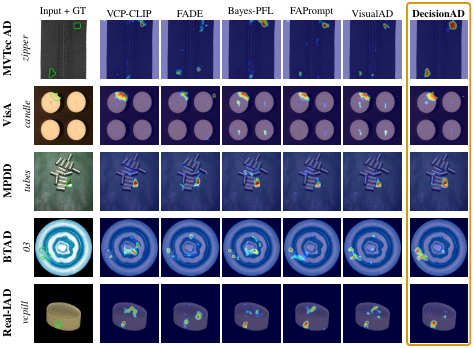}
\caption{\textbf{Qualitative comparison} with the best training-free method and the four best auxiliary-trained methods in mean AUPRO. Green contours mark the ground truth, and each map colors its top 5\% of pixels.}
\label{fig:qual}
\end{figure}

\subsection{Qualitative results}
\label{sec:exp_qual}
\Cref{fig:qual} compares the maps, and \cref{sec:supp_qualfig} shows all 12 baselines.
On the MVTec AD zipper, \method covers each of several defects.
On the VisA candle and the MPDD tubes, it marks small defects among repeated objects.
On BTAD and the Real-IAD vcpill, most baselines highlight other parts of the object, whereas \method concentrates on the defect.
\method fails mainly on three types of anomaly.
The first is an anomaly of the whole object, such as a flipped nut or a pill of another type, where every patch appears intact.
The others are missing components of populated boards and very small defects.
One normal reference image raises the mean image AUROC of the whole-object types from 57.0 to 75.6 (\cref{sec:supp_failure}).

\section{Conclusion}
\label{sec:conclusion}

Visual inspection needs an answer and its evidence.
ZSAD methods return only a map, and commercial decision models only an answer that changes under negation.
\method returns both an answer and a map, without training.
Its answers are consistent under negation whenever the polarity is parsed, and each answer comes with the map that enters it.
On five industrial datasets, \method achieves the highest mean image AUROC among 12 ZSAD methods and seven decision models.
It processes images about 15$\times$ faster than these models under sequential requests.
With few normal images, it exceeds few-shot baselines.

\paragraph{Limitations.}
$\Pyes$ is not a calibrated probability of a defect.
As calibration needs labels of the target, \method instead measures the evidence against unlabeled natural images.
Their choice has a marginal effect on the results (\cref{sec:supp_ablation}).
When a few normal images are available, the answer score is instead centered on them (\cref{sec:exp_fewshot}).
On medical images, the decision models achieve a higher image AUROC, as analyzed in \cref{sec:supp_medical}.
They still return no map, whereas \method has the best AUPRO among the ZSAD methods on three of the six pixel-level datasets.
\method misses anomalies of the whole object, where every patch appears intact.
The 12 baselines miss them as well, and a single normal reference image reduces these misses (\cref{sec:supp_failure}).
\method answers only yes/no questions about a defect, not about its kind, whereas a decision model accepts other answer types.
Its rule-based parser misreads double and implicit negations, and an embedding-based parser reads fewer questions (\cref{sec:supp_parser}).
Label-free calibration and other answer types are future work.

\section*{LLM Usage Statement}
We used Claude (Anthropic), a large language model assistant, to edit the text of this paper, to search the related literature, and to write and debug code for the experiments, tables, and figures.
It also wrote 60 of the 104 questions of \cref{sec:supp_parser}, without access to the parser rules.
The authors checked all text, references, code, and reported numbers, and take full responsibility for the content.
The commercial decision models in \cref{sec:pilot} are objects of evaluation, not tools used in writing.

{
    \small
    \bibliographystyle{ieeenat_fullname}
    \bibliography{main}
}

\clearpage
\setcounter{table}{0}\setcounter{figure}{0}\setcounter{equation}{0}\setcounter{algorithm}{0}
\renewcommand{\theHtable}{S\arabic{table}}\renewcommand{\theHfigure}{S\arabic{figure}}
\renewcommand{\theHequation}{S\arabic{equation}}\renewcommand{\theHalgorithm}{S\arabic{algorithm}}
\renewcommand{\SuppSec}{Sec.}\renewcommand{\SuppSecs}{Secs.}\renewcommand{\SuppSection}{Section}\renewcommand{\SuppSections}{Sections}
\onecolumn
\renewcommand{\thetable}{S\arabic{table}}
\renewcommand{\thefigure}{S\arabic{figure}}
\renewcommand{\theequation}{S\arabic{equation}}
\renewcommand{\thealgorithm}{S\arabic{algorithm}}
\begin{center}
{\Large\textbf{\thetitle}}\\[0.5em]
{\Large Supplementary Material}
\end{center}
\vspace{1em}
\appendix
\crefalias{section}{appendix}\crefalias{subsection}{subappendix}\crefalias{subsubsection}{subsubappendix}
\noindent Section letters (A--I) and numbers with the prefix S refer to the supplementary material. All other numbers refer to the main paper.
In the tables, AUC denotes AUROC and PRO denotes AUPRO@0.30, as in the main paper, and F1 denotes F1-max.
Where values are marked, the best is in bold and the second-best is underlined, unless a caption states otherwise.
\medskip

\setcounter{topnumber}{3}\setcounter{bottomnumber}{2}\setcounter{totalnumber}{4}
\renewcommand{\topfraction}{0.9}\renewcommand{\bottomfraction}{0.6}\renewcommand{\textfraction}{0.1}\renewcommand{\floatpagefraction}{0.75}

\noindent
\begin{tabular}{@{}l@{\hspace{0.8em}}p{0.94\linewidth}@{}}
\ref{sec:supp_decision} & Decision models: questions, per-dataset results, polarity inconsistency, box answers, requests and served models \\
\ref{sec:supp_impl} & Implementation details: settings, inference procedure, prompt sets, natural-image reference, datasets, baselines \\
\ref{sec:supp_question} & Handling the question: how the question is used, rule-based parser, question types \\
\ref{sec:supp_full} & Full zero-shot results: all image and pixel metrics, intervals of the margins \\
\ref{sec:supp_efficiency} & Efficiency: model size, computation, memory, speed, timing protocol, cost of each variant \\
\ref{sec:supp_ablation} & Ablations: layers and encoders, encoder and method, windows, prompt pair, refinement, map weights, prompt templates, natural-image reference, resolution \\
\ref{sec:supp_fewshot} & Few-shot extension: method, results, normal references \\
\ref{sec:supp_qual} & Analysis: qualitative comparison, results by defect size and image content, location of the map maximum, failure analysis \\
\ref{sec:supp_medical} & Medical images: comparison with ZSAD methods and decision models, role of the answer score \\
\end{tabular}

\FloatBarrier
\section{Decision models}
\label{sec:supp_decision}
This section presents the questions, the per-dataset results, the polarity study, the box answers, and the requests and served versions of the decision models of \cref{sec:pilot}.

\subsection{Questions and per-dataset results}
Each request sends one image, letterboxed to a 512\,px square, and one yes/no question of \cref{tab:dm_questions}.
The returned value $\Pyes$ is the probability of the answer yes, and no labels, file names, or categories are sent.
The presence questions yield the image AUROC, and the presence and absence questions together yield the polarity inconsistency.
\method is evaluated with the same questions.
Its parser reads the presence questions as $\pi=+1$ and the absence questions as $\pi=-1$, as the industrial absence question contains the cue ``free of'' and the medical absence question the cues ``normal'' and ``without''.
Its results therefore come from the presence questions, and its polarity inconsistency is zero.
\Cref{tab:decision} summarizes the seven models and \method, and \cref{tab:supp_decision} lists the image AUROC per dataset.
On the three medical datasets, the decision models have the higher image AUROC (95.7--98.5 against 93.1), which \cref{sec:supp_medical} analyzes.
\begin{table}[!htbp]
\centering
\caption{\textbf{Questions sent to the decision models} (yes/no questions, the same questions are given to \method).}
\label{tab:dm_questions}
\small
\begin{tabular}{@{}l l l l@{}}
\toprule
Domain & Polarity & Question & Datasets \\
\midrule
industrial & presence & Does this industrial inspection image contain a visible defect or anomaly? & all five \\
industrial & absence & Is this industrial inspection image free of any visible defect or anomaly? & all five \\
medical & presence & Does this medical image contain a visible abnormality or lesion? & HeadCT, BrainMRI, Br35H \\
medical & absence & Is this medical image normal, without any visible abnormality or lesion? & HeadCT, BrainMRI, Br35H \\
\bottomrule
\end{tabular}
\end{table}
\begin{table}[!htbp]
\centering
\caption{\textbf{Image AUROC of the decision models and \method per dataset} (\%).}
\label{tab:supp_decision}
\fitbox{\footnotesize
\setlength{\tabcolsep}{2.6pt}
\begin{tabular}{l cccccccccc}
\toprule
\multicolumn{1}{l}{Model} & MVTec AD & VisA & MPDD & BTAD & Real-IAD & HeadCT & BrainMRI & Br35H & Ind.\ mean & Med.\ mean \\
\midrule
Jev 1.14 & 92.5 & 85.3 & 62.8 & 92.3 & \underline{81.1} & 98.2 & 93.2 & 96.4 & 82.8 & 95.9 \\
Decider V1.2 & 92.1 & \underline{87.5} & 79.9 & 91.5 & 80.9 & \underline{99.6} & 94.6 & 96.7 & 86.4 & 96.9 \\
GPT-6 Luna Decisions & 87.2 & 80.2 & 76.2 & 79.0 & 76.6 & \textbf{99.8} & 94.7 & 97.3 & 79.8 & 97.3 \\
Clef & 91.5 & 78.9 & 77.3 & 81.0 & 76.7 & \underline{99.6} & 95.0 & 97.6 & 81.1 & 97.4 \\
Mammoth 1 & \underline{93.0} & \textbf{88.2} & \underline{80.4} & \underline{92.7} & 81.0 & 99.5 & \textbf{97.3} & \textbf{98.7} & \underline{87.1} & \textbf{98.5} \\
Drex Max & 90.7 & 85.6 & 73.6 & 89.9 & 79.9 & 99.3 & 92.6 & 95.2 & 83.9 & 95.7 \\
Span-02 & 90.4 & 82.5 & 77.7 & 88.3 & 78.0 & \underline{99.6} & \underline{96.5} & \underline{98.2} & 83.4 & \underline{98.1} \\
\midrule
\textbf{DecisionAD} & \textbf{96.0} & \underline{87.5} & \textbf{80.5} & \textbf{96.0} & \textbf{81.3} & 96.3 & 90.5 & 92.7 & \textbf{88.2} & 93.1 \\
\bottomrule
\end{tabular}
}
\end{table}

\subsection{Polarity inconsistency}
\begin{figure}[!htbp]
\centering
\includegraphics[width=\linewidth]{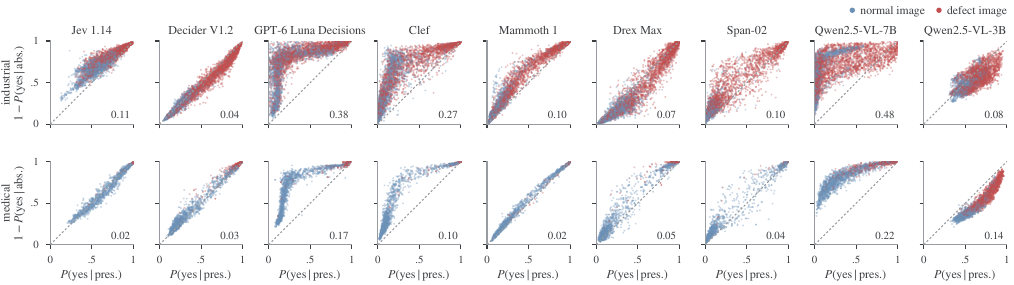}
\caption{\textbf{Presence and absence answers of the seven decision models and two open models} (one dot per image).
The top row shows four industrial datasets (MVTec AD, VisA, MPDD, BTAD), and the bottom row three medical datasets (HeadCT, BrainMRI, Br35H).
On the diagonal, the two probabilities sum to one. The number is the mean polarity inconsistency over the five industrial or the three medical datasets (\cref{tab:supp_gap}).}
\label{fig:supp_flip}
\end{figure}

For one image, a consistent model yields the same probability of a defect for a presence question $Q^{+}$ and an absence question $Q^{-}$.
The \emph{polarity inconsistency} is $|P(\text{yes}\mid Q^{+})+P(\text{yes}\mid Q^{-})-1|$, which needs no threshold.
\Cref{tab:supp_gap} lists its mean per dataset and the share of industrial images on which it exceeds 0.2, and \cref{fig:supp_flip} shows the two answers of every model per image.
The study covers 56 model--dataset pairs, namely the seven decision models on the five industrial and the three image-level medical datasets.
The main paper reports the 35 industrial pairs, and \cref{sec:supp_medical} discusses the medical pairs.
\begin{table}[!htbp]
\centering
\caption{\textbf{Polarity inconsistency per dataset} (mean over the images, category macro, lower is better). The last column is the mean share of industrial images with an inconsistency above 0.2 (\%). The two Qwen2.5-VL rows are open vision-language models.}
\label{tab:supp_gap}
\fitbox{\footnotesize
\setlength{\tabcolsep}{3pt}
\begin{tabular}{l cccccccc cc c}
\toprule
\multicolumn{1}{l}{Model} & MVTec AD & VisA & MPDD & BTAD & Real-IAD & HeadCT & BrainMRI & Br35H & Ind.\ mean & Med.\ mean & $>0.2$ (\%) \\
\midrule
Jev 1.14 & 0.07 & 0.09 & 0.16 & 0.11 & 0.12 & 0.02 & 0.01 & 0.02 & 0.11 & 0.02 & 15.2 \\
Decider V1.2 & 0.05 & 0.03 & 0.06 & 0.03 & 0.04 & 0.05 & 0.02 & 0.03 & 0.04 & 0.03 & 0.5 \\
GPT-6 Luna Decisions & 0.29 & 0.43 & 0.46 & 0.34 & 0.38 & 0.23 & 0.11 & 0.16 & 0.38 & 0.17 & 71.8 \\
Clef & 0.19 & 0.22 & 0.26 & 0.42 & 0.26 & 0.14 & 0.06 & 0.09 & 0.27 & 0.10 & 59.9 \\
Mammoth 1 & 0.07 & 0.08 & 0.08 & 0.16 & 0.09 & 0.02 & 0.01 & 0.02 & 0.10 & 0.02 & 12.5 \\
Drex Max & 0.06 & 0.08 & 0.09 & 0.07 & 0.07 & 0.07 & 0.03 & 0.05 & 0.07 & 0.05 & 4.9 \\
Span-02 & 0.09 & 0.10 & 0.12 & 0.11 & 0.09 & 0.07 & 0.02 & 0.03 & 0.10 & 0.04 & 9.0 \\
\midrule
Qwen2.5-VL-7B~\cite{bai2025qwen25vl} & 0.36 & 0.43 & 0.51 & 0.65 & 0.47 & 0.22 & 0.19 & 0.24 & 0.48 & 0.22 & 89.9 \\
Qwen2.5-VL-3B~\cite{bai2025qwen25vl} & 0.11 & 0.07 & 0.06 & 0.08 & 0.09 & 0.14 & 0.13 & 0.14 & 0.08 & 0.14 & 4.9 \\
\midrule
\textbf{DecisionAD} & 0.00 & 0.00 & 0.00 & 0.00 & 0.00 & 0.00 & 0.00 & 0.00 & 0.00 & 0.00 & 0.0 \\
\bottomrule
\end{tabular}
}
\end{table}

\paragraph{Ranking metrics hide the inconsistency.}
The AUROCs of the two questions differ by at most 3\%p for 51 of the 56 pairs and for 31 of the 35 industrial pairs.
The median Spearman correlation of the two answers is 0.95 over all pairs and 0.94 over the industrial pairs (\cref{tab:supp_pairs}).
The inconsistency nevertheless ranges from 0.01 to 0.46 across the 56 pairs.
The mean sum of the two probabilities is below one for 45 of the 56 pairs (26 of the 35 industrial pairs).
In these pairs, the absence question yields a higher probability of a defect, and their dots in \cref{fig:supp_flip} lie mostly above the diagonal.
\begin{table}[!htbp]
\centering
\caption{\textbf{Every pair of a decision model and a dataset in the polarity study.}
Image AUROC of the presence and the absence question (\%, for the absence question, the score is $1-\Pyes$), Spearman correlation $\rho$ of the two answers, mean of $P^{+}+P^{-}$ with $P^{\pm}=P(\text{yes}\mid Q^{\pm})$, and polarity inconsistency (category macro, images answered under both questions).}
\label{tab:supp_pairs}
\begin{minipage}[t]{0.49\linewidth}\centering\scriptsize
\setlength{\tabcolsep}{5pt}
\renewcommand{\arraystretch}{0.9}
\begin{tabular}[t]{l cc c c c}
\toprule
\multicolumn{1}{l}{\multirow{2}{*}[-0.5\dimexpr\aboverulesep+\belowrulesep\relax]{Model}} & \multicolumn{2}{c}{Image AUC} & \multirow{2}{*}[-0.5\dimexpr\aboverulesep+\belowrulesep\relax]{$\rho$} & Mean of & \multirow{2}{*}[-0.5\dimexpr\aboverulesep+\belowrulesep\relax]{Incons.} \\
\cmidrule(lr){2-3}
 & Pres. & Abs. &  & $P^{+}+P^{-}$ &  \\
\midrule
\multicolumn{6}{l}{\textit{MVTec AD}} \\
Jev 1.14 & 92.5 & 92.9 & 0.98 & 0.93 & 0.07 \\
Decider V1.2 & 92.1 & 91.6 & 0.99 & 1.04 & 0.05 \\
GPT-6 Luna Decisions & 87.2 & 84.6 & 0.85 & 0.71 & 0.29 \\
Clef & 91.5 & 90.3 & 0.96 & 0.82 & 0.19 \\
Mammoth 1 & 93.0 & 92.5 & 0.98 & 0.93 & 0.07 \\
Drex Max & 90.7 & 91.5 & 0.98 & 1.05 & 0.06 \\
Span-02 & 90.4 & 90.2 & 0.97 & 0.92 & 0.09 \\
\midrule
\multicolumn{6}{l}{\textit{VisA}} \\
Jev 1.14 & 85.3 & 86.1 & 0.94 & 0.91 & 0.09 \\
Decider V1.2 & 87.5 & 87.4 & 0.99 & 1.03 & 0.03 \\
GPT-6 Luna Decisions & 80.2 & 81.3 & 0.77 & 0.57 & 0.43 \\
Clef & 78.9 & 80.7 & 0.92 & 0.78 & 0.22 \\
Mammoth 1 & 88.2 & 87.7 & 0.98 & 0.92 & 0.08 \\
Drex Max & 85.6 & 86.4 & 0.97 & 1.07 & 0.08 \\
Span-02 & 82.5 & 82.7 & 0.90 & 0.91 & 0.10 \\
\midrule
\multicolumn{6}{l}{\textit{MPDD}} \\
Jev 1.14 & 62.8 & 67.7 & 0.90 & 0.84 & 0.16 \\
Decider V1.2 & 79.9 & 81.6 & 0.97 & 1.06 & 0.06 \\
GPT-6 Luna Decisions & 76.2 & 68.9 & 0.83 & 0.54 & 0.46 \\
Clef & 77.3 & 79.1 & 0.93 & 0.74 & 0.26 \\
Mammoth 1 & 80.4 & 80.4 & 0.96 & 0.93 & 0.08 \\
Drex Max & 73.6 & 76.5 & 0.95 & 1.08 & 0.09 \\
Span-02 & 77.7 & 73.5 & 0.88 & 0.88 & 0.12 \\
\midrule
\multicolumn{6}{l}{\textit{BTAD}} \\
Jev 1.14 & 92.3 & 91.8 & 0.61 & 0.89 & 0.11 \\
Decider V1.2 & 91.5 & 89.5 & 0.94 & 0.99 & 0.03 \\
GPT-6 Luna Decisions & 79.0 & 75.8 & 0.71 & 0.66 & 0.34 \\
Clef & 81.0 & 82.9 & 0.74 & 0.58 & 0.42 \\
Mammoth 1 & 92.7 & 92.7 & 0.96 & 0.84 & 0.16 \\
Drex Max & 89.9 & 89.8 & 0.87 & 1.00 & 0.07 \\
Span-02 & 88.3 & 88.7 & 0.80 & 0.89 & 0.11 \\
\bottomrule
\end{tabular}
\end{minipage}\hfill
\begin{minipage}[t]{0.49\linewidth}\centering\scriptsize
\setlength{\tabcolsep}{5pt}
\renewcommand{\arraystretch}{0.9}
\begin{tabular}[t]{l cc c c c}
\toprule
\multicolumn{1}{l}{\multirow{2}{*}[-0.5\dimexpr\aboverulesep+\belowrulesep\relax]{Model}} & \multicolumn{2}{c}{Image AUC} & \multirow{2}{*}[-0.5\dimexpr\aboverulesep+\belowrulesep\relax]{$\rho$} & Mean of & \multirow{2}{*}[-0.5\dimexpr\aboverulesep+\belowrulesep\relax]{Incons.} \\
\cmidrule(lr){2-3}
 & Pres. & Abs. &  & $P^{+}+P^{-}$ &  \\
\midrule
\multicolumn{6}{l}{\textit{Real-IAD}} \\
Jev 1.14 & 81.1 & 80.7 & 0.92 & 0.88 & 0.12 \\
Decider V1.2 & 80.9 & 80.3 & 0.98 & 1.03 & 0.04 \\
GPT-6 Luna Decisions & 76.6 & 75.3 & 0.87 & 0.62 & 0.38 \\
Clef & 76.7 & 76.8 & 0.94 & 0.74 & 0.26 \\
Mammoth 1 & 81.0 & 80.2 & 0.98 & 0.91 & 0.09 \\
Drex Max & 79.9 & 79.6 & 0.97 & 1.06 & 0.07 \\
Span-02 & 78.0 & 76.9 & 0.92 & 0.92 & 0.09 \\
\midrule
\multicolumn{6}{l}{\textit{HeadCT}} \\
Jev 1.14 & 98.2 & 97.6 & 0.94 & 1.00 & 0.02 \\
Decider V1.2 & 99.6 & 99.6 & 0.98 & 0.95 & 0.05 \\
GPT-6 Luna Decisions & 99.8 & 99.7 & 0.93 & 0.77 & 0.23 \\
Clef & 99.6 & 99.5 & 0.98 & 0.86 & 0.14 \\
Mammoth 1 & 99.5 & 99.5 & 0.99 & 0.99 & 0.02 \\
Drex Max & 99.3 & 99.5 & 0.94 & 0.94 & 0.07 \\
Span-02 & 99.6 & 99.6 & 0.98 & 0.94 & 0.07 \\
\midrule
\multicolumn{6}{l}{\textit{BrainMRI}} \\
Jev 1.14 & 93.2 & 91.2 & 0.92 & 1.01 & 0.01 \\
Decider V1.2 & 94.6 & 95.1 & 0.98 & 0.98 & 0.02 \\
GPT-6 Luna Decisions & 94.7 & 94.3 & 0.92 & 0.89 & 0.11 \\
Clef & 95.0 & 95.4 & 0.95 & 0.94 & 0.06 \\
Mammoth 1 & 97.3 & 97.1 & 0.99 & 0.99 & 0.01 \\
Drex Max & 92.6 & 95.9 & 0.85 & 0.97 & 0.03 \\
Span-02 & 96.5 & 97.1 & 0.93 & 0.98 & 0.02 \\
\midrule
\multicolumn{6}{l}{\textit{Br35H}} \\
Jev 1.14 & 96.4 & 94.0 & 0.94 & 1.01 & 0.02 \\
Decider V1.2 & 96.7 & 97.1 & 0.98 & 0.98 & 0.03 \\
GPT-6 Luna Decisions & 97.3 & 96.7 & 0.96 & 0.84 & 0.16 \\
Clef & 97.6 & 97.7 & 0.97 & 0.91 & 0.09 \\
Mammoth 1 & 98.7 & 98.7 & 0.99 & 0.98 & 0.02 \\
Drex Max & 95.2 & 97.3 & 0.90 & 0.96 & 0.05 \\
Span-02 & 98.2 & 98.3 & 0.96 & 0.97 & 0.03 \\
\bottomrule
\end{tabular}
\end{minipage}
\end{table}

\paragraph{Either question yields a similar AUROC.}
\Cref{tab:supp_polarity_acc} compares the image AUROC with the presence question, with the absence question (score $1-\Pyes$), and with the mean of the two scores.
With the questions of \cref{tab:dm_questions}, the three differ by at most 3.1\%p for a model, and averaging the two questions changes the image AUROC of the presence question by only $-0.2$ to $+1.3$\%p.
For these questions, the polarity thus shifts the probabilities of a model, while its mean image AUROC changes by at most 3.1\%p.
\begin{table}[!htbp]
\centering
\caption{\textbf{Image AUROC under each polarity} (industrial mean over five datasets, category macro, \%, images answered under both questions). For the absence question, the score is $1-\Pyes$. Both is the mean of the two scores.}
\label{tab:supp_polarity_acc}
\footnotesize
\setlength{\tabcolsep}{6pt}
\begin{tabular}{l ccc}
\toprule
\multicolumn{1}{l}{Model} & Presence & Absence & Both \\
\midrule
Jev 1.14 & 82.8 & 83.8 & 83.5 \\
Decider V1.2 & 86.4 & 86.1 & 86.4 \\
GPT-6 Luna Decisions & 79.8 & 77.2 & 80.3 \\
Clef & 81.1 & 81.9 & 82.4 \\
Mammoth 1 & 87.1 & 86.7 & 86.9 \\
Drex Max & 83.9 & 84.8 & 84.9 \\
Span-02 & 83.4 & 82.4 & 83.3 \\
\bottomrule
\end{tabular}

\end{table}

\paragraph{Other wordings.}
We repeat the measurement with six more question pairs, namely pairs 2--7 of \cref{tab:question_list}.
They were fixed before any of them was sent to a model.
On Real-IAD, these pairs use a fixed 10\% sample of the images of every category.
On this sample, the first pair yields the same inconsistency as on all images, to two decimals.
\Cref{tab:supp_wording_pairs} lists the five-dataset means.
No pair brings the inconsistency of a model to zero, as it ranges from 0.04 to 0.49 over the seven pairs and the seven models.
For four of the seven models, the first pair yields the smallest value.
The wording also changes the ranking.
The image AUROC of a model differs by up to 15.6\%p between the two questions of a pair, and the presence AUROC of Mammoth~1 ranges from 74.0 to 86.9.
\method reads the polarity of all 14 questions. Its answers are therefore the same for all pairs, with an image AUROC of 88.1 on these images, against at most 87.2 for a decision model over all 14 questions.
\begin{table}[!htbp]
\centering
\caption{\textbf{Seven question pairs} (mean over the five industrial datasets, category macro, image AUROC in \%, images answered under both questions). Pair 1 is the pair of \cref{tab:dm_questions}. Real-IAD uses the 10\% sample in every row. For the absence question, the score is $1-\Pyes$. The third block is the absolute difference between the two image AUROCs of a pair. The last block is the image AUROC when one question is always sent and the answer is inverted for the other polarity.}
\label{tab:supp_wording_pairs}
\fitbox{\footnotesize
\setlength{\tabcolsep}{4pt}
\begin{tabular}{l ccccccc c}
\toprule
\multicolumn{1}{l}{Pair} & Jev 1.14 & Decider V1.2 & GPT-6 Luna Decisions & Clef & Mammoth 1 & Drex Max & Span-02 & \textbf{DecisionAD} \\
\midrule
\multicolumn{9}{l}{\textit{Polarity inconsistency}} \\
Pair 1 & 0.11 & 0.04 & 0.38 & 0.27 & 0.10 & 0.07 & 0.10 & 0.00 \\
Pair 2 & 0.21 & 0.05 & 0.34 & 0.16 & 0.14 & 0.08 & 0.14 & 0.00 \\
Pair 3 & 0.20 & 0.09 & 0.28 & 0.31 & 0.12 & 0.14 & 0.49 & 0.00 \\
Pair 4 & 0.11 & 0.09 & 0.27 & 0.16 & 0.11 & 0.11 & 0.34 & 0.00 \\
Pair 5 & 0.24 & 0.08 & 0.45 & 0.34 & 0.14 & 0.14 & 0.21 & 0.00 \\
Pair 6 & 0.13 & 0.19 & 0.22 & 0.24 & 0.16 & 0.22 & 0.25 & 0.00 \\
Pair 7 & 0.16 & 0.15 & 0.11 & 0.22 & 0.10 & 0.15 & 0.27 & 0.00 \\
\midrule
\multicolumn{9}{l}{\textit{Image AUC, presence / absence question}} \\
Pair 1 & 82.6 / 83.7 & 86.2 / 85.8 & 79.7 / 77.0 & 80.8 / 81.7 & 86.9 / 86.5 & 83.8 / 84.5 & 83.2 / 82.2 & 88.1 / 88.1 \\
Pair 2 & 82.7 / 84.6 & 85.6 / 85.9 & 80.2 / 68.3 & 84.3 / 83.0 & 86.9 / 87.2 & 83.2 / 83.9 & 84.1 / 83.7 & 88.1 / 88.1 \\
Pair 3 & 83.1 / 83.8 & 84.6 / 84.3 & 80.0 / 77.0 & 83.9 / 78.2 & 86.1 / 85.2 & 83.8 / 81.7 & 82.9 / 77.4 & 88.1 / 88.1 \\
Pair 4 & 81.2 / 80.2 & 81.8 / 79.5 & 69.8 / 74.5 & 76.1 / 76.9 & 85.4 / 80.9 & 81.2 / 78.1 & 80.0 / 79.5 & 88.1 / 88.1 \\
Pair 5 & 84.4 / 84.2 & 85.7 / 85.4 & 78.8 / 75.6 & 82.5 / 83.0 & 86.2 / 85.9 & 83.4 / 83.1 & 83.6 / 83.1 & 88.1 / 88.1 \\
Pair 6 & 73.6 / 82.2 & 69.2 / 84.5 & 69.5 / 73.9 & 70.2 / 79.8 & 74.0 / 83.4 & 67.2 / 82.8 & 76.1 / 81.1 & 88.1 / 88.1 \\
Pair 7 & 82.0 / 70.8 & 84.3 / 69.4 & 75.6 / 73.1 & 80.2 / 68.9 & 85.0 / 76.3 & 81.5 / 67.6 & 82.2 / 69.1 & 88.1 / 88.1 \\
\midrule
\multicolumn{9}{l}{\textit{Difference between the two image AUCs (\%p)}} \\
Pair 1 & 1.1 & 0.3 & 2.7 & 0.9 & 0.4 & 0.8 & 1.0 & 0.0 \\
Pair 2 & 1.9 & 0.3 & 11.9 & 1.3 & 0.3 & 0.8 & 0.4 & 0.0 \\
Pair 3 & 0.8 & 0.3 & 3.0 & 5.7 & 0.9 & 2.0 & 5.5 & 0.0 \\
Pair 4 & 1.0 & 2.3 & 4.8 & 0.8 & 4.5 & 3.1 & 0.5 & 0.0 \\
Pair 5 & 0.2 & 0.3 & 3.2 & 0.5 & 0.3 & 0.3 & 0.5 & 0.0 \\
Pair 6 & 8.6 & 15.3 & 4.4 & 9.6 & 9.4 & 15.6 & 5.0 & 0.0 \\
Pair 7 & 11.2 & 14.9 & 2.5 & 11.2 & 8.7 & 13.9 & 13.1 & 0.0 \\
\midrule
\multicolumn{9}{l}{\textit{Image AUC behind the wrapper, by the question sent}} \\
Pair 1, presence & 82.6 & 86.2 & 79.7 & 80.8 & 86.9 & 83.8 & 83.2 & 88.1 \\
Mean of the 14 & 81.4 & 82.3 & 75.2 & 79.3 & 84.0 & 80.4 & 80.6 & 88.1 \\
Best of the 14 & 84.6 & 86.2 & 80.2 & 84.3 & 87.2 & 84.5 & 84.1 & 88.1 \\
Worst of the 14 & 70.8 & 69.2 & 68.3 & 68.9 & 74.0 & 67.2 & 69.1 & 88.1 \\
\bottomrule
\end{tabular}
}
\end{table}

\paragraph{Open models.}
The inconsistency is not specific to the decision models.
We give the questions of \cref{tab:dm_questions} and the same images to two open vision-language models, Qwen2.5-VL-7B and Qwen2.5-VL-3B~\cite{bai2025qwen25vl}.
The instruction ``Answer with Yes or No.'' follows each question, and $\Pyes$ is read from the first answer token.
The mean polarity inconsistency of the larger model is 0.48 on the industrial and 0.22 on the medical datasets, above that of every decision model with the same questions.
That of the smaller model is 0.08 and 0.14 (\cref{tab:supp_gap,fig:supp_flip}).

\paragraph{Same interface.}
A decision model also becomes consistent if it always receives one question of a pair and its answer is inverted for the other.
Such a wrapper needs the polarity of the question, as \method does, and a choice of the question that is sent.
It changes neither the number of requests nor the time per image.
Behind it, the image AUROC of a model is that of the question sent (last block of \cref{tab:supp_wording_pairs}).
The presence question of pair 1, fixed before any request, is within 3.5\%p of the best of the 14 questions for every model.
The best question differs between the models and has to be chosen with labels (\cref{sec:supp_parser}).
On all images, a model that receives the presence question of pair 1 reaches the image AUROC of \cref{tab:industrial,tab:decision}.
These tables therefore already compare the methods under the same interface.
What separates them is the ranking, the map, and the cost.

\FloatBarrier
\subsection{Box answers}
\label{sec:supp_box}
Of the seven decision models, only Jev 1.14 supports the box answer type~\cite{typesafe2026jev}, which returns one bounding box for a described target with a confidence value, even when the target is not in the image.
We asked it for ``the visible defect or anomaly in this industrial inspection image''.
For medical images, the target is ``the visible abnormality or lesion in this medical image''.
The images are the same 512\,px letterboxed images as for the yes/no questions, from the five industrial and the six pixel-level medical datasets.
Each box is mapped back to its image and drawn as a binary anomaly map.
The confidence value is not used for the map.
Used as an image score, it yields a mean industrial image AUROC of 75.1 (65.1--82.4 on the five datasets), against 82.8 for the yes/no answer of the same model.
\Cref{tab:supp_box} compares the box with the map of \method in pixel AUROC, pixel AP, and AUPRO, computed by the same evaluator for both at a 256\,px working size.
On industrial images, the box also covers part of every normal image, and it is lower than the map in all metrics.
Its AUPRO is 32.0--47.8 on the five datasets, against 85.8--92.8 for the map.
On the medical datasets, whose lesions are large and which have no normal images, the gap is smaller.
The map is higher in 13 of the 18 medical cells of \cref{tab:supp_box}. The box is higher in pixel AUROC on Kvasir and TN3K, in pixel AP on Kvasir and Endo, and in AUPRO on Endo.
The box is a separate answer.
As it does not enter $\Pyes$ and is returned for any value of $\Pyes$, it is not tied to the answer.
\begin{table}[!htbp]
\centering
\caption{\textbf{Box answers of Jev 1.14 and the evidence map of \method} (pixel metrics, category macro, \%). Better in bold. Normal box area is the mean share of a normal image that the box of Jev 1.14 covers. The medical datasets have no normal images.}
\label{tab:supp_box}
\footnotesize
\setlength{\tabcolsep}{2.6pt}
\begin{tabular}{l cc cc cc c}
\toprule
\multicolumn{1}{l}{\multirow{2}{*}[-0.5\dimexpr\aboverulesep+\belowrulesep\relax]{Dataset}} & \multicolumn{2}{c}{Pixel AUC} & \multicolumn{2}{c}{Pixel AP} & \multicolumn{2}{c}{PRO} & Normal \\
\cmidrule(lr){2-3} \cmidrule(lr){4-5} \cmidrule(lr){6-7}
 & Jev 1.14 & DecisionAD & Jev 1.14 & DecisionAD & Jev 1.14 & DecisionAD & box area \\
\midrule
MVTec AD & 70.3 & \textbf{91.6} & 10.2 & \textbf{45.8} & 47.8 & \textbf{90.2} & 9.4 \\
VisA & 65.9 & \textbf{95.1} & 2.8 & \textbf{23.9} & 39.1 & \textbf{90.9} & 4.1 \\
MPDD & 65.1 & \textbf{97.0} & 4.0 & \textbf{32.6} & 36.6 & \textbf{91.1} & 2.8 \\
BTAD & 69.9 & \textbf{96.8} & 6.1 & \textbf{53.8} & 32.0 & \textbf{85.8} & 27.2 \\
Real-IAD & 71.3 & \textbf{96.7} & 1.2 & \textbf{31.1} & 44.7 & \textbf{92.8} & 5.3 \\
\midrule
ISIC & 86.1 & \textbf{90.4} & 62.1 & \textbf{75.6} & 59.3 & \textbf{81.6} & -- \\
ClinicDB & 85.5 & \textbf{89.6} & 40.9 & \textbf{53.5} & 68.9 & \textbf{82.4} & -- \\
ColonDB & 80.9 & \textbf{83.3} & 29.4 & \textbf{36.8} & 63.1 & \textbf{80.7} & -- \\
Kvasir & \textbf{84.9} & 84.1 & \textbf{50.3} & 41.8 & 52.0 & \textbf{52.8} & -- \\
Endo & 85.8 & \textbf{85.9} & \textbf{52.3} & 41.0 & \textbf{70.6} & 68.2 & -- \\
TN3K & \textbf{80.2} & 78.8 & 40.3 & \textbf{40.4} & 48.6 & \textbf{57.8} & -- \\
\bottomrule
\end{tabular}

\end{table}

\FloatBarrier
\subsection{Requests and served models}
\label{sec:supp_requests}
Requests go to the decisions API~\cite{openrouter2026decisions} at \url{https://openrouter.ai/api/alpha/decisions} with one PNG asset referenced by the \texttt{state} field and one question with the answer type \texttt{noul}. The returned \texttt{noul} value is $\Pyes$.
No temperature, seed, system prompt, or token limit is sent.
A request times out after 180\,s and is retried up to six times on network errors and HTTP 408, 429, 500, 502, 503, 520, 522, 524, and 529. Refusals are not retried, and failed images are left out of the metrics. In the runs below, they amount to at most 0.5\% of the requests of a model.
\Cref{tab:repro_api} lists the served model versions.
The industrial presence runs (November 2, 2026) answered all 35,525 requests, the medical presence runs (November 3) all but one refusal of 24,171, and the absence runs (November 3) all but ten of 59,696, of which eight were refusals and two routing errors. The Real-IAD presence run (November 5) answered all 114,551 images for every model, and the Real-IAD absence run (November 9) all but four refusals of 801,857 requests. The box requests to Jev 1.14 (November 9, \cref{sec:supp_box}) were answered for all 5,075 industrial, 114,551 Real-IAD, and 3,185 medical images. The six additional question pairs (November 12, \cref{tab:supp_wording_pairs}) were answered for all but 279 of 426,300 requests on the four smaller industrial datasets and all but 17 of 964,068 requests on the Real-IAD sample. These 296 requests are all refusals of GPT-6 Luna Decisions.

\begin{table}[!htbp]
\centering
\caption{\textbf{Requested and served decision models.}}
\label{tab:repro_api}
\small
\begin{tabular}{@{}l l l@{}}
\toprule
Requested model & Served model & Provider \\
\midrule
\texttt{typesafe/jev-1.14} & \texttt{typesafe/jev-1.14-20261014} & TypeSafe \\
\texttt{perplexity/pplx-decider-v1.2} & \texttt{perplexity/pplx-decider-v1.2-20261016} & Perplexity \\
\texttt{openai/gpt-6-luna-decisions} & \texttt{openai/gpt-6-luna-decisions-20261006} & OpenAI \\
\texttt{cloudflare/clef} & \texttt{cloudflare/clef} & Cloudflare \\
\texttt{baseten/mammoth-1} & \texttt{baseten/mammoth-1-20261021} & BaseTen \\
\texttt{nace-ai/drex-max} & \texttt{nace-ai/drex-max-20261016} & DeepInfra \\
\texttt{respan/span-02} & \texttt{respan/span-02-20261015} & Respan \\
\bottomrule
\end{tabular}
\end{table}

\FloatBarrier
\section{Implementation details}
\label{sec:supp_impl}
This section lists the settings and inputs needed to rerun the experiments.
The code will be released.

\subsection{Settings}
\Cref{tab:impl} lists every setting of \cref{sec:method}.
All values are shared by the 14 datasets, and only the prompt pair differs between industrial and medical images.
Because layers are set by relative depth, the same rule yields layers 12--21 and attribution layer 17 for the 32-layer C-RADIOv4-H.
\begin{table}[!htbp]
\centering
\caption{\textbf{Settings of \method.} $d$ is the number of layers of the encoder. The last column points to the experiments that vary each setting.}
\label{tab:impl}
\small
\begin{tabular}{@{}l p{0.6\linewidth} l@{}}
\toprule
Item & Value & Varied in \\
\midrule
Encoder & C-RADIOv4-SO400M~\cite{nvidia2026cradiov4}, 27 layers, patch size 16. SigLIP2 and SAM3 heads & \cref{sec:supp_abl_layers,sec:supp_encoder} \\
Input & 592\,px square (bicubic, anti-aliased), $37{\times}37$ patch grid. CLS and register tokens excluded from maps & \cref{fig:ablation}c, \cref{sec:supp_abl_other} \\
Rarity layers $\mathcal{L}$ & $[\mathrm{round}(10d/27),\,\mathrm{round}(18d/27)]$, \ie, layers 10--18 & \cref{sec:supp_abl_layers} \\
Attribution layer $\ell_A$ & $\mathrm{round}(14d/27)$, \ie, layer 14 & \cref{fig:ablation}a, \cref{sec:supp_abl_layers} \\
Prompt sets & $|\mathcal{T}|{=}14$ templates, $|\mathcal{S}_{\mathrm{n}}|{=}7$ normal and $|\mathcal{S}_{\mathrm{a}}|{=}5$ anomalous states (98 and 70 sentences). Medical sets of the same sizes & \cref{sec:supp_abl_design}, \cref{sec:supp_abl_other} \\
Example sentences & ``a photo of a flawless object.'', ``a blurry photo of the damaged object.'', ``a medical image of a healthy tissue.'', ``a blurry medical image of the tissue with lesion.'' & -- \\
Feature rarity & $K{=}10$. The patch and its $3{\times}3$ neighbors excluded. $\hat h$ is LayerNorm without affine parameters & -- \\
Rarity maps & 9 layers $\times$ 2 statistics $=$ 18 maps & \cref{sec:supp_abl_layers} \\
Natural-image reference $\nat$ & 480 images drawn from COCO val2017 with a fixed seed & \cref{fig:ablation}d, \cref{sec:supp_abl_other} \\
Fusion & $Z=(\zA+\zR)/\sqrt2$, $u=(\zL+\zs)/\sqrt2$ (equal weights) & \cref{tab:ablation}, \cref{sec:supp_abl_design} \\
Refinement & SAM3-head features, $|\mathcal{K}_i|{=}16$ with $i$ among them, $\gamma{=}0.07$ & \cref{tab:ablation}, \cref{sec:supp_abl_design} \\
Windows & $\win=\{1,2,4,8,16\}$ patches, stride 1 & \cref{fig:ablation}b, \cref{sec:supp_abl_design} \\
Offset $\tau$ & $\Phi^{-1}(0.95)=1.645$. Ranking metrics do not depend on it & -- \\
\bottomrule
\end{tabular}
\end{table}

\subsection{Inference procedure}
\label{sec:supp_algorithm}
\Cref{alg:inference} summarizes the computation of \cref{sec:method} for a single image.
The reference values $\mu_{\nat}(q)$ and $\sigma_{\nat}(q)$ of every standardized quantity $q$ are obtained once by applying the same computation to $\nat$.
\begin{algorithm}[!htbp]
\caption{\method inference for one image. $\mu_{\nat}(q),\sigma_{\nat}(q)$ are constants computed once on $\nat$.}
\label{alg:inference}
\begin{algorithmic}[1]
\Require image $x$, question $Q$, frozen encoder, prompt-pair embeddings $(\ta,\tn)$, $\mu_{\nat}(\cdot),\sigma_{\nat}(\cdot)$
\State $\pi\gets\textsc{Parse}(Q)$ \Comment{$+1$ presence, $-1$ absence}
\State $\{h^{(\ell)},\alpha^{(\ell)}\}_{\ell}, v, f\gets\textsc{Encoder}(x)$ \Comment{one forward pass}
\State $s\gets\cos(v,\ta)-\cos(v,\tn)$
\State $A_i\gets\log\|\partial s/\partial h^{(\ell_A)}_i\|_2$ for every patch $i$ \Comment{backward pass above $\ell_A$ only}
\State $\zR\gets z\big(\mathrm{mean}_{\ell\in\mathcal{L}}[\,z(\kappa^{(\ell)}),\,z(\lambda^{(\ell)})\,]\big)$
\State $Z\gets(\zA+\zR)/\sqrt2$ \Comment{$\zA=z(A)$}
\State $\Zt\gets\textsc{Refine}(Z,f)$
\State $\zL\gets\mathrm{mean}_{w\in\win}\,z\big(\max_p\mathrm{avg}_{\Omega_w(p)}\Zt\big)$
\State $u\gets(\zL+\zs)/\sqrt2$ \Comment{$\zs=z(s)$}
\State \Return $\Pyes=\Phi(\pi(u-\tau))$, $M=\textsc{Upsample}(\Zt)$
\end{algorithmic}
\end{algorithm}

\subsection{Prompt sets}
Industrial sentences replace \texttt{object} in each template of \cref{tab:repro_templates} by each state phrase below. No category name is used.
The normal states are \texttt{object}, \texttt{flawless object}, \texttt{perfect object}, \texttt{unblemished object}, \texttt{object without flaw}, \texttt{object without defect}, and \texttt{object without damage}. The anomalous states are \texttt{damaged object}, \texttt{broken object}, \texttt{object with flaw}, \texttt{object with defect}, and \texttt{object with damage} (98 normal and 70 anomalous sentences).
Medical sentences use the word \texttt{tissue}, the same templates with \texttt{photo} replaced by \texttt{medical image}, the normal states \texttt{tissue}, \texttt{healthy tissue}, \texttt{normal tissue}, \texttt{unremarkable tissue}, \texttt{tissue without abnormality}, \texttt{tissue without lesion}, and \texttt{tissue without disease}, and the anomalous states \texttt{abnormal tissue}, \texttt{diseased tissue}, \texttt{tissue with abnormality}, \texttt{tissue with lesion}, and \texttt{tissue with disease}.
Each sentence embedding of the SigLIP2 text encoder is $\ell_2$-normalized, the embeddings of each side are averaged, and the mean is $\ell_2$-normalized.

\begin{table}[!htbp]
\centering
\caption{\textbf{Prompt templates} (industrial). \texttt{object} is replaced by each state phrase, \eg, \texttt{flawless object} or \texttt{object with defect}. No category name is inserted. Medical templates replace \texttt{photo} by \texttt{medical image} and \texttt{object} by \texttt{tissue}.}
\label{tab:repro_templates}
\small
\begin{tabular}{@{}r l r l@{}}
\toprule
1 & \texttt{a bright photo of a object.} & 8 & \texttt{a blurry photo of a object.} \\
2 & \texttt{a bright photo of the object.} & 9 & \texttt{a photo of a object.} \\
3 & \texttt{a dark photo of the object.} & 10 & \texttt{a photo of the object.} \\
4 & \texttt{a dark photo of a object.} & 11 & \texttt{a photo of the object for visual inspection.} \\
5 & \texttt{a jpeg corrupted photo of a object.} & 12 & \texttt{a photo of a object for visual inspection.} \\
6 & \texttt{a jpeg corrupted photo of the object.} & 13 & \texttt{a photo of the object for anomaly detection.} \\
7 & \texttt{a blurry photo of the object.} & 14 & \texttt{a photo of a object for anomaly detection.} \\
\bottomrule
\end{tabular}
\end{table}

\subsection{Natural-image reference}
\label{sec:supp_refstats}
The reference $\nat$ is drawn from the 5,000 COCO val2017 images, sorted by file name, with \texttt{random.Random(17).sample(files, 480)}. \Cref{tab:repro_coco} lists their image IDs.
Other draws of the same size change the image AUROC by at most 0.2\%p (\cref{sec:supp_abl_other}).
The industrial and medical prompt pairs have separate statistics on the same images, and every ablation recomputes the statistics of the quantities it changes, such as prompts, layers, map combination, or refinement, before those of the windows and of $s$.
The standardized quantities are the 18 per-layer rarity statistics, their mean, $A$, $s$, and the five window statistics.
For a map, the mean and the standard deviation are taken over all $480{\times}37^2$ patches of $\nat$, and for $s$ and the window statistics, over the 480 images.
On $\nat$, the natural logarithm of the gradient magnitude has a standard deviation of 1.21. Under a normal approximation, the central 95\% of the patches span a factor of about 100 in gradient magnitude.
Over all patches of $\nat$, the magnitudes span four orders of magnitude.
The logarithm compresses this range before the standardization, and without it, the mean pixel AP on the five industrial datasets decreases by 5.4\%p (\cref{tab:supp_attr}).
The answer statistic $u$ has a standard deviation of 1.06 on $\nat$.
\begin{table}[!htbp]
\centering
\caption{\textbf{COCO val2017 image IDs of $\nat$} (in sampling order).}
\label{tab:repro_coco}
\parbox{\linewidth}{\tiny\raggedright 493772, 393014, 288430, 347254, 274460, 166509, 512776, 263403, 106330, 22935, 233370, 364166, 398028, 236690, 474854, 300039, 378605, 132796, 523175, 57760, 134722, 184978, 146363, 506004, 530854, 196141, 312421, 512648, 120584, 64574, 292330, 387387, 79031, 479732, 451144, 571718, 136466, 386210, 478474, 319534, 15272, 410510, 353027, 545594, 47828, 336309, 46804, 453860, 356169, 553339, 4134, 383443, 225670, 113589, 529148, 200252, 233033, 499768, 342128, 53909, 252332, 60102, 261061, 181816, 575372, 528314, 506279, 526103, 126110, 234757, 291551, 329455, 301563, 163258, 313454, 89761, 548267, 551780, 257865, 317999, 7818, 47010, 329456, 15956, 445658, 451084, 50331, 186345, 292908, 243075, 70229, 460160, 17178, 536038, 498032, 133418, 380711, 563267, 370270, 173183, 184321, 235836, 146358, 106563, 198960, 46497, 509260, 289586, 297578, 298251, 45728, 155179, 397327, 88951, 294695, 235057, 381971, 456303, 514797, 81766, 515445, 227511, 212072, 505565, 250137, 549220, 157138, 280930, 45550, 461405, 560011, 371042, 461036, 209972, 526706, 385029, 488664, 423104, 74058, 437351, 205647, 215259, 458790, 300913, 178982, 345385, 570448, 23937, 551439, 541055, 491725, 401250, 50149, 311190, 551804, 162092, 528524, 217753, 296224, 349302, 164637, 197528, 411530, 337987, 426329, 564023, 89296, 491216, 101420, 494913, 8532, 181499, 549136, 170278, 426795, 441491, 423617, 476770, 132587, 508730, 521601, 451714, 430056, 9891, 279774, 248334, 174004, 6894, 445722, 320490, 197658, 48153, 94614, 490470, 13004, 23230, 109441, 449432, 286849, 190676, 94751, 22479, 433134, 509014, 209753, 25386, 377670, 119677, 406129, 53994, 511999, 55150, 177934, 511599, 246968, 255401, 338901, 329219, 516143, 554266, 395575, 383606, 6471, 498747, 322352, 78426, 29596, 270122, 308430, 304291, 114049, 315492, 124659, 511384, 229216, 468124, 268831, 419601, 121591, 239537, 234607, 68765, 398810, 293300, 187362, 458768, 347174, 125257, 200162, 477288, 545958, 146831, 279541, 80022, 413689, 415536, 500478, 368212, 495054, 576031, 148999, 531134, 137294, 157847, 560178, 507667, 299609, 466416, 267191, 384661, 61171, 222991, 161879, 222559, 405195, 520264, 304365, 512929, 420230, 475064, 125472, 273711, 228436, 468965, 235064, 475365, 180383, 353096, 6012, 119452, 317433, 466125, 73153, 408120, 82986, 488251, 357978, 229601, 163118, 66135, 282046, 67896, 273642, 322724, 440336, 303908, 143931, 144784, 251065, 292005, 363461, 308466, 548246, 486040, 138550, 414170, 99810, 242678, 223955, 203580, 199681, 516804, 350122, 356424, 255749, 75612, 291619, 382696, 190140, 297698, 9769, 31248, 542089, 555005, 55299, 334399, 465675, 382030, 402096, 581482, 244099, 287667, 426376, 156076, 393469, 199442, 179265, 193926, 344614, 184338, 216516, 359937, 66771, 433192, 185802, 336628, 474095, 122046, 557672, 459662, 418961, 84431, 127624, 414034, 372260, 214200, 378873, 358427, 460927, 316054, 225946, 309391, 365642, 200839, 165713, 9483, 450488, 94157, 384468, 457884, 95707, 470773, 25424, 249219, 449661, 125850, 186632, 151857, 241319, 141328, 31749, 121506, 276284, 417911, 84270, 281759, 244592, 504415, 293858, 490171, 248314, 535253, 29984, 26465, 509008, 270244, 507575, 66886, 51598, 260261, 279927, 333956, 458054, 198928, 474344, 104424, 39477, 352684, 455301, 424521, 311394, 466256, 348045, 206411, 6763, 572462, 109992, 329827, 144003, 361730, 111086, 530162, 76416, 36494, 370478, 105923, 439180, 138639, 185292, 520531, 419408, 42889, 253452, 15338, 565607, 102356, 91779, 135561, 453001, 207306, 292225, 286422, 287874, 577932, 312340, 524742, 14007, 415238, 84752, 363875, 2157, 439522, 99242, 308793, 509258, 306139, 437239, 52565, 462371, 89045, 507081, 446117, 282298, 579091, 119828, 148957, 66561, 304560, 34071, 187513.}
\end{table}

\subsection{Datasets}
\Cref{tab:datasets} summarizes the evaluation data.
We use the official test sets only, and pixel metrics are computed on full-resolution masks without threshold subsampling.
MPDD, BTAD, and the medical datasets are read from the processed \texttt{meta.json} test entries in the format of the AdaCLIP and AnomalyCLIP repositories.
VisA is read from the test rows of \texttt{1cls.csv}, and Real-IAD from the test entries of the JSON files of its 512\,px release.
CVC-ClinicDB and CVC-ColonDB are written ClinicDB and ColonDB in the tables and in the text.
Masks are binarized at values above zero and kept at the native image size, and every method's map is upsampled bilinearly to that size before the metrics are computed.
\begin{table}[!htbp]
\centering
\caption{\textbf{Evaluation datasets} (official test sets).}
\label{tab:datasets}
\small
\begin{tabular}{@{}l l c r l@{}}
\toprule
Dataset & Domain & Categories & Test images & Metrics \\
\midrule
MVTec AD~\cite{bergmann2019mvtec} & industrial & 15 & 1,725 & image, pixel \\
VisA~\cite{zou2022visa} (1-class split) & industrial & 12 & 2,162 & image, pixel \\
MPDD~\cite{jezek2021mpdd} & industrial & 6 & 458 & image, pixel \\
BTAD~\cite{mishra2021btad} & industrial & 3 & 730 & image, pixel \\
Real-IAD~\cite{wang2024realiad} & industrial & 30 & 114,551 & image, pixel \\
\midrule
HeadCT~\cite{salehi2021mkd} & brain CT & 1 & 200 & image \\
BrainMRI~\cite{salehi2021mkd} & brain MRI & 1 & 253 & image \\
Br35H~\cite{hamada2020br35h} & brain MRI & 1 & 3,000 & image \\
ISIC~\cite{gutman2016isic} & skin & 1 & 379 & pixel \\
CVC-ClinicDB~\cite{bernal2015clinicdb} & colonoscopy & 1 & 612 & pixel \\
CVC-ColonDB~\cite{tajbakhsh2016colondb} & colonoscopy & 1 & 380 & pixel \\
Kvasir~\cite{jha2020kvasir} & colonoscopy & 1 & 1,000 & pixel \\
Endo~\cite{hicks2021endotect} & endoscopy & 1 & 200 & pixel \\
TN3K~\cite{gong2021tn3k} & thyroid ultrasound & 1 & 614 & pixel \\
\bottomrule
\end{tabular}
\end{table}

\subsection{Baselines}
\label{sec:supp_baselines}
\Cref{tab:repro_baselines} lists the code and settings of the reproduced baselines.
Auxiliary-trained methods follow the source--target pairing of their papers.
Policy A uses weights trained on VisA for MVTec AD and on MVTec AD for every other dataset, and policy B uses weights trained on MVTec AD for VisA and on VisA for every other dataset.
AA-CLIP and MoECLIP release no weights and are trained with their official code on the full source test set with its labels and masks, as in their papers.
AA-CLIP uses Adam with learning rates of $10^{-5}$ for 5 text epochs and $5{\times}10^{-4}$ for 20 image epochs.
MoECLIP uses Adam with a learning rate of $5{\times}10^{-5}$ for 20 epochs and seed 111.
Both use the last epoch.
For BTAD, WinCLIP, AnoVL, FADE, APRIL-GAN, and Bayes-PFL include the category names 01, 02, and 03 in their prompts.
AA-CLIP and MoECLIP use the descriptions of their official code, namely ``Bright concentric rings in neon yellow and blue tones against a dark blue background, resembling a stylized wave or energy field radiating outward.'' for 01, ``vertical fabric lines in warm, dusty pink and beige tones'' for 02, and ``oval concentric circular rings in gradient shades of blue and white'' for 03.
On medical images, category-based methods receive \texttt{headct}, \texttt{brain\_mri}, \texttt{br35h}, \texttt{isic}, \texttt{ClinicDB}, \texttt{ColonDB}, \texttt{colon} (Kvasir, Endo), and \texttt{tn3k}. AA-CLIP and MoECLIP receive \texttt{scan}, \texttt{skin}, \texttt{thyroid}, and \texttt{colon endoscopy image}.
Image scores outside $[0,1]$ are min--max normalized per run before evaluation. Bayes-PFL, AA-CLIP, and MoECLIP normalize the scores of each category as in their official code. \method uses no category name and no statistic of the evaluated images.

\begin{table}[!htbp]
\centering
\caption{\textbf{Reproduced ZSAD baselines.} Code is the GitHub repository and commit. B/16+ denotes ViT-B/16-plus-240 (LAION-400M), and L/14 denotes OpenAI CLIP ViT-L/14@336. For Weights, -- means none, and A and B are the source--target policies of the text.}
\label{tab:repro_baselines}
\scriptsize
\setlength{\tabcolsep}{3pt}
\begin{tabular}{@{}l l l l l p{0.27\linewidth}@{}}
\toprule
Method & Code & Encoder, input & Weights & Category text & Notes \\
\midrule
WinCLIP & zqhang/Accurate-WinCLIP-pytorch \texttt{eef4f8c} & B/16+, 240\,px & -- & name & reproduction, 32 and 48\,px windows \\
AnoVL & hq-deng/AnoVL \texttt{3a70bfd} & B/16+, 240\,px & -- & name & training-free path (no adapter) \\
FADE & BMVC-FADE/BMVC-FADE \texttt{43d237b} & GEM B/16+, 240--896\,px & -- & name (image) & language-only image score \\
SAA+ & caoyunkang/Segment-Any-Anomaly \texttt{ff564ed} & GroundingDINO-T, SAM-H & -- & official / generic & box and text thresholds 0.1 \\
\midrule
APRIL-GAN & ByChelsea/VAND-APRIL-GAN \texttt{f13b8a6} & L/14, 518\,px & A, released & name & layers 6/12/18/24 \\
AnomalyCLIP & zqhang/AnomalyCLIP \texttt{3911738} & L/14, 518\,px & A, released & object-agnostic & epoch 15 \\
VCP-CLIP & xiaozhen228/VCP-CLIP \texttt{f87a3f6} & L/14, 518\,px & A, released & visual context & -- \\
Bayes-PFL & xiaozhen228/Bayes-PFL \texttt{8f155a0} & L/14, 518\,px & A, released & name & 10 samples, seed 111 \\
AA-CLIP & Mwxinnn/AA-CLIP \texttt{53db195} & L/14, 518\,px & B, trained & description & official code, last epoch \\
FAPrompt & mala-lab/FAPrompt \texttt{506b6bf} & L/14, 518\,px & A, released & object-agnostic & -- \\
MoECLIP & CoCoRessa/MoECLIP \texttt{f9b2574} & L/14, 518\,px & B, trained & description & official code, last epoch \\
VisualAD & 7HHHHH/VisualAD \texttt{ba7a866} & L/14 (visual), 518\,px & B, released & -- & language-free \\
\bottomrule
\end{tabular}
\end{table}

\FloatBarrier
\section{Handling the question}
\label{sec:supp_question}
\method reads only the polarity of the question (\cref{sec:parser}).
This section shows why the question enters only through its polarity, how reliably its polarity is read, and how each type of question is handled.

\subsection{How the question is used}
\Cref{tab:question} compares four ways of using the question on pair 1 and on pairs 3--7 of \cref{tab:question_list}.
\begin{enumerate*}[label=(\roman*)]
\item Without polarity handling, the ranking of absence questions is inverted (image AUROC 10.0).
\item Feeding the question itself to the text encoder lowers the presence question to 83.0 and the absence question to 18.7, since the text encoder does not handle negation. This variant is evaluated on pair 1 only.
\item Building the prompt pair from the defect words of the question is slightly worse than the fixed pair.
\end{enumerate*}
We therefore fix the prompt pair and read only the polarity from the question.
The question-word variant of \cref{tab:question} matches the defect words defect, anomaly, damage, flaw, scratch, crack, dent, hole, contamination, stain, broken, fault, abnormality, irregularity, imperfection, and deformation, with their inflections.
It replaces the anomalous states by \texttt{object with <word>}, or by \texttt{broken object} for broken.
\begin{table}[!htbp]
\centering
\caption{\textbf{How the question is used.}
Mean image AUROC over four industrial datasets (MVTec AD, VisA, MPDD, BTAD, \%).
For absence questions, normal images are the positive (yes) class.
Pair 1 has one presence and one absence question, and pairs 3--7 have five of each.}
\label{tab:question}
\footnotesize
\setlength{\tabcolsep}{3pt}
\begin{tabular}{l cc cc}
\toprule
\multicolumn{1}{l}{\multirow{2}{*}[-0.5\dimexpr\aboverulesep+\belowrulesep\relax]{Variant}} & \multicolumn{2}{c}{Pair 1} & \multicolumn{2}{c}{Pairs 3--7} \\
\cmidrule(lr){2-3} \cmidrule(lr){4-5}
 & Pres. & Abs. & Pres. & Abs. \\
\midrule
Fixed prompt pair + polarity (ours) & \textbf{90.0} & \textbf{90.0} & \textbf{90.0} & \textbf{90.0} \\
Question-word prompt pair + polarity & \underline{88.5} & \underline{88.5} & \underline{88.6} & \underline{89.5} \\
Fixed prompt pair, no polarity & \textbf{90.0} & 10.0 & \textbf{90.0} & 10.0 \\
Question sentence as prompt & 83.0 & 18.7 & -- & -- \\
\bottomrule
\end{tabular}

\end{table}
\begin{table}[!htbp]
\centering
\caption{\textbf{Questions of the question-handling evaluation.}
Pair 1 and pairs 3--7 are also evaluated on images in \cref{tab:question}.}
\label{tab:question_list}
\scriptsize
\setlength{\tabcolsep}{3pt}
\begin{tabular}{@{}l p{0.44\linewidth} p{0.44\linewidth}@{}}
\toprule
Pair & Presence question & Absence question \\
\midrule
1 & Does this industrial inspection image contain a visible defect or anomaly? & Is this industrial inspection image free of visible defects or anomalies? \\
2 & Is there any visible defect or anomaly anywhere in this industrial inspection image? & Are visible defects and anomalies absent throughout this industrial inspection image? \\
3 & Is there any defect visible on this item? & Is this item in perfect condition? \\
4 & Can you see anything wrong with the part in this photo? & Would this part pass a visual quality check? \\
5 & Does this product show any signs of damage or flaws? & Does the object look free of any flaws? \\
6 & Is something abnormal about the object shown here? & Is the product shown here defect-free? \\
7 & Would an inspector find a fault anywhere on this piece? & Can this piece be shipped as a good unit? \\
\bottomrule
\end{tabular}

\end{table}

\subsection{Rule-based parser}
\label{sec:supp_rules}
The parser uses regular expressions on the lowercased question and reads three kinds of cues.
A defect cue indicates a presence question. The defect cues are the forms of \texttt{defect}, \texttt{anomaly}, \texttt{damage}, \texttt{flaw}, \texttt{scratch}, \texttt{crack}, \texttt{dent}, \texttt{hole}, \texttt{contamination}, \texttt{stain}, \texttt{broken}, \texttt{fault}, \texttt{abnormal}, \texttt{irregularity}, \texttt{imperfection}, \texttt{deformation}, \texttt{wrong}, \texttt{problem}, \texttt{issue}, \texttt{bad}, \texttt{differ}, \texttt{deviate}, \texttt{reject}, \texttt{repair}, \texttt{rework}, \texttt{scrap}, \texttt{flag}, \texttt{fail}, \texttt{blemish}, \texttt{worn}, \texttt{wear}, \texttt{missing}, \texttt{unacceptable}, \texttt{impurity}, \texttt{mark}, \texttt{spot}, \texttt{something off}, \texttt{anything off}, \texttt{corrosion}, \texttt{rust}, \texttt{chip}, \texttt{tear}, \texttt{torn}, and \texttt{bent}.
A quality cue indicates an absence question. The quality cues are \texttt{flawless}, \texttt{intact}, \texttt{undamaged}, \texttt{normal}, \texttt{good}, \texttt{ok}, \texttt{okay}, \texttt{acceptable}, \texttt{pass}, \texttt{clean}, \texttt{perfect}, \texttt{unblemished}, \texttt{healthy}, \texttt{all right}, \texttt{alright}, \texttt{fine}, \texttt{spotless}, \texttt{pristine}, \texttt{immaculate}, \texttt{right with}, \texttt{as it should}, \texttt{standard}, \texttt{specification}, \texttt{spec}, \texttt{conform}, \texttt{compliant}, \texttt{sound}, \texttt{in order}, \texttt{proper}, and \texttt{correct}.
Cues match whole words with their inflections, so \texttt{flawless}, \texttt{undamaged}, and \texttt{unblemished} are quality cues.
A defect cue takes precedence over a quality cue.
Each occurrence of a negator flips the reading. The negators are \texttt{no}, \texttt{not}, \texttt{n't}, \texttt{without}, \texttt{free of}, \texttt{free from}, \texttt{-free}, \texttt{nothing}, \texttt{none}, \texttt{never}, \texttt{absent}, \texttt{absence}, \texttt{devoid}, and \texttt{lack}.
``Is it not free of defects?'' is thus read as a presence question, with one defect cue and two negators.
A contracted negation that opens the question, as in ``Isn't there a defect?'', is not counted.
A question without a defect or quality cue is declined.
Its polarity is then declared with the question.
The parser reads the polarity of all 14 questions of \cref{tab:question_list} correctly.

\subsection{Question types}
\label{sec:supp_parser}
We evaluate the handling of the question on 104 whole-image questions of six types (\cref{tab:qtypes}).
A question of the first two types names a defect or a quality. The next two types negate a defect or a quality, as in ``Is there nothing wrong with this part?'' and ``Is this part not acceptable?''
The fifth type uses a double negation or expresses the negation by a predicate, as in ``Is the piece anything other than flawless?''
The sixth type asks how the item is handled and names neither a defect nor a quality, as in ``Is this unit ready to ship?''
The types were assigned from the form of each sentence, and \cref{tab:qlist_a,tab:qlist_b} list every question.
\begin{table*}[!htbp]
\centering
\caption{\textbf{Results by question type} (MVTec AD). For \method, the questions read correctly / declined / misread and the image AUROC over the questions that are not declined. For each decision model, the image AUROC with the intended polarity. AUROC cells list the maximum / minimum over the questions of the type (\%). For \method, a question read correctly receives 96.0 and a misread question 4.0. Answered counts the questions for which every decision model answered at least 95\% of the images.}
\label{tab:qtypes}
\fitbox{\scriptsize
\setlength{\tabcolsep}{3pt}
\begin{tabular}{l c c c c ccccccc}
\toprule
Type & Questions & Read / declined / misread & \textbf{DecisionAD} & Answered & Jev 1.14 & Decider V1.2 & GPT-6 Luna Decisions & Clef & Mammoth 1 & Drex Max & Span-02 \\
\midrule
Defect named & 29 & 28 / 1 / 0 & 96.0 / 96.0 & 29 & 92.5 / 74.5 & 93.6 / 72.4 & 88.2 / 59.5 & 93.5 / 76.5 & 94.1 / 74.0 & 92.7 / 71.1 & 92.4 / 70.9 \\
Quality named & 17 & 17 / 0 / 0 & 96.0 / 96.0 & 11 & 93.3 / 81.6 & 93.0 / 73.2 & 84.1 / 59.4 & 91.0 / 76.5 & 94.0 / 74.6 & 91.2 / 71.1 & 91.6 / 76.1 \\
Negated defect & 24 & 22 / 1 / 1 & 96.0 / 4.0 & 23 & 93.5 / 69.2 & 93.7 / 51.2 & 87.2 / 58.8 & 93.3 / 74.7 & 93.5 / 66.4 & 91.7 / 49.6 & 92.1 / 65.0 \\
Negated quality & 8 & 8 / 0 / 0 & 96.0 / 96.0 & 7 & 92.5 / 88.8 & 89.8 / 82.1 & 81.7 / 58.5 & 90.8 / 87.0 & 92.9 / 84.4 & 90.5 / 84.6 & 90.5 / 78.9 \\
Double or implicit negation & 6 & 1 / 1 / 4 & 96.0 / 4.0 & 5 & 92.1 / 87.5 & 91.2 / 88.6 & 77.6 / 57.7 & 90.3 / 9.8 & 93.1 / 88.2 & 90.3 / 85.1 & 89.2 / 9.8 \\
Handling of the item & 20 & 10 / 10 / 0 & 96.0 / 96.0 & 14 & 92.5 / 51.4 & 91.2 / 68.2 & 83.4 / 50.2 & 90.9 / 80.9 & 93.7 / 72.6 & 90.7 / 63.6 & 89.7 / 44.0 \\
\midrule
All & 104 & 86 / 13 / 5 & 96.0 / 4.0 & 89 & 93.5 / 51.4 & 93.7 / 51.2 & 88.2 / 50.2 & 93.5 / 9.8 & 94.1 / 66.4 & 92.7 / 49.6 & 92.4 / 9.8 \\
\bottomrule
\end{tabular}
}
\end{table*}

\paragraph{Origin of the questions.}
The 104 questions come from three sets.
The first set is the 14 questions of \cref{tab:question_list}, whose pairs 2--7 are evaluated on images in \cref{tab:supp_wording_pairs}.
With 30 further questions, we revised the rules into those of \cref{sec:supp_rules}. A first version, which read a question as an absence question whenever it contained a negator or a quality cue, misread 10 of them.
The remaining 60 questions are held out. They were written after the rules were fixed, by a language model that received only a description of the task and not the rules, and we evaluated them once.
On the held-out questions, the rules read 44, decline 11, and misread 5, whereas the first version declines none and misreads 18.

\paragraph{Results by type.}
As the prompt pair is fixed, a question receives the image AUROC of \method (96.0 on MVTec AD) when its polarity is read correctly and the inverted value (4.0) otherwise.
The rules read 75 of the 78 questions that name or negate a defect or a quality, decline two, and misread one.
The errors concentrate in the last two types.
Of the six questions with a double or implicit negation, the rules misread four. These are 6 of the 104 questions, and Clef and Span-02 also fall to 9.8 on one of them.
Of the 20 questions about the handling of the item, the rules read 10 and decline the other 10, without a misread.

\paragraph{Decision models.}
The seven decision models received the 104 questions on MVTec AD.
\Cref{tab:qtypes} reports the 89 questions for which every model answered at least 95\% of the 1,725 images. On the other 15, GPT-6 Luna Decisions refused more images.
Negation alone rarely changes their ranking.
For each model, the median over the negated-defect questions is within 1.6\%p of that over the defect questions, and the maximum within 1.1\%p.
The lowest negated-defect result of every model is on ``Is the item rust-free?'', which names one kind of defect.
Of the 623 results of the 89 questions and seven models, only two are far below chance, and the next lowest is 44.0.
Both are for the double negation ``Would it be wrong to call this part faulty?'', with 9.8 for Clef and Span-02.
As in \cref{sec:pilot}, a negation shifts the probabilities of a decision model more than its ranking.

The image AUROC of the decision models falls when the question names one kind of defect, as in ``Is the item rust-free?''
We marked these questions in \cref{tab:qlist_a,tab:qlist_b} after seeing the results.
On the 19 of them with complete answers, the median of every model is 3.2--11.5\%p lower than on the 56 other questions outside the handling type.
The labels count any defect, so an item with a scratch is anomalous although it is rust-free.
A decision model answers the question as worded.
\method takes only the polarity and answers whether the item has any defect, as the labels do.
It does not distinguish kinds of defects.

Among these 56 questions, the image AUROC of a model still varies by at least 7.3\%p, and the seven models have five different best questions.
The models largely agree on which questions are hard, with a median pairwise Spearman correlation of 0.73 over the 89 questions, but not on the best one.
The best wording therefore has to be found per model with labeled images, whereas \method returns one ranking for every question it reads.

The questions about handling show the opposite strength.
A decision model answers them without a cue, with a maximum of 83.4--93.7 and a minimum of 44.0--80.9.
The lowest result is for ``Can this product be delivered to the customer?''
\method declines half of these questions and needs their polarity declared.

\paragraph{Embedding-based parser.}
An embedding-based parser embeds the question with the same SigLIP2 text encoder and reads the polarity from the closer side of the fixed prompt pair, where a question closer to $t_{\mathrm{n}}$ is read as an absence question.
A variant reads the polarity from the closer question of pair 1.
The text encoder does not separate the two polarities and labels almost every question as an absence question (\cref{tab:parser}).
With the prompt-pair anchors, the parser reads 2 of the 29 questions that name a defect and none of the 8 that negate a quality.
The two variants read 54 and 64 of the 104 questions, against 86 for the rules.
They also take 9.9\,ms per question against 0.04\,ms for the rules.

\paragraph{Language-model parser.}
A small language model can also read the polarity.
We prompt the instruction-tuned Qwen2.5 with 0.5B and 1.5B parameters~\cite{yang2024qwen25} and SmolLM2 with 1.7B parameters~\cite{allal2025smollm2}.
The prompt holds a fixed instruction and in-context examples that are not among the 104 questions.
Each model chooses among a presence question, an absence question, and declining.
Qwen2.5-1.5B, the best of them, reads 94 of the 104 questions but never declines, so it misreads 10, twice as many as the rules (\cref{tab:parser}).
Left uncorrected, a misread inverts every answer, whereas a declined question only has its polarity declared at setup.
With the order of the answer options swapped in the prompt, the same model misreads 22 questions.
The three models take 19--25\,ms per question against 0.04\,ms for the rules.
\begin{table*}[!htbp]
\centering
\caption{\textbf{Rule-based, embedding-based, and language-model parsers by question type.} Each cell lists the questions read correctly / declined / misread. The time is per question, on one A100 for the embedding-based and language-model parsers and on a CPU for the rules.}
\label{tab:parser}
\fitbox{\footnotesize
\setlength{\tabcolsep}{4pt}
\begin{tabular}{l ccccccc r}
\toprule
Parser & Defect (29) & Quality (17) & Neg.\ defect (24) & Neg.\ quality (8) & Double/implicit neg. (6) & Handling (20) & All (104) & ms/question \\
\midrule
Rules (ours) & 28 / 1 / 0 & 17 / 0 / 0 & 22 / 1 / 1 & 8 / 0 / 0 & 1 / 1 / 4 & 10 / 10 / 0 & 86 / 13 / 5 & 0.04 \\
Rules, first version & 29 / 0 / 0 & 12 / 0 / 5 & 21 / 0 / 3 & 1 / 0 / 7 & 0 / 0 / 6 & 13 / 0 / 7 & 76 / 0 / 28 & 0.04 \\
Embedding, prompt-pair anchors & 2 / 0 / 27 & 17 / 0 / 0 & 24 / 0 / 0 & 0 / 0 / 8 & 2 / 0 / 4 & 9 / 0 / 11 & 54 / 0 / 50 & 9.92 \\
Embedding, pair-1 anchors & 11 / 0 / 18 & 17 / 0 / 0 & 24 / 0 / 0 & 0 / 0 / 8 & 2 / 0 / 4 & 10 / 0 / 10 & 64 / 0 / 40 & 9.92 \\
\midrule
Qwen2.5-0.5B~\cite{yang2024qwen25} & 14 / 15 / 0 & 0 / 15 / 2 & 0 / 15 / 9 & 0 / 8 / 0 & 1 / 4 / 1 & 6 / 12 / 2 & 21 / 69 / 14 & 20.52 \\
Qwen2.5-1.5B~\cite{yang2024qwen25} & 29 / 0 / 0 & 16 / 0 / 1 & 20 / 0 / 4 & 8 / 0 / 0 & 4 / 0 / 2 & 17 / 0 / 3 & 94 / 0 / 10 & 24.55 \\
SmolLM2-1.7B~\cite{allal2025smollm2} & 29 / 0 / 0 & 0 / 10 / 7 & 0 / 9 / 15 & 7 / 1 / 0 & 2 / 3 / 1 & 11 / 0 / 9 & 49 / 23 / 32 & 18.69 \\
\bottomrule
\end{tabular}
}
\end{table*}

\paragraph{Use.}
The question of an inspection task is fixed when the system is set up and not chosen per image.
At setup, the user confirms the polarity that the parser reads and declares it for a declined question, like the answer type of a decision API.
A misread is thus corrected before any image is answered.
The answer statistic $u$ and the evidence map do not depend on the polarity, so declaring the other polarity needs no recomputation.
\begin{table*}[!htbp]
\centering
\caption{\textbf{The 104 questions (part 1).} Set R denotes the questions of \cref{tab:question_list}, D those used to develop the rules, and H the held-out questions. In the last column, a \checkmark\ marks a polarity read as intended, $\times$ a misread polarity, and -- a declined question. A $\dagger$ marks a question that names one kind of defect.}
\label{tab:qlist_a}
\scriptsize
\setlength{\tabcolsep}{3pt}
\begin{tabular}{@{}l l c l c@{}}
\toprule
Question & Type & Set & Intended & Read \\
\midrule
Does this industrial inspection image contain a visible defect or anomaly? & defect & R & presence & \checkmark \\
Is there any visible defect or anomaly anywhere in this industrial inspection image? & defect & R & presence & \checkmark \\
Is there any defect visible on this item? & defect & R & presence & \checkmark \\
Can you see anything wrong with the part in this photo? & defect & R & presence & \checkmark \\
Does this product show any signs of damage or flaws? & defect & R & presence & \checkmark \\
Is something abnormal about the object shown here? & defect & R & presence & \checkmark \\
Would an inspector find a fault anywhere on this piece? & defect & R & presence & \checkmark \\
Is anything wrong with this part? & defect & D & presence & \checkmark \\
Can you spot a scratch, crack, or dent on this object?$^{\dagger}$ & defect & D & presence & \checkmark \\
Is this unit faulty? & defect & D & presence & \checkmark \\
Does the surface show any damage? & defect & D & presence & \checkmark \\
Is there a problem with this product? & defect & D & presence & \checkmark \\
Is this item defective? & defect & D & presence & \checkmark \\
Are there irregularities on this part? & defect & D & presence & \checkmark \\
Has this piece been damaged in any way? & defect & D & presence & \checkmark \\
Is there a defect on this part? & defect & H & presence & \checkmark \\
Does this item show any damage? & defect & H & presence & \checkmark \\
Is the product defective? & defect & H & presence & \checkmark \\
Can you see an anomaly in this image? & defect & H & presence & \checkmark \\
Does anything look wrong with this piece? & defect & H & presence & \checkmark \\
Is there a scratch on the surface?$^{\dagger}$ & defect & H & presence & \checkmark \\
Does the component have a crack?$^{\dagger}$ & defect & H & presence & \checkmark \\
Is the unit contaminated with dirt or residue?$^{\dagger}$ & defect & H & presence & \checkmark \\
Are there any dents in this object?$^{\dagger}$ & defect & H & presence & \checkmark \\
Is the item bent or deformed?$^{\dagger}$ & defect & H & presence & \checkmark \\
Does the surface show rust or corrosion?$^{\dagger}$ & defect & H & presence & \checkmark \\
Is a piece missing from this component?$^{\dagger}$ & defect & H & presence & \checkmark \\
Can you spot a hole or puncture in the product?$^{\dagger}$ & defect & H & presence & \checkmark \\
Is the printed marking on this item smudged or misaligned?$^{\dagger}$ & defect & H & presence & -- \\
Is this item in perfect condition? & quality & R & absence & \checkmark \\
Is this part in good shape? & quality & D & absence & \checkmark \\
Does this item meet the quality standard? & quality & D & absence & \checkmark \\
Is the surface spotless?$^{\dagger}$ & quality & D & absence & \checkmark \\
Does this item look as it should? & quality & D & absence & \checkmark \\
Is everything fine with this part? & quality & D & absence & \checkmark \\
Is the product undamaged? & quality & D & absence & \checkmark \\
Does this part look new and intact? & quality & D & absence & \checkmark \\
Is this object flawless? & quality & D & absence & \checkmark \\
Is the part in normal condition? & quality & D & absence & \checkmark \\
Is this part in good condition? & quality & H & absence & \checkmark \\
Does the item look normal? & quality & H & absence & \checkmark \\
Is the product intact? & quality & H & absence & \checkmark \\
Is this piece acceptable? & quality & H & absence & \checkmark \\
Does the component look as it should? & quality & H & absence & \checkmark \\
Is the surface clean and smooth?$^{\dagger}$ & quality & H & absence & \checkmark \\
Is every component present and properly seated?$^{\dagger}$ & quality & H & absence & \checkmark \\
Is this industrial inspection image free of visible defects or anomalies? & neg.\ defect & R & absence & \checkmark \\
Are visible defects and anomalies absent throughout this industrial inspection image? & neg.\ defect & R & absence & \checkmark \\
Does the object look free of any flaws? & neg.\ defect & R & absence & \checkmark \\
Is the product shown here defect-free? & neg.\ defect & R & absence & \checkmark \\
Is this item without any scratches or cracks?$^{\dagger}$ & neg.\ defect & D & absence & \checkmark \\
Is there nothing wrong with this part? & neg.\ defect & D & absence & \checkmark \\
\bottomrule
\end{tabular}

\end{table*}
\begin{table*}[!htbp]
\centering
\caption{\textbf{The 104 questions (part 2).} Notation as in \cref{tab:qlist_a}.}
\label{tab:qlist_b}
\scriptsize
\setlength{\tabcolsep}{3pt}
\begin{tabular}{@{}l l c l c@{}}
\toprule
Question & Type & Set & Intended & Read \\
\midrule
Is this piece not defective? & neg.\ defect & D & absence & \checkmark \\
Is it free from any visible fault? & neg.\ defect & D & absence & \checkmark \\
Is the surface free of scratches?$^{\dagger}$ & neg.\ defect & H & absence & \checkmark \\
Is the component without cracks or chips?$^{\dagger}$ & neg.\ defect & H & absence & \checkmark \\
Does the unit have no dents?$^{\dagger}$ & neg.\ defect & H & absence & \checkmark \\
Is the object clear of stains and discoloration?$^{\dagger}$ & neg.\ defect & H & absence & $\times$ \\
Is the piece straight, with no bending or warping?$^{\dagger}$ & neg.\ defect & H & absence & -- \\
Is the item rust-free?$^{\dagger}$ & neg.\ defect & H & absence & \checkmark \\
Does the product show no signs of contamination?$^{\dagger}$ & neg.\ defect & H & absence & \checkmark \\
Is there no defect on this part? & neg.\ defect & H & absence & \checkmark \\
Is the item without any flaws? & neg.\ defect & H & absence & \checkmark \\
Is nothing wrong with this unit? & neg.\ defect & H & absence & \checkmark \\
Does this piece show no damage at all? & neg.\ defect & H & absence & \checkmark \\
Is the product free of any anomaly? & neg.\ defect & H & absence & \checkmark \\
Are there no visible problems with this object? & neg.\ defect & H & absence & \checkmark \\
Is this component not defective? & neg.\ defect & H & absence & \checkmark \\
Is it true that the item is not damaged? & neg.\ defect & H & absence & \checkmark \\
Is the surface not flawed in any way? & neg.\ defect & H & absence & \checkmark \\
Is something not right with this item? & neg.\ quality & D & presence & \checkmark \\
Is this part not acceptable? & neg.\ quality & D & presence & \checkmark \\
Is the product not in good condition? & neg.\ quality & D & presence & \checkmark \\
Does this object differ from a normal one? & neg.\ quality & D & presence & \checkmark \\
Is this part not in acceptable condition? & neg.\ quality & H & presence & \checkmark \\
Does this unit fail to meet the quality standard? & neg.\ quality & H & presence & \checkmark \\
Is the surface no longer intact? & neg.\ quality & H & presence & \checkmark \\
Would you say this product is not up to spec? & neg.\ quality & H & presence & \checkmark \\
Is the piece anything other than flawless? & double/implicit neg. & H & presence & $\times$ \\
Is the item not free of flaws? & double/implicit neg. & H & presence & \checkmark \\
Is it untrue that this component is defect-free? & double/implicit neg. & H & presence & $\times$ \\
Would calling this item perfect be an overstatement? & double/implicit neg. & H & presence & $\times$ \\
Would it be wrong to call this part faulty? & double/implicit neg. & H & absence & $\times$ \\
Is there nothing that would keep this unit from passing? & double/implicit neg. & H & absence & -- \\
Should this part be rejected? & handling & D & presence & \checkmark \\
Does this item need to be sent back for repair? & handling & D & presence & \checkmark \\
Would a quality inspector flag this piece? & handling & D & presence & \checkmark \\
Is this object something you could not approve? & handling & H & presence & -- \\
Should this part be sent back for rework? & handling & H & presence & \checkmark \\
Would you pull this item off the line? & handling & H & presence & -- \\
Does this unit belong in the reject bin? & handling & H & presence & \checkmark \\
Should this product be held back from shipping? & handling & H & presence & -- \\
Would a customer have grounds to return this item? & handling & H & presence & -- \\
Does this piece need to be flagged for the quality team? & handling & H & presence & \checkmark \\
Is this object headed for the scrap pile? & handling & H & presence & \checkmark \\
Would this part pass a visual quality check? & handling & R & absence & \checkmark \\
Can this piece be shipped as a good unit? & handling & R & absence & \checkmark \\
Can this product be delivered to the customer? & handling & D & absence & -- \\
Is this unit ready for sale? & handling & D & absence & -- \\
Would this item pass inspection? & handling & H & absence & \checkmark \\
Is this unit ready to ship? & handling & H & absence & -- \\
Could this product go out to a customer as is? & handling & H & absence & -- \\
Would you sign off on this piece? & handling & H & absence & -- \\
Can this part move on to the next stage of production? & handling & H & absence & -- \\
\bottomrule
\end{tabular}

\end{table*}

\FloatBarrier
\section{Full zero-shot results}
\label{sec:supp_full}
\Cref{tab:supp_image,tab:supp_pixel} report all image and pixel metrics of the methods in \cref{tab:industrial}, with the same notation. The medical results are in \cref{sec:supp_medical}.
The decision models return no map and are listed in \cref{tab:supp_image} only.
On average over the five datasets, \method is first in all three image metrics and in all four pixel metrics.
It leads 12 of the 15 image cells and is within 1.6\%p of the best in the other three.
Among the training-free baselines, only SAA+ in F1-max on VisA is higher in a pixel cell.
In F1-max, \method ranks first on all five datasets, with a mean of 87.2 against 84.9 for both VisualAD and Mammoth~1, the best ZSAD baseline and the best decision model.
\begin{table}[!htbp]
\centering
\caption{\textbf{All image metrics on industrial datasets} (AUROC, AP, F1-max, category macro, \%).}
\label{tab:supp_image}
\fitbox{\scriptsize
\setlength{\tabcolsep}{2.2pt}
\begin{tabular}{l l c ccc ccc ccc ccc ccc}
\toprule
\multicolumn{1}{l}{\multirow{2}{*}[-0.5\dimexpr\aboverulesep+\belowrulesep\relax]{Method}} & \multirow{2}{*}[-0.5\dimexpr\aboverulesep+\belowrulesep\relax]{Venue} & \multirow{2}{*}[-0.5\dimexpr\aboverulesep+\belowrulesep\relax]{Train} & \multicolumn{3}{c}{MVTec AD} & \multicolumn{3}{c}{VisA} & \multicolumn{3}{c}{MPDD} & \multicolumn{3}{c}{BTAD} & \multicolumn{3}{c}{Real-IAD} \\
\cmidrule(lr){4-6} \cmidrule(lr){7-9} \cmidrule(lr){10-12} \cmidrule(lr){13-15} \cmidrule(lr){16-18}
 &  &  & AUC & AP & F1 & AUC & AP & F1 & AUC & AP & F1 & AUC & AP & F1 & AUC & AP & F1 \\
\midrule
WinCLIP~\cite{jeong2023winclip} & CVPR'23 & -- & 90.4 & 95.6 & 92.7 & 75.5 & 78.7 & 78.2 & 61.5 & 69.2 & 77.5 & 69.6 & 70.9 & 68.1 & 67.1 & 63.0 & 65.3 \\
AnoVL~\cite{deng2023anovl} & arXiv'23 & -- & 92.5 & 96.8 & 93.2 & 79.2 & 81.8 & 79.6 & 68.6 & 71.8 & 75.9 & 80.7 & 72.1 & 72.0 & 67.0 & 63.2 & 65.2 \\
FADE~\cite{li2024fade} & BMVC'24 & -- & 90.1 & 95.5 & 92.0 & 74.7 & 78.5 & 78.3 & 62.8 & 70.3 & 78.0 & 76.1 & 72.5 & 69.9 & 67.7 & 63.8 & 65.6 \\
SAA+~\cite{cao2023saa} & TCYB'25 & -- & 62.4 & 81.1 & 87.0 & 71.0 & 77.2 & 76.0 & 56.3 & 67.5 & 74.9 & 57.0 & 76.4 & 86.1 & 56.1 & 52.9 & 63.9 \\
\midrule
APRIL-GAN~\cite{chen2023aprilgan} & arXiv'23 & AD & 86.0 & 93.4 & 90.4 & 77.5 & 80.8 & 78.6 & 76.5 & 82.6 & 80.6 & 75.5 & 70.7 & 70.4 & 68.4 & 64.5 & 65.3 \\
AnomalyCLIP~\cite{zhou2024anomalyclip} & ICLR'24 & AD & 91.6 & 96.3 & 92.8 & 82.0 & 85.3 & 80.4 & 77.5 & 82.5 & 80.4 & 88.8 & 90.2 & 86.3 & 68.9 & 63.5 & 65.8 \\
VCP-CLIP~\cite{qu2024vcpclip} & ECCV'24 & AD & 89.8 & 95.5 & 91.8 & 83.8 & 87.6 & 81.5 & 69.3 & 72.9 & 78.2 & 89.7 & 94.0 & 90.7 & 72.4 & 69.9 & 68.0 \\
Bayes-PFL~\cite{qu2025bayespfl} & CVPR'25 & AD & 92.2 & 96.2 & 93.1 & 86.9 & \underline{89.3} & 83.6 & 76.9 & 78.7 & 81.3 & 91.2 & 90.6 & 87.5 & 72.6 & 68.9 & 67.4 \\
AA-CLIP~\cite{ma2025aaclip} & CVPR'25 & AD & 89.8 & 95.1 & 90.7 & 78.1 & 81.8 & 78.3 & 74.3 & 79.2 & 81.3 & \underline{95.4} & \underline{98.0} & 94.8 & 75.0 & 73.2 & 68.8 \\
FAPrompt~\cite{zhu2025faprompt} & ICCV'25 & AD & 91.8 & 95.6 & 92.3 & 85.0 & 86.7 & 82.6 & 79.8 & \underline{83.0} & 82.2 & 93.0 & 96.2 & 93.0 & 77.4 & 74.8 & 70.4 \\
MoECLIP~\cite{park2026moeclip} & CVPR'26 & AD & \underline{93.0} & 97.0 & 93.3 & 82.8 & 84.8 & 81.0 & 68.3 & 73.9 & 77.6 & 90.3 & 95.2 & 92.7 & 79.0 & 76.8 & 71.8 \\
VisualAD~\cite{hou2026visualad} & CVPR'26 & AD & 92.2 & 96.7 & 93.2 & 84.6 & 87.6 & 82.4 & 71.5 & 79.5 & 81.6 & 95.0 & \underline{98.0} & \underline{95.9} & 78.5 & 76.2 & 71.3 \\
\midrule
Jev 1.14 & API & -- & 92.5 & 96.2 & 93.0 & 85.3 & 87.5 & 83.1 & 62.8 & 71.9 & 76.5 & 92.3 & 88.8 & 85.7 & \underline{81.1} & 79.0 & \underline{73.0} \\
Decider V1.2 & API & -- & 92.1 & 96.9 & 93.9 & \underline{87.5} & \textbf{90.8} & 84.1 & 79.9 & \underline{83.0} & 81.7 & 91.5 & 91.7 & 87.3 & 80.9 & \underline{79.7} & 72.6 \\
GPT-6 Luna Decisions & API & -- & 87.2 & 94.6 & 91.5 & 80.2 & 83.2 & 79.6 & 76.2 & 77.9 & 78.7 & 79.0 & 71.5 & 71.7 & 76.6 & 74.4 & 69.5 \\
Clef & API & -- & 91.5 & 96.4 & 93.4 & 78.9 & 84.1 & 79.1 & 77.3 & 82.3 & 81.4 & 81.0 & 83.6 & 82.7 & 76.7 & 75.4 & 69.3 \\
Mammoth 1 & API & -- & \underline{93.0} & \underline{97.2} & \underline{94.9} & \textbf{88.2} & \textbf{90.8} & \underline{84.8} & \underline{80.4} & 82.0 & \underline{83.0} & 92.7 & 95.3 & 89.0 & 81.0 & \textbf{79.8} & 72.8 \\
Drex Max & API & -- & 90.7 & 96.3 & 92.7 & 85.6 & 88.7 & 82.3 & 73.6 & 78.8 & 78.8 & 89.9 & 90.0 & 87.0 & 79.9 & 78.3 & 71.5 \\
Span-02 & API & -- & 90.4 & 95.5 & 91.5 & 82.5 & 85.9 & 80.5 & 77.7 & 82.3 & 81.6 & 88.3 & 88.4 & 86.5 & 78.0 & 77.0 & 70.6 \\
\midrule
\textbf{DecisionAD} & -- & -- & \textbf{96.0} & \textbf{97.9} & \textbf{95.0} & \underline{87.5} & 89.2 & \textbf{86.1} & \textbf{80.5} & \textbf{83.6} & \textbf{83.5} & \textbf{96.0} & \textbf{99.1} & \textbf{96.7} & \textbf{81.3} & 79.4 & \textbf{74.8} \\
\bottomrule
\end{tabular}
}
\end{table}
\begin{table}[!htbp]
\centering
\caption{\textbf{All pixel metrics on industrial datasets} (AUROC, AP, AUPRO@0.30, F1-max, category macro, \%).}
\label{tab:supp_pixel}
\fitbox{\scriptsize
\setlength{\tabcolsep}{1.8pt}
\begin{tabular}{l l c cccc cccc cccc cccc cccc}
\toprule
\multicolumn{1}{l}{\multirow{2}{*}[-0.5\dimexpr\aboverulesep+\belowrulesep\relax]{Method}} & \multirow{2}{*}[-0.5\dimexpr\aboverulesep+\belowrulesep\relax]{Venue} & \multirow{2}{*}[-0.5\dimexpr\aboverulesep+\belowrulesep\relax]{Train} & \multicolumn{4}{c}{MVTec AD} & \multicolumn{4}{c}{VisA} & \multicolumn{4}{c}{MPDD} & \multicolumn{4}{c}{BTAD} & \multicolumn{4}{c}{Real-IAD} \\
\cmidrule(lr){4-7} \cmidrule(lr){8-11} \cmidrule(lr){12-15} \cmidrule(lr){16-19} \cmidrule(lr){20-23}
 &  &  & AUC & AP & PRO & F1 & AUC & AP & PRO & F1 & AUC & AP & PRO & F1 & AUC & AP & PRO & F1 & AUC & AP & PRO & F1 \\
\midrule
WinCLIP~\cite{jeong2023winclip} & CVPR'23 & -- & 82.3 & 18.2 & 62.2 & 24.8 & 73.0 & 5.5 & 51.3 & 9.1 & 71.5 & 14.1 & 41.9 & 15.4 & 72.6 & 12.9 & 26.6 & 18.5 & 84.2 & 3.4 & 55.6 & 7.9 \\
AnoVL~\cite{deng2023anovl} & arXiv'23 & -- & 86.5 & 24.0 & 70.9 & 30.1 & 85.1 & 9.3 & 60.2 & 14.7 & 60.0 & 13.8 & 33.0 & 15.2 & 75.2 & 18.3 & 41.8 & 23.4 & 80.8 & 4.3 & 48.6 & 9.2 \\
FADE~\cite{li2024fade} & BMVC'24 & -- & 89.2 & 34.3 & 84.2 & 39.8 & 91.0 & 10.1 & 77.6 & 16.9 & 95.2 & 19.1 & 85.7 & 24.7 & 96.4 & 40.8 & 76.7 & 47.0 & 93.0 & 9.1 & 82.5 & 17.0 \\
SAA+~\cite{cao2023saa} & TCYB'25 & -- & 73.9 & 29.3 & 55.0 & 37.8 & 76.4 & 22.1 & 58.3 & 33.2 & 81.9 & 8.4 & 66.5 & 19.4 & 59.3 & 5.3 & 35.9 & 12.4 & 75.6 & 3.7 & 48.4 & 7.5 \\
\midrule
APRIL-GAN~\cite{chen2023aprilgan} & arXiv'23 & AD & 86.7 & 39.8 & 84.6 & 43.3 & 94.2 & 25.8 & 86.9 & 32.3 & 94.3 & 26.6 & 84.7 & 31.3 & 89.3 & 36.7 & 69.0 & 40.8 & 95.2 & 30.2 & 86.6 & 35.8 \\
AnomalyCLIP~\cite{zhou2024anomalyclip} & ICLR'24 & AD & 91.1 & 34.5 & 85.3 & 39.1 & 95.5 & 21.3 & 86.6 & 28.3 & 96.5 & 28.9 & 89.0 & 34.2 & 94.3 & 45.6 & 74.4 & 49.8 & 95.0 & 28.3 & 85.0 & 35.6 \\
VCP-CLIP~\cite{qu2024vcpclip} & ECCV'24 & AD & \textbf{92.1} & \textbf{49.1} & 87.7 & \textbf{49.3} & \underline{95.8} & \textbf{29.9} & 90.0 & \textbf{35.3} & 96.8 & 27.8 & 90.0 & 31.1 & 93.8 & 45.3 & 75.0 & 48.8 & 96.4 & \underline{31.2} & 87.7 & 35.4 \\
Bayes-PFL~\cite{qu2025bayespfl} & CVPR'25 & AD & \underline{91.9} & \underline{48.7} & \underline{88.1} & \textbf{49.3} & \textbf{95.9} & \underline{29.3} & 88.8 & 34.5 & \textbf{97.1} & 29.9 & \underline{90.8} & 32.9 & 92.2 & 40.3 & 77.8 & 43.9 & 96.0 & 25.9 & 85.8 & 31.1 \\
AA-CLIP~\cite{ma2025aaclip} & CVPR'25 & AD & 91.3 & 45.4 & 87.6 & 46.8 & 94.7 & 24.3 & 81.8 & 29.7 & 96.3 & 26.0 & 87.0 & 29.0 & \textbf{97.1} & \textbf{54.0} & \underline{79.7} & \underline{55.2} & 95.0 & 29.1 & 81.2 & 34.8 \\
FAPrompt~\cite{zhu2025faprompt} & ICCV'25 & AD & 89.4 & 36.9 & 85.5 & 40.8 & \underline{95.8} & 22.6 & 87.9 & 28.8 & 96.3 & 29.1 & 88.3 & 34.7 & 96.4 & 49.1 & 78.5 & 52.2 & 94.7 & 27.9 & 84.8 & 34.7 \\
MoECLIP~\cite{park2026moeclip} & CVPR'26 & AD & 91.5 & 45.3 & 86.2 & 46.0 & 94.8 & 23.7 & 83.9 & 29.8 & 96.8 & 24.7 & 87.8 & 27.7 & 96.4 & 47.2 & 76.7 & 52.5 & 94.5 & 23.8 & 79.7 & 29.9 \\
VisualAD~\cite{hou2026visualad} & CVPR'26 & AD & 90.8 & 41.2 & 87.9 & 43.6 & 95.7 & 28.3 & \underline{90.2} & \underline{34.6} & \textbf{97.1} & \textbf{35.4} & \textbf{92.5} & \underline{40.7} & 91.1 & 43.4 & 79.2 & 50.2 & \underline{96.6} & \textbf{35.0} & \underline{91.8} & \textbf{40.2} \\
\midrule
\textbf{DecisionAD} & -- & -- & 91.6 & 45.8 & \textbf{90.4} & \underline{47.7} & 95.1 & 23.9 & \textbf{90.5} & 29.1 & \underline{97.0} & \underline{32.5} & \underline{90.8} & \textbf{40.8} & \underline{96.8} & \underline{53.8} & \textbf{85.6} & \textbf{57.6} & \textbf{96.7} & \underline{31.2} & \textbf{92.8} & \underline{36.1} \\
\bottomrule
\end{tabular}
}
\end{table}

Over 1,000 paired bootstrap replicates, stratified by category and label, the 95\% interval of the margin of \method in mean image AUROC excludes zero for all 12 ZSAD baselines.
It is 1.6--4.1\%p for the closest one, FAPrompt.
Among the seven decision models, the interval excludes zero for six, with 0.6--3.2\%p for Decider~V1.2, and is $-$0.1 to 2.4\%p for Mammoth~1.
\method thus ranks images on par with the strongest decision model and differs from it in the consistency, the map, and the throughput.

\FloatBarrier
\section{Efficiency}
\label{sec:supp_efficiency}
This section reports the size, the computation, the memory, and the speed of the methods, the timing protocol, and the cost of each variant of \method.

\subsection{Model size, computation, memory, and speed}
\method trains no parameters, as all of its 1,200M parameters belong to frozen pretrained modules.
It runs at 25.1\,FPS, faster than 11 of the 12 baselines (\cref{tab:efficiency}).
Its peak GPU memory is 5.7\,GB, which includes the backward pass of answer attribution.
The answer score alone takes 15.8\,ms per image, and the evidence map raises this to 38.7\,ms (\cref{tab:supp_cost}).
It is nevertheless faster than the decision models because it runs one frozen encoder with its heads, 1.2B parameters in total, on site.
The decision models with a published size have 27B parameters and read 346--434 input tokens per request.
This is approximately 20 TFLOPs per image for the language model alone, against 2.5 TFLOPs for \method.
Every request also crosses the network to a shared service. The median latency of a request is 0.60--1.63\,s on MVTec AD.
Concurrent requests raise the throughput of an API but do not lower the latency of a request, and the images still leave the site.
An open vision-language model running on site is also slower.
On one RTX 4090, Qwen2.5-VL-7B~\cite{bai2025qwen25vl} takes 141.1\,ms per image up to its first answer token, including image loading, and 16.4\,GB of memory, against 39.1\,ms and 5.7\,GB for \method (\cref{sec:supp_timing}).
\begin{table}[!htbp]
\centering
\caption{\textbf{Model size, computation, memory, and speed} (MVTec AD cable, one A100, batch size 1).
Total counts every module run at inference, and Trained counts the parameters fitted by the method. Train as in \cref{tab:industrial}. Mem.\ is the peak allocated GPU memory. Lower is better for Trained, GFLOPs, and Mem., and higher for FPS.}
\label{tab:efficiency}
\footnotesize
\setlength{\tabcolsep}{3pt}
\begin{tabular}{l c r r r r r}
\toprule
\multicolumn{1}{l}{\multirow{2}{*}[-0.5\dimexpr\aboverulesep+\belowrulesep\relax]{Method}} & \multirow{2}{*}[-0.5\dimexpr\aboverulesep+\belowrulesep\relax]{Train} & \multicolumn{2}{c}{Params (M)} & \multirow{2}{*}[-0.5\dimexpr\aboverulesep+\belowrulesep\relax]{GFLOPs} & \multirow{2}{*}[-0.5\dimexpr\aboverulesep+\belowrulesep\relax]{Mem.\ (GB)} & \multirow{2}{*}[-0.5\dimexpr\aboverulesep+\belowrulesep\relax]{FPS} \\
\cmidrule(lr){3-4}
 &  & Total & Trained &  &  &  \\
\midrule
WinCLIP~\cite{jeong2023winclip} & -- & 193 & \textbf{0.00} & \textbf{674} & \textbf{0.9} & 15.1 \\
AnoVL~\cite{deng2023anovl} & -- & 203 & \textbf{0.00} & 1418 & \underline{1.2} & 5.4 \\
FADE~\cite{li2024fade} & -- & 198 & \textbf{0.00} & 2822 & 6.2 & 2.4 \\
SAA+~\cite{cao2023saa} & -- & 837 & \textbf{0.00} & 8572 & 8.0 & 0.8 \\
\midrule
APRIL-GAN~\cite{chen2023aprilgan} & AD & 399 & \underline{3.15} & \underline{1022} & 2.4 & \textbf{31.1} \\
AnomalyCLIP~\cite{zhou2024anomalyclip} & AD & 416 & 5.66 & 1216 & 2.0 & 8.3 \\
VCP-CLIP~\cite{qu2024vcpclip} & AD & 404 & 10.76 & 1072 & 2.7 & 14.2 \\
Bayes-PFL~\cite{qu2025bayespfl} & AD & 423 & 36.82 & 1837 & 2.3 & 2.5 \\
AA-CLIP~\cite{ma2025aaclip} & AD & 406 & 12.58 & 1042 & 2.1 & 10.4 \\
FAPrompt~\cite{zhu2025faprompt} & AD & 420 & 10.21 & 1509 & 2.0 & 6.1 \\
MoECLIP~\cite{park2026moeclip} & AD & 398 & 4.81 & 1042 & 2.0 & 7.7 \\
VisualAD~\cite{hou2026visualad} & AD & 310 & 31.52 & 1085 & 2.0 & 11.8 \\
\midrule
\textbf{DecisionAD} & -- & 1200 & \textbf{0.00} & 2522 & 5.7 & \underline{25.1} \\
\bottomrule
\end{tabular}

\end{table}

\subsection{Timing protocol}
\label{sec:supp_timing}
All methods are timed in one session with the same protocol, on one NVIDIA A100 80GB PCIe with batch size 1 on the MVTec AD category cable.
Each method keeps the precision of its official code.
The time per image is estimated as $(T_{110}-T_{10})/100$, where $T_m$ is the wall time of a run over $m$ images, and we take the median of three repetitions.
The subtraction removes fixed costs such as model loading.
The timing script also times the image loop of the last repetition directly and reports the loop time when it lies more than 10\% from all three estimates.
For \method, the estimate is 23.9\,ms and the loop time 39.9\,ms per image including image loading. We report the slower loop time (25.1\,FPS).
On one RTX 4090 with 24\,GB, the same protocol yields 39.1\,ms per image (25.6\,FPS), so the speed holds on a single consumer GPU.
Qwen2.5-VL-7B is timed with the same image loop on this GPU.
With the images already in memory, the backward pass adds 9.0\,ms and the rarity statistics 10.8\,ms per image (\cref{tab:supp_cost}).
As the text embeddings are computed once, the number of prompt templates does not change the time per image.
\begin{table}[!htbp]
\centering
\caption{\textbf{Cost of each variant} (one A100, batch size 1, images in memory, our setting shaded).
The 25.1\,FPS of the main paper refers to the timed loop including image loading.}
\label{tab:supp_cost}
\footnotesize
\setlength{\tabcolsep}{4pt}
\begin{tabular}{l r r r}
\toprule
\multicolumn{1}{l}{Variant} & ms/image & GFLOPs & FPS \\
\midrule
\rowcolor{oursrow}\textbf{DecisionAD} & 38.7 & 2522 & 25.9 \\
\midrule
without answer attribution (no backward pass) & 29.7 & 1691 & 33.7 \\
without self-rarity (no rarity statistics) & 27.9 & 2345 & 35.8 \\
without refinement (no SAM3 head) & 38.2 & 2496 & 26.2 \\
answer score alone (no map) & 15.8 & 1488 & 63.4 \\
\midrule
attribution layer 6 & 44.8 & 3033 & 22.3 \\
attribution layer 26 & 31.5 & 1755 & 31.7 \\
rarity layers 1--27 & 68.4 & 2876 & 14.6 \\
rarity layer 14 only & 29.8 & 2364 & 33.5 \\
\bottomrule
\end{tabular}

\end{table}

\FloatBarrier
\section{Ablations}
\label{sec:supp_ablation}
This section varies the layers, the encoder, and the remaining settings of \method, and examines the contribution of the encoder and that of the method.
Ablations use MVTec AD and VisA, and \cref{sec:supp_encoder} uses the five industrial datasets. In the tables on MVTec AD and VisA, the best of each group is in bold and the second underlined, and our setting is shaded.
Our setting is not the best in every column. A single state, map weights of 1:2, and a 768\,px input each yield a higher image AUROC on VisA (89.0, 88.1, and 89.2 against 87.5).

\subsection{Layers and encoders}
\label{sec:supp_abl_layers}
On both datasets, earlier attribution layers yield higher pixel AP. On VisA, later layers yield higher image AUROC. Our layer 14, at half the depth, lies in between (\cref{tab:supp_layers}).
Rarity layers 10--18, at one to two thirds of the depth, perform similarly to layers 1--9, and layers 19--27 perform worse.
A single layer, layer 14, remains within 0.5\%p of the nine layers in every metric and takes 29.8\,ms instead of 38.7\,ms per image (\cref{tab:supp_cost}).
Of the two statistics, attention locality alone is higher in image AUROC and pixel AP on MVTec AD, and feature rarity alone in image AUROC on VisA. Their combination stays within 1.0\%p of the better one in every metric and yields a higher AUPRO than either on both datasets.
Applying the same relative-depth rule to C-RADIOv4-H yields similar results on MVTec AD and lower results on VisA.
CLIP ViT-L/14@336 is lower in all metrics, and DINOv2-L, which has no text space, builds the map from $R$ alone.
With $R$ alone, DINOv2-L is higher in pixel AP on MVTec AD (47.4 \vs 45.8) and lower in image AUROC (89.8 \vs 96.0).
\Cref{sec:supp_encoder} examines the contribution of the encoder and that of the method.
\begin{table}[!htbp]
\centering
\caption{\textbf{Layers and encoders} (MVTec AD, VisA, \%, our setting shaded).}
\label{tab:supp_layers}
\fitbox{\scriptsize
\setlength{\tabcolsep}{2.2pt}
\begin{tabular}{l cccccc}
\toprule
\multicolumn{1}{l}{\multirow{3}{*}[-1\dimexpr\aboverulesep+\belowrulesep\relax]{Setting}} & \multicolumn{3}{c}{MVTec AD} & \multicolumn{3}{c}{VisA} \\
\cmidrule(lr){2-4} \cmidrule(lr){5-7}
 & Image & \multicolumn{2}{c}{Pixel} & Image & \multicolumn{2}{c}{Pixel} \\
\cmidrule(lr){2-2} \cmidrule(lr){3-4} \cmidrule(lr){5-5} \cmidrule(lr){6-7}
 & AUC & AP & PRO & AUC & AP & PRO \\
\midrule
\multicolumn{7}{l}{\textit{Answer-attribution layer $\ell_A$ (of 27)}} \\
layer 6 & 95.4 & \textbf{46.1} & \underline{90.2} & 86.7 & \textbf{24.5} & \textbf{91.3} \\
layer 10 & 95.7 & \underline{45.9} & \textbf{90.4} & 86.9 & \underline{24.2} & \underline{91.0} \\
\rowcolor{oursrow}layer 14 & \textbf{96.0} & 45.8 & \textbf{90.4} & 87.5 & 23.9 & 90.5 \\
layer 18 & \textbf{96.0} & 42.9 & 89.9 & 87.3 & 20.0 & 89.0 \\
layer 22 & \underline{95.9} & 42.0 & 89.7 & \underline{87.8} & 18.7 & 88.7 \\
layer 26 & 95.4 & 40.5 & 87.5 & \textbf{88.4} & 18.0 & 86.4 \\
\midrule
\multicolumn{7}{l}{\textit{Self-rarity layers $\mathcal{L}$}} \\
1--9 & 95.9 & \textbf{46.9} & 89.4 & 87.3 & \textbf{24.5} & \textbf{90.6} \\
\rowcolor{oursrow}10--18 & \underline{96.0} & 45.8 & \textbf{90.4} & \underline{87.5} & \underline{23.9} & \underline{90.5} \\
19--27 & 95.2 & 44.1 & 89.3 & 87.0 & 21.9 & 89.2 \\
1--27 & 95.9 & 46.2 & 90.0 & 87.4 & 23.8 & 90.3 \\
14 only & \underline{96.0} & 45.7 & \underline{90.3} & 87.0 & 23.8 & \underline{90.5} \\
attention locality only & \textbf{96.4} & \underline{46.8} & 89.7 & 86.8 & 23.5 & 90.2 \\
feature rarity only & 95.7 & 41.1 & 89.0 & \textbf{87.9} & 22.2 & 89.3 \\
\midrule
\multicolumn{7}{l}{\textit{Encoder}} \\
\rowcolor{oursrow}C-RADIOv4-SO400M (27 layers) & \underline{96.0} & 45.8 & \textbf{90.4} & \textbf{87.5} & \textbf{23.9} & \textbf{90.5} \\
C-RADIOv4-H (32 layers) & \textbf{96.2} & \underline{46.9} & \underline{90.1} & \underline{85.2} & 22.2 & 90.1 \\
CLIP ViT-L/14@336 & 91.4 & 37.3 & 84.2 & 81.9 & 17.6 & 86.4 \\
DINOv2-L (map $=R$ only) & 89.8 & \textbf{47.4} & 87.0 & 83.4 & \underline{23.6} & \underline{90.3} \\
\bottomrule
\end{tabular}
}
\end{table}

\subsection{Encoder and method}
\label{sec:supp_encoder}
The encoder accounts for part of the margins in \cref{tab:industrial}.
We separate the encoder from the method on the five industrial datasets, first with the encoder fixed (\cref{tab:supp_encoder,tab:supp_attr}) and then with the method fixed (\cref{tab:supp_clip}).
Last, we train four auxiliary-trained methods on the encoder of \method (\cref{tab:supp_ports}).
\begin{table*}[!htbp]
\centering
\caption{\textbf{Same encoder.} Controls on C-RADIOv4-SO400M on the five industrial datasets (\%, \method shaded).}
\label{tab:supp_encoder}
\fitbox{\scriptsize
\setlength{\tabcolsep}{2.2pt}
\begin{tabular}{l cccccccccccccccccc}
\toprule
\multicolumn{1}{l}{\multirow{3}{*}[-1\dimexpr\aboverulesep+\belowrulesep\relax]{Method}} & \multicolumn{3}{c}{MVTec AD} & \multicolumn{3}{c}{VisA} & \multicolumn{3}{c}{MPDD} & \multicolumn{3}{c}{BTAD} & \multicolumn{3}{c}{Real-IAD} & \multicolumn{3}{c}{Mean} \\
\cmidrule(lr){2-4} \cmidrule(lr){5-7} \cmidrule(lr){8-10} \cmidrule(lr){11-13} \cmidrule(lr){14-16} \cmidrule(lr){17-19}
 & Image & \multicolumn{2}{c}{Pixel} & Image & \multicolumn{2}{c}{Pixel} & Image & \multicolumn{2}{c}{Pixel} & Image & \multicolumn{2}{c}{Pixel} & Image & \multicolumn{2}{c}{Pixel} & Image & \multicolumn{2}{c}{Pixel} \\
\cmidrule(lr){2-2} \cmidrule(lr){3-4} \cmidrule(lr){5-5} \cmidrule(lr){6-7} \cmidrule(lr){8-8} \cmidrule(lr){9-10} \cmidrule(lr){11-11} \cmidrule(lr){12-13} \cmidrule(lr){14-14} \cmidrule(lr){15-16} \cmidrule(lr){17-17} \cmidrule(lr){18-19}
 & AUC & AP & PRO & AUC & AP & PRO & AUC & AP & PRO & AUC & AP & PRO & AUC & AP & PRO & AUC & AP & PRO \\
\midrule
Patch--text similarity & 51.9 & 12.4 & 50.2 & 54.5 & 2.7 & 46.6 & 50.6 & 7.9 & 50.8 & 60.4 & 11.3 & 32.9 & 56.0 & 5.4 & 65.7 & 54.7 & 7.9 & 49.2 \\
Answer score alone & 95.6 & -- & -- & 86.0 & -- & -- & 81.2 & -- & -- & 93.4 & -- & -- & 78.1 & -- & -- & 86.8 & -- & -- \\
\rowcolor{oursrow}\textbf{DecisionAD} & 96.0 & 45.8 & 90.4 & 87.5 & 23.9 & 90.5 & 80.5 & 32.5 & 90.8 & 96.0 & 53.8 & 85.6 & 81.3 & 31.2 & 92.8 & 88.2 & 37.5 & 90.0 \\
\bottomrule
\end{tabular}
}
\end{table*}

\paragraph{Same encoder.}
Existing ZSAD methods compare each patch token with the text of a prompt pair.
We apply this to C-RADIOv4-SO400M with the patch tokens of its SigLIP2 head and the prompt pair of \method.
The map is the difference of the two patch--text similarities, and the image score is its maximum.
This control reaches a mean image AUROC of 54.7 and a mean AUPRO of 49.2, against 88.2 and 90.0 for \method (\cref{tab:supp_encoder}).
Without training, the patch tokens of this encoder are thus poorly aligned with the text, and the map of \method comes from answer attribution and self-rarity.
The answer score alone reaches a mean image AUROC of 86.8 and yields no map.
The map adds 1.4\%p on average, with 3.2\%p on Real-IAD and 2.6\%p on BTAD, and lowers the image AUROC on MPDD by 0.7\%p.
Over 1,000 paired bootstrap replicates, stratified by category and label, the 95\% interval of the mean gain is 0.7--2.2\%p.
The interval of a single dataset excludes zero on VisA, BTAD, and Real-IAD and includes it on MVTec AD and MPDD, so the decrease on MPDD is within the sampling variation.

\begin{table*}[!htbp]
\centering
\caption{\textbf{Other attributions} in place of answer attribution on C-RADIOv4-SO400M (\%, \method shaded). The first group keeps all other steps, and the second group also removes self-rarity. The best of each group is in bold.}
\label{tab:supp_attr}
\fitbox{\scriptsize
\setlength{\tabcolsep}{2.2pt}
\begin{tabular}{l cccccccccccccccccc}
\toprule
\multicolumn{1}{l}{\multirow{3}{*}[-1\dimexpr\aboverulesep+\belowrulesep\relax]{Method}} & \multicolumn{3}{c}{MVTec AD} & \multicolumn{3}{c}{VisA} & \multicolumn{3}{c}{MPDD} & \multicolumn{3}{c}{BTAD} & \multicolumn{3}{c}{Real-IAD} & \multicolumn{3}{c}{Mean} \\
\cmidrule(lr){2-4} \cmidrule(lr){5-7} \cmidrule(lr){8-10} \cmidrule(lr){11-13} \cmidrule(lr){14-16} \cmidrule(lr){17-19}
 & Image & \multicolumn{2}{c}{Pixel} & Image & \multicolumn{2}{c}{Pixel} & Image & \multicolumn{2}{c}{Pixel} & Image & \multicolumn{2}{c}{Pixel} & Image & \multicolumn{2}{c}{Pixel} & Image & \multicolumn{2}{c}{Pixel} \\
\cmidrule(lr){2-2} \cmidrule(lr){3-4} \cmidrule(lr){5-5} \cmidrule(lr){6-7} \cmidrule(lr){8-8} \cmidrule(lr){9-10} \cmidrule(lr){11-11} \cmidrule(lr){12-13} \cmidrule(lr){14-14} \cmidrule(lr){15-16} \cmidrule(lr){17-17} \cmidrule(lr){18-19}
 & AUC & AP & PRO & AUC & AP & PRO & AUC & AP & PRO & AUC & AP & PRO & AUC & AP & PRO & AUC & AP & PRO \\
\midrule
\multicolumn{19}{l}{\textit{With self-rarity}} \\
Gradient $\times$ feature & 94.5 & 30.2 & 76.1 & 87.2 & 13.0 & 78.8 & 73.8 & 15.8 & 72.7 & 92.9 & 35.2 & 77.0 & 79.1 & 15.6 & 82.9 & 85.5 & 21.9 & 77.5 \\
$|$Gradient $\times$ feature$|$ & 95.6 & 35.3 & 88.3 & 84.8 & 14.9 & 88.0 & 77.0 & 23.2 & 89.5 & 93.1 & 42.4 & 82.2 & 79.5 & 17.4 & 91.3 & 86.0 & 26.6 & 87.9 \\
Grad-CAM & 95.6 & 36.2 & 85.8 & 87.0 & 14.3 & 85.4 & 79.7 & 23.5 & 87.4 & 95.2 & 41.7 & 83.7 & 77.8 & 17.2 & 91.5 & 87.0 & 26.6 & 86.7 \\
Gradient norm & 95.8 & 38.2 & 89.5 & 86.1 & 18.2 & 89.5 & 77.3 & 29.2 & \textbf{90.8} & 94.9 & 51.8 & 84.6 & 80.6 & 23.2 & 92.6 & 86.9 & 32.1 & 89.4 \\
\rowcolor{oursrow}\textbf{DecisionAD} & \textbf{96.0} & \textbf{45.8} & \textbf{90.4} & \textbf{87.5} & \textbf{23.9} & \textbf{90.5} & \textbf{80.5} & \textbf{32.5} & \textbf{90.8} & \textbf{96.0} & \textbf{53.8} & \textbf{85.6} & \textbf{81.3} & \textbf{31.2} & \textbf{92.8} & \textbf{88.2} & \textbf{37.5} & \textbf{90.0} \\
\midrule
\multicolumn{19}{l}{\textit{Without self-rarity}} \\
Gradient $\times$ feature & 93.7 & 17.1 & 57.9 & 85.4 & 8.2 & 60.6 & 73.7 & 9.4 & 47.2 & 91.1 & 6.4 & 45.5 & 78.4 & 11.3 & 67.6 & 84.5 & 10.5 & 55.8 \\
$|$Gradient $\times$ feature$|$ & 94.9 & 25.0 & 79.9 & 84.0 & 11.1 & 76.8 & 75.5 & 17.7 & 78.9 & 90.6 & 13.3 & 63.7 & 78.8 & 13.0 & 81.8 & 84.8 & 16.0 & 76.2 \\
Grad-CAM & 94.9 & 20.2 & 69.7 & 83.4 & 8.5 & 65.5 & 75.6 & 17.4 & 60.5 & 95.3 & 22.9 & 76.0 & 75.2 & 8.9 & 79.0 & 84.9 & 15.6 & 70.1 \\
Gradient norm & \textbf{95.4} & 32.7 & 85.9 & 85.7 & 15.9 & 84.7 & 76.9 & 26.2 & 86.7 & 94.1 & 37.2 & 79.9 & 80.2 & 20.7 & 89.3 & 86.5 & 26.5 & 85.3 \\
Answer attribution & \textbf{95.4} & \textbf{38.8} & \textbf{87.9} & \textbf{86.9} & \textbf{20.2} & \textbf{86.4} & \textbf{79.3} & \textbf{28.8} & \textbf{88.1} & \textbf{96.0} & \textbf{49.0} & \textbf{82.9} & \textbf{80.7} & \textbf{26.3} & \textbf{90.6} & \textbf{87.7} & \textbf{32.6} & \textbf{87.2} \\
\bottomrule
\end{tabular}
}
\end{table*}

\paragraph{Other attributions.}
We also replace answer attribution by four simpler gradient attributions of the answer score at the same layer and keep all other steps (\cref{tab:supp_attr}).
They are gradient $\times$ feature, its absolute value, Grad-CAM~\cite{selvaraju2017gradcam}, and the gradient norm without the logarithm.
Gradient $\times$ feature sums the product of the gradient and the token over the channels, and Grad-CAM weights each channel by its mean gradient over the tokens.
\method is the best or tied with every one of them on every dataset and metric.
The mean image AUROC decreases by 1.2--2.7\%p.
The answer score carries most of the image-level answer, so the attribution matters more for the map.
The map depends on the attribution, as the mean pixel AP decreases from 37.5 to 21.9--32.1.
The logarithm alone accounts for 5.4\%p.
Without self-rarity, the mean pixel AP of these attributions is 10.5--26.5, against 32.6 for answer attribution.

\begin{table*}[!htbp]
\centering
\caption{\textbf{Same method} on CLIP ViT-L/14@336 (\%, \method shaded), with the metrics of \cref{tab:industrial}. Train marks auxiliary training. The best zero-shot result is in bold and the second underlined. The last four rows use $k$ normal references (mean of three seeds).}
\label{tab:supp_clip}
\fitbox{\scriptsize
\setlength{\tabcolsep}{1.5pt}
\begin{tabular}{l l c cccc cccc cccc cccc cccc cccc}
\toprule
\multicolumn{1}{l}{\multirow{3}{*}[-1\dimexpr\aboverulesep+\belowrulesep\relax]{Method}} & \multirow{3}{*}[-1\dimexpr\aboverulesep+\belowrulesep\relax]{Venue} & \multirow{3}{*}[-1\dimexpr\aboverulesep+\belowrulesep\relax]{Train} & \multicolumn{4}{c}{MVTec AD} & \multicolumn{4}{c}{VisA} & \multicolumn{4}{c}{MPDD} & \multicolumn{4}{c}{BTAD} & \multicolumn{4}{c}{Real-IAD} & \multicolumn{4}{c}{Mean} \\
\cmidrule(lr){4-7} \cmidrule(lr){8-11} \cmidrule(lr){12-15} \cmidrule(lr){16-19} \cmidrule(lr){20-23} \cmidrule(lr){24-27}
 &  &  & \multicolumn{2}{c}{Image} & \multicolumn{2}{c}{Pixel} & \multicolumn{2}{c}{Image} & \multicolumn{2}{c}{Pixel} & \multicolumn{2}{c}{Image} & \multicolumn{2}{c}{Pixel} & \multicolumn{2}{c}{Image} & \multicolumn{2}{c}{Pixel} & \multicolumn{2}{c}{Image} & \multicolumn{2}{c}{Pixel} & \multicolumn{2}{c}{Image} & \multicolumn{2}{c}{Pixel} \\
\cmidrule(lr){4-5} \cmidrule(lr){6-7} \cmidrule(lr){8-9} \cmidrule(lr){10-11} \cmidrule(lr){12-13} \cmidrule(lr){14-15} \cmidrule(lr){16-17} \cmidrule(lr){18-19} \cmidrule(lr){20-21} \cmidrule(lr){22-23} \cmidrule(lr){24-25} \cmidrule(lr){26-27}
 &  &  & AUC & AP & AUC & PRO & AUC & AP & AUC & PRO & AUC & AP & AUC & PRO & AUC & AP & AUC & PRO & AUC & AP & AUC & PRO & AUC & AP & AUC & PRO \\
\midrule
APRIL-GAN & arXiv'23 & \checkmark & 86.0 & 93.4 & 86.7 & 84.6 & 77.5 & 80.8 & 94.2 & 86.9 & 76.5 & 82.6 & 94.3 & 84.7 & 75.5 & 70.7 & 89.3 & 69.0 & 68.4 & 64.5 & 95.2 & 86.6 & 76.8 & 78.4 & 91.9 & 82.4 \\
AnomalyCLIP & ICLR'24 & \checkmark & 91.6 & 96.3 & 91.1 & 85.3 & 82.0 & 85.3 & 95.5 & 86.6 & 77.5 & 82.5 & 96.5 & 89.0 & 88.8 & 90.2 & 94.3 & 74.4 & 68.9 & 63.5 & 95.0 & 85.0 & 81.8 & 83.6 & 94.5 & 84.1 \\
VCP-CLIP & ECCV'24 & \checkmark & 89.8 & 95.5 & \textbf{92.1} & 87.7 & 83.8 & \underline{87.6} & \underline{95.8} & \underline{90.0} & 69.3 & 72.9 & \underline{96.8} & 90.0 & 89.7 & 94.0 & 93.8 & 75.0 & 72.4 & 69.9 & \underline{96.4} & 87.7 & 81.0 & 84.0 & \textbf{95.0} & 86.1 \\
Bayes-PFL & CVPR'25 & \checkmark & \underline{92.2} & 96.2 & \underline{91.9} & \textbf{88.1} & \textbf{86.9} & \textbf{89.3} & \textbf{95.9} & 88.8 & 76.9 & 78.7 & \textbf{97.1} & \underline{90.8} & 91.2 & 90.6 & 92.2 & 77.8 & 72.6 & 68.9 & 96.0 & 85.8 & 84.0 & 84.7 & 94.6 & \underline{86.2} \\
AA-CLIP & CVPR'25 & \checkmark & 89.8 & 95.1 & 91.3 & 87.6 & 78.1 & 81.8 & 94.7 & 81.8 & 74.3 & 79.2 & 96.3 & 87.0 & \textbf{95.4} & \textbf{98.0} & \textbf{97.1} & \textbf{79.7} & 75.0 & 73.2 & 95.0 & 81.2 & 82.5 & 85.4 & \underline{94.9} & 83.5 \\
FAPrompt & ICCV'25 & \checkmark & 91.8 & 95.6 & 89.4 & 85.5 & \underline{85.0} & 86.7 & \underline{95.8} & 87.9 & \textbf{79.8} & \underline{83.0} & 96.3 & 88.3 & 93.0 & \underline{96.2} & \underline{96.4} & 78.5 & 77.4 & 74.8 & 94.7 & 84.8 & \textbf{85.4} & \underline{87.3} & 94.5 & 85.0 \\
MoECLIP & CVPR'26 & \checkmark & \textbf{93.0} & \textbf{97.0} & 91.5 & 86.2 & 82.8 & 84.8 & 94.8 & 83.9 & 68.3 & 73.9 & \underline{96.8} & 87.8 & 90.3 & 95.2 & \underline{96.4} & 76.7 & \textbf{79.0} & \textbf{76.8} & 94.5 & 79.7 & 82.7 & 85.6 & 94.8 & 82.8 \\
VisualAD & CVPR'26 & \checkmark & \underline{92.2} & \underline{96.7} & 90.8 & \underline{87.9} & 84.6 & \underline{87.6} & 95.7 & \textbf{90.2} & 71.5 & 79.5 & \textbf{97.1} & \textbf{92.5} & \underline{95.0} & \textbf{98.0} & 91.1 & \underline{79.2} & \underline{78.5} & \underline{76.2} & \textbf{96.6} & \textbf{91.8} & \underline{84.4} & \textbf{87.6} & 94.3 & \textbf{88.3} \\
\rowcolor{oursrow}\textbf{DecisionAD} & -- & -- & 91.4 & 96.0 & 89.4 & 84.2 & 81.9 & 84.9 & 93.7 & 86.4 & 78.3 & \textbf{84.8} & 95.9 & 88.2 & 89.9 & 93.3 & 88.9 & 73.1 & 73.2 & 70.4 & 95.0 & 87.5 & 83.0 & 85.9 & 92.6 & 83.9 \\
\rowcolor{oursrow}\textbf{DecisionAD}, no refinement & -- & -- & 91.1 & 96.0 & 90.0 & 84.9 & 80.9 & 84.0 & 93.9 & 87.1 & \underline{78.6} & \textbf{84.8} & 96.0 & 88.2 & 90.0 & 93.6 & 89.0 & 73.2 & 73.3 & 70.6 & 95.4 & \underline{89.4} & 82.8 & 85.8 & 92.9 & 84.5 \\
\cmidrule{1-27}
\rowcolor{oursrow}\textbf{DecisionAD}, $k{=}1$ & -- & -- & 93.9 & 97.3 & 92.0 & 88.7 & 88.1 & 90.1 & 96.0 & 89.0 & 83.0 & 85.9 & 96.9 & 91.8 & 95.7 & 97.3 & 92.6 & 78.8 & 71.3 & 67.6 & 95.3 & 86.2 & 86.4 & 87.6 & 94.5 & 86.9 \\
\rowcolor{oursrow}\textbf{DecisionAD}, $k{=}2$ & -- & -- & 94.7 & 97.5 & 92.4 & 89.3 & 89.0 & 90.8 & 96.2 & 89.5 & 86.4 & 88.8 & 97.1 & 92.2 & 96.0 & 97.7 & 92.8 & 78.7 & 73.0 & 69.0 & 95.6 & 87.3 & 87.8 & 88.8 & 94.8 & 87.4 \\
\rowcolor{oursrow}\textbf{DecisionAD}, $k{=}4$ & -- & -- & 96.4 & 98.2 & 92.7 & 89.8 & 90.0 & 91.6 & 96.4 & 90.0 & 89.1 & 92.0 & 97.3 & 92.9 & 96.4 & 98.2 & 93.0 & 79.0 & 76.7 & 73.1 & 96.3 & 89.4 & 89.7 & 90.6 & 95.1 & 88.2 \\
\rowcolor{oursrow}\textbf{DecisionAD}, $k{=}8$ & -- & -- & 97.1 & 98.5 & 93.0 & 90.3 & 90.8 & 92.3 & 96.7 & 90.5 & 91.6 & 94.2 & 97.5 & 93.6 & 96.3 & 98.7 & 93.2 & 79.0 & 79.3 & 76.3 & 96.6 & 90.7 & 91.0 & 92.0 & 95.4 & 88.8 \\
\bottomrule
\end{tabular}
}
\end{table*}

\begin{table*}[!htbp]
\centering
\caption{\textbf{Baselines on the same encoder.} Four auxiliary-trained methods retrained on C-RADIOv4-SO400M (\%, \method shaded), with the metrics of \cref{tab:industrial}. Train marks auxiliary training.}
\label{tab:supp_ports}
\fitbox{\scriptsize
\setlength{\tabcolsep}{1.6pt}
\begin{tabular}{l l c cccc cccc cccc cccc cccc cccc}
\toprule
\multicolumn{1}{l}{\multirow{3}{*}[-1\dimexpr\aboverulesep+\belowrulesep\relax]{Method}} & \multirow{3}{*}[-1\dimexpr\aboverulesep+\belowrulesep\relax]{Venue} & \multirow{3}{*}[-1\dimexpr\aboverulesep+\belowrulesep\relax]{Train} & \multicolumn{4}{c}{MVTec AD} & \multicolumn{4}{c}{VisA} & \multicolumn{4}{c}{MPDD} & \multicolumn{4}{c}{BTAD} & \multicolumn{4}{c}{Real-IAD} & \multicolumn{4}{c}{Mean} \\
\cmidrule(lr){4-7} \cmidrule(lr){8-11} \cmidrule(lr){12-15} \cmidrule(lr){16-19} \cmidrule(lr){20-23} \cmidrule(lr){24-27}
 &  &  & \multicolumn{2}{c}{Image} & \multicolumn{2}{c}{Pixel} & \multicolumn{2}{c}{Image} & \multicolumn{2}{c}{Pixel} & \multicolumn{2}{c}{Image} & \multicolumn{2}{c}{Pixel} & \multicolumn{2}{c}{Image} & \multicolumn{2}{c}{Pixel} & \multicolumn{2}{c}{Image} & \multicolumn{2}{c}{Pixel} & \multicolumn{2}{c}{Image} & \multicolumn{2}{c}{Pixel} \\
\cmidrule(lr){4-5} \cmidrule(lr){6-7} \cmidrule(lr){8-9} \cmidrule(lr){10-11} \cmidrule(lr){12-13} \cmidrule(lr){14-15} \cmidrule(lr){16-17} \cmidrule(lr){18-19} \cmidrule(lr){20-21} \cmidrule(lr){22-23} \cmidrule(lr){24-25} \cmidrule(lr){26-27}
 &  &  & AUC & AP & AUC & PRO & AUC & AP & AUC & PRO & AUC & AP & AUC & PRO & AUC & AP & AUC & PRO & AUC & AP & AUC & PRO & AUC & AP & AUC & PRO \\
\midrule
APRIL-GAN~\cite{chen2023aprilgan} & arXiv'23 & \checkmark & \underline{95.3} & \underline{97.3} & 85.8 & \underline{84.2} & \underline{85.6} & 87.3 & \textbf{95.8} & \underline{89.1} & \textbf{81.2} & \underline{84.0} & 94.7 & 85.0 & 92.4 & 95.3 & \underline{95.9} & 75.3 & 77.2 & 74.1 & 96.1 & 85.3 & \underline{86.3} & \underline{87.6} & 93.7 & 83.8 \\
AnomalyCLIP~\cite{zhou2024anomalyclip} & ICLR'24 & \checkmark & 92.0 & 96.3 & 85.2 & 76.6 & 80.5 & 84.2 & 91.6 & 81.9 & 78.2 & \textbf{84.7} & \underline{96.8} & 89.5 & 84.3 & 92.0 & 81.0 & 53.5 & 69.9 & 66.7 & 95.2 & 85.1 & 81.0 & 84.8 & 89.9 & 77.3 \\
FAPrompt~\cite{zhu2025faprompt} & ICCV'25 & \checkmark & 91.0 & 95.7 & 88.2 & 82.3 & 83.6 & 87.1 & 91.5 & 77.1 & 76.2 & 80.8 & 95.3 & 85.2 & 88.5 & 95.6 & 79.6 & 53.4 & 78.6 & \underline{77.4} & 96.2 & 87.1 & 83.6 & 87.3 & 90.2 & 77.0 \\
VisualAD~\cite{hou2026visualad} & CVPR'26 & \checkmark & 92.6 & 96.2 & \textbf{92.1} & \textbf{90.4} & 84.6 & \underline{87.6} & \underline{95.6} & \textbf{90.5} & 68.9 & 76.6 & 96.3 & \underline{89.6} & \underline{93.2} & \underline{98.1} & 95.7 & \underline{82.5} & \underline{78.9} & 77.1 & \textbf{97.8} & \underline{92.7} & 83.6 & 87.1 & \textbf{95.5} & \underline{89.1} \\
\midrule
\rowcolor{oursrow}\textbf{DecisionAD} & -- & -- & \textbf{96.0} & \textbf{97.9} & \underline{91.6} & \textbf{90.4} & \textbf{87.5} & \textbf{89.2} & 95.1 & \textbf{90.5} & \underline{80.5} & 83.6 & \textbf{97.0} & \textbf{90.8} & \textbf{96.0} & \textbf{99.1} & \textbf{96.8} & \textbf{85.6} & \textbf{81.3} & \textbf{79.4} & \underline{96.7} & \textbf{92.8} & \textbf{88.2} & \textbf{89.8} & \underline{95.4} & \textbf{90.0} \\
\bottomrule
\end{tabular}
}
\end{table*}

\paragraph{Same method.}
The eight auxiliary-trained methods of \cref{tab:industrial} use CLIP ViT-L/14@336.
Its vision encoder has 304M parameters in 24 layers, against 431M in 27 layers for C-RADIOv4-SO400M.
We run \method on this encoder without tuning.
Both passes use 518\,px, the input size of these methods (\cref{tab:repro_baselines}), which gives the same $37{\times}37$ patch grid. The neighbor settings, the refinement settings, and the windows of \cref{tab:impl} are therefore unchanged.
The layers are transferred by relative depth, to layers 9--16 of 24 for self-rarity and layer 12 for answer attribution.
As this encoder has no SAM3 head, the refinement uses the patch features of its last layer.
The reference statistics are recomputed on $\nat$.
Without training, \method reaches a mean image AUROC of 83.0 (\cref{tab:supp_clip}). It exceeds five of the eight methods, all of which are trained on auxiliary defect data.
Its mean image AP of 85.9 exceeds six of them, its mean AUPRO of 83.9 three, and its mean pixel AUROC of 92.6 one.
Without refinement, the mean image AUROC decreases by 0.2\%p, and the mean pixel AUROC and AUPRO increase by 0.3 and 0.6\%p. On this encoder, the refinement with last-layer features thus changes each mean by at most 0.6\%p.
The last four rows of \cref{tab:supp_clip} add $k$ normal references as in \cref{sec:supp_fewshot}, still without training.
One reference raises the mean image AUROC from 83.0 to 86.4, above all eight auxiliary-trained methods.
The counterparts of these rows are the five few-shot baselines of \cref{tab:fewshot}.
In mean image AUROC, \method on this encoder matches or exceeds the best of them from $k{=}2$, against a margin of 2.7--3.1\%p with C-RADIOv4-SO400M.
On Real-IAD, one or two references lower the image AUROC (71.3 and 73.0 \vs 73.2), as with C-RADIOv4-SO400M (\cref{sec:supp_fewshot}).

\paragraph{Baselines on the same encoder.}
In the other direction, we train four auxiliary-trained methods on C-RADIOv4-SO400M, namely APRIL-GAN, AnomalyCLIP, FAPrompt, and VisualAD (\cref{tab:supp_ports}).
Each keeps its official losses and schedule and is trained under policy A of \cref{sec:supp_baselines}, with the 592\,px input of \method.
The CLIP layers that a method reads are transferred by relative depth, to layers 7, 14, 20, and 27 of 27. FAPrompt reads only the last of them, as in its official code.
APRIL-GAN, AnomalyCLIP, and FAPrompt take their text features from the text encoder of the SigLIP2 head, and VisualAD uses no text.
The encoder raises the mean image AUROC of APRIL-GAN from 76.8 to 86.3, and it does not raise that of AnomalyCLIP (81.0 \vs 81.8) or FAPrompt (83.6 \vs 85.4).
VisualAD follows policy B in \cref{tab:industrial}, and on MVTec AD and VisA, where the two policies use the same auxiliary data, the encoder changes its image AUROC by at most 0.4\%p (92.6 and 84.6 \vs 92.2 and 84.6).
The gain from the encoder thus depends on the method.
Without training, \method achieves the highest mean image AUROC, image AP, and AUPRO in \cref{tab:supp_ports}.
Its margins in mean image AUROC and AUPRO are 1.9 and 0.9\%p, against 2.8 and 1.7\%p over the best auxiliary-trained methods of \cref{tab:industrial}.
\method thus stays ahead in these means when these four methods use its encoder, so its margins do not come from the encoder alone.

\subsection{Windows, prompt pair, refinement, and map weights}
\label{sec:supp_abl_design}
Wider window sets generally raise image AUROC, though not monotonically, and beyond $\{1,2,4,8,16\}$ the results remain almost the same (\cref{tab:supp_design}).
On industrial images, a single template remains within 0.3\%p of all 14 templates in image AUROC, and a single state raises the image AUROC on VisA (89.0 \vs 87.5) but lowers pixel AP and AUPRO. On BrainMRI (\cref{sec:supp_medical}), the medical prompt sets outperform a single sentence pair in image AUROC by 7.0\%p.
For refinement, backbone output features instead of the SAM3 head yield a higher pixel AP, by 1.9\%p on MVTec AD and 0.5\%p on VisA, with an AUPRO within 0.2\%p.
A larger temperature $\gamma$ raises pixel AP, and no other $\gamma$ yields a higher AUPRO than our $\gamma{=}0.07$.
Unequal weights of $\zA$ and $\zR$ lower pixel AP and AUPRO on both datasets and change the image AUROC by $-0.3$ to $+0.6$\%p.
Equal weights, which need no tuning, thus yield the best map of the three.
The single-sentence row of \cref{tab:supp_design} uses \texttt{a photo of a flawless object.} and \texttt{a photo of a damaged object.}, the single-state row \texttt{flawless object} and \texttt{damaged object}, and the single-template row \texttt{a photo of a object.}
\begin{table}[!htbp]
\centering
\caption{\textbf{Windows, prompt pair, number of templates, refinement, and map weights} (MVTec AD, VisA, \%, our setting shaded). Window sets change only the answer, and a dot marks their unchanged map.}
\label{tab:supp_design}
\fitbox{\scriptsize
\setlength{\tabcolsep}{2.2pt}
\begin{tabular}{l cccccc}
\toprule
\multicolumn{1}{l}{\multirow{3}{*}[-1\dimexpr\aboverulesep+\belowrulesep\relax]{Setting}} & \multicolumn{3}{c}{MVTec AD} & \multicolumn{3}{c}{VisA} \\
\cmidrule(lr){2-4} \cmidrule(lr){5-7}
 & Image & \multicolumn{2}{c}{Pixel} & Image & \multicolumn{2}{c}{Pixel} \\
\cmidrule(lr){2-2} \cmidrule(lr){3-4} \cmidrule(lr){5-5} \cmidrule(lr){6-7}
 & AUC & AP & PRO & AUC & AP & PRO \\
\midrule
\multicolumn{7}{l}{\textit{Window set $\mathcal{W}$}} \\
\{1\} & 94.0 & $\cdot$ & $\cdot$ & 85.0 & $\cdot$ & $\cdot$ \\
\{16\} & 95.0 & $\cdot$ & $\cdot$ & 87.0 & $\cdot$ & $\cdot$ \\
\{1,2\} & 95.0 & $\cdot$ & $\cdot$ & 84.2 & $\cdot$ & $\cdot$ \\
\{1,2,4\} & 95.5 & $\cdot$ & $\cdot$ & 85.0 & $\cdot$ & $\cdot$ \\
\{1,2,4,8\} & \underline{95.9} & $\cdot$ & $\cdot$ & 86.8 & $\cdot$ & $\cdot$ \\
\rowcolor{oursrow}\{1,2,4,8,16\} & \textbf{96.0} & 45.8 & 90.4 & \underline{87.5} & 23.9 & 90.5 \\
\{1,2,4,\dots,32\} & \textbf{96.0} & $\cdot$ & $\cdot$ & \textbf{87.8} & $\cdot$ & $\cdot$ \\
\midrule
\multicolumn{7}{l}{\textit{Prompt pair}} \\
one sentence each & 95.6 & 45.6 & 90.1 & \underline{87.9} & 23.5 & 90.1 \\
1 template $\times$ all states & 95.7 & \textbf{45.9} & \underline{90.3} & 87.3 & \underline{23.7} & \underline{90.3} \\
14 templates $\times$ 1 state & \textbf{96.1} & \underline{45.8} & \underline{90.3} & \textbf{89.0} & 23.5 & \underline{90.3} \\
\rowcolor{oursrow}14 templates $\times$ all states & \underline{96.0} & \underline{45.8} & \textbf{90.4} & 87.5 & \textbf{23.9} & \textbf{90.5} \\
\midrule
\multicolumn{7}{l}{\textit{Number of prompt templates $|\mathcal{T}|$ (mean of three random subsets)}} \\
1 & \underline{95.9} & 45.8 & 90.4 & 87.2 & 23.6 & \underline{90.4} \\
2 & \textbf{96.0} & 45.8 & 90.4 & \underline{87.4} & \underline{23.8} & \underline{90.4} \\
4 & \textbf{96.0} & 45.8 & 90.4 & \underline{87.4} & \textbf{23.9} & \textbf{90.5} \\
7 & \textbf{96.0} & 45.8 & 90.4 & \underline{87.4} & \underline{23.8} & \textbf{90.5} \\
\rowcolor{oursrow}14 & \textbf{96.0} & 45.8 & 90.4 & \textbf{87.5} & \textbf{23.9} & \textbf{90.5} \\
\midrule
\multicolumn{7}{l}{\textit{Affinity refinement}} \\
\rowcolor{oursrow}SAM3 head, $|\mathcal{K}_i|{=}16$, $\gamma{=}0.07$ & \underline{96.0} & 45.8 & \underline{90.4} & \underline{87.5} & 23.9 & \textbf{90.5} \\
SigLIP2 head features & \textbf{96.1} & 46.8 & \underline{90.4} & \textbf{87.8} & 23.0 & 90.2 \\
backbone output features & \textbf{96.1} & \underline{47.7} & \textbf{90.6} & 87.2 & \underline{24.4} & \underline{90.3} \\
$|\mathcal{K}_i|{=}8$ & \underline{96.0} & 45.5 & 90.3 & 87.3 & 23.8 & \textbf{90.5} \\
$|\mathcal{K}_i|{=}32$ & \underline{96.0} & 46.0 & \underline{90.4} & \underline{87.5} & 23.9 & \textbf{90.5} \\
$\gamma{=}0.03$ & 95.9 & 44.2 & 90.1 & 86.9 & 22.7 & \textbf{90.5} \\
$\gamma{=}0.15$ & \textbf{96.1} & \textbf{47.8} & 90.1 & 87.2 & \textbf{24.7} & 89.7 \\
\midrule
\multicolumn{7}{l}{\textit{Map weights $z_A{:}z_R$}} \\
\rowcolor{oursrow}1:1 & \underline{96.0} & \textbf{45.8} & \textbf{90.4} & \underline{87.5} & \textbf{23.9} & \textbf{90.5} \\
2:1 & \textbf{96.1} & 44.5 & \underline{90.1} & 87.3 & \underline{23.4} & 89.8 \\
1:2 & 95.7 & \underline{45.7} & 90.0 & \textbf{88.1} & \underline{23.4} & \underline{90.4} \\
\bottomrule
\end{tabular}
}
\end{table}

\subsection{Prompt templates, natural-image reference, and resolution}
\label{sec:supp_abl_other}
\begin{figure}[!htbp]
\centering
\includegraphics[width=0.45\linewidth]{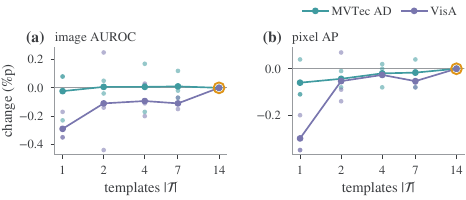}
\caption{\textbf{Number of prompt templates $|\mathcal{T}|$} (all states, change from our 14 templates, \%p).
Dots are three random subsets of the templates, lines their mean, and the ring our setting.}
\label{fig:template_count}
\end{figure}

\paragraph{Prompt templates.}
\Cref{fig:template_count} varies the number of templates with all states, using three random subsets of the 14 templates for each size, and \cref{tab:supp_design} lists their means.
All subsets remain within 0.5\%p of our setting in image AUROC and pixel AP.
The subsets are drawn with seed $1000m+r$ for $m$ templates and draw $r$.

\paragraph{Natural-image reference.}
Individual draws of 60 COCO images move the image AUROC on MVTec AD within 95.8--96.1 (\cref{tab:supp_reference}), and \cref{fig:ablation}d plots the same results.
The disjoint COCO set and the Imagenette set are within 0.1\%p of our setting.
Five other draws of 480 images, drawn as $\nat$ with seeds 1--5, yield an image AUROC of 95.9--96.0 on MVTec AD and 87.3--87.6 on VisA.
Their pixel AP and AUPRO differ from ours by less than 0.1\%p.
The reference supplies only a mean and a standard deviation per quantity, and 480 images estimate both stably.
Without a reference, when each map is standardized within its own image, pixel AP decreases by 5.7\%p on MVTec AD and 3.3\%p on VisA.
The scale must thus come from outside the image, and which natural images supply it matters little.
The 60-, 120-, and 240-image subsets are \texttt{random.Random(r).sample(range(480), m)} for $m$ images and draw $r\in\{0,1,2\}$. The disjoint COCO set is drawn with seed 17 after removing the 480 images, and the Imagenette set with seed 17 from its validation split.
\begin{table}[!htbp]
\centering
\begin{minipage}[t]{0.485\linewidth}
\centering
\caption{\textbf{Natural-image reference $\nat$} (MVTec AD, VisA, \%, our setting shaded). ``draw'' indexes the random draws. The last block summarizes five other draws of 480 images. Values are not ranked.}
\label{tab:supp_reference}
\fitbox{\scriptsize
\setlength{\tabcolsep}{2.2pt}
\begin{tabular}{l cccccc}
\toprule
\multicolumn{1}{l}{\multirow{3}{*}[-1\dimexpr\aboverulesep+\belowrulesep\relax]{Setting}} & \multicolumn{3}{c}{MVTec AD} & \multicolumn{3}{c}{VisA} \\
\cmidrule(lr){2-4} \cmidrule(lr){5-7}
 & Image & \multicolumn{2}{c}{Pixel} & Image & \multicolumn{2}{c}{Pixel} \\
\cmidrule(lr){2-2} \cmidrule(lr){3-4} \cmidrule(lr){5-5} \cmidrule(lr){6-7}
 & AUC & AP & PRO & AUC & AP & PRO \\
\midrule
\multicolumn{7}{l}{\textit{Natural-image reference $\mathcal{N}$}} \\
COCO 60 (draw 0) & 95.8 & 45.9 & 90.4 & 87.2 & 23.9 & 90.5 \\
COCO 60 (draw 1) & 96.1 & 45.8 & 90.4 & 87.6 & 23.9 & 90.5 \\
COCO 60 (draw 2) & 95.9 & 45.9 & 90.4 & 87.3 & 23.9 & 90.5 \\
COCO 120 (draw 0) & 96.0 & 45.9 & 90.4 & 87.4 & 23.9 & 90.5 \\
COCO 120 (draw 1) & 96.1 & 45.8 & 90.4 & 87.7 & 23.9 & 90.5 \\
COCO 120 (draw 2) & 95.9 & 45.9 & 90.4 & 87.4 & 23.9 & 90.5 \\
COCO 240 (draw 0) & 96.0 & 45.9 & 90.4 & 87.5 & 23.9 & 90.5 \\
COCO 240 (draw 1) & 96.0 & 45.8 & 90.4 & 87.5 & 23.9 & 90.5 \\
COCO 240 (draw 2) & 95.9 & 45.9 & 90.4 & 87.3 & 23.9 & 90.5 \\
\rowcolor{oursrow}COCO 480 & 96.0 & 45.8 & 90.4 & 87.5 & 23.9 & 90.5 \\
COCO 480 disjoint & 96.0 & 45.9 & 90.4 & 87.5 & 23.9 & 90.5 \\
Imagenette 480 & 95.9 & 45.8 & 90.4 & 87.4 & 23.9 & 90.5 \\
no reference (per-image) & 95.2 & 40.1 & 89.5 & 86.2 & 20.6 & 89.9 \\
\midrule
\multicolumn{7}{l}{\textit{Five other draws of 480 COCO images}} \\
\quad mean & 96.0 & 45.9 & 90.4 & 87.5 & 23.9 & 90.5 \\
\quad std & 0.06 & 0.01 & 0.00 & 0.12 & 0.00 & 0.00 \\
\quad min & 95.9 & 45.8 & 90.4 & 87.3 & 23.9 & 90.5 \\
\quad max & 96.0 & 45.9 & 90.4 & 87.6 & 23.9 & 90.5 \\
\bottomrule
\end{tabular}
}
\end{minipage}\hfill
\begin{minipage}[t]{0.485\linewidth}
\centering
\caption{\textbf{Input resolution} (MVTec AD, VisA, \%, our setting shaded). Time per image relative to our setting (one A100, images in memory) and GFLOPs per image.}
\label{tab:supp_resolution}
\fitbox{\scriptsize
\setlength{\tabcolsep}{2.2pt}
\begin{tabular}{l cccccccc}
\toprule
\multicolumn{1}{l}{\multirow{3}{*}[-1\dimexpr\aboverulesep+\belowrulesep\relax]{Setting}} & \multicolumn{3}{c}{MVTec AD} & \multicolumn{3}{c}{VisA} & \multirow{3}{*}[-1\dimexpr\aboverulesep+\belowrulesep\relax]{Time} & \multirow{3}{*}[-1\dimexpr\aboverulesep+\belowrulesep\relax]{GFLOPs} \\
\cmidrule(lr){2-4} \cmidrule(lr){5-7}
 & Image & \multicolumn{2}{c}{Pixel} & Image & \multicolumn{2}{c}{Pixel} &  &  \\
\cmidrule(lr){2-2} \cmidrule(lr){3-4} \cmidrule(lr){5-5} \cmidrule(lr){6-7}
 & AUC & AP & PRO & AUC & AP & PRO &  &  \\
\midrule
\multicolumn{9}{l}{\textit{Input resolution}} \\
384 px & 95.4 & 41.5 & 89.1 & 85.9 & 19.9 & 88.2 & 0.92$\times$ & 919 \\
448 px & 95.5 & 43.1 & 89.7 & 87.3 & 21.1 & 89.7 & 0.93$\times$ & 1300 \\
512 px & \textbf{96.2} & 44.5 & 90.2 & 87.0 & 22.9 & 90.0 & 1.00$\times$ & 1774 \\
\rowcolor{oursrow}592 px & \underline{96.0} & 45.8 & 90.4 & 87.5 & 23.9 & 90.5 & 1.00$\times$ & 2522 \\
672 px & 95.8 & \underline{46.5} & \underline{90.5} & \underline{87.9} & \underline{24.7} & \underline{90.9} & 1.44$\times$ & 3473 \\
768 px & 95.9 & \textbf{47.2} & \textbf{90.6} & \textbf{89.2} & \textbf{25.3} & \textbf{91.2} & 2.03$\times$ & 4940 \\
\bottomrule
\end{tabular}
}
\end{minipage}
\end{table}

\paragraph{Resolution.}
Larger inputs raise pixel AP at the cost of speed (\cref{tab:supp_resolution}).
At 592\,px, the grid matches the $37{\times}37$ grid of the baselines.

\FloatBarrier
\section{Few-shot extension}
\label{sec:supp_fewshot}
This section describes the few-shot variant of \method and reports its results.

\paragraph{Method.}
Given $k$ normal reference images, we replace self-rarity by \emph{reference rarity} and move the center of the answer score to the mean over the references. The shift does not change ranking metrics within a category.
Reference rarity is the distance to the most similar reference patch, one minus the maximum cosine similarity, averaged over layers 10--18.
Its reference statistics are obtained by measuring the same distance for every COCO image against the next $k$ COCO images.
All other steps are unchanged, no weights are trained, and values are averaged over three seeds.

\paragraph{Normal references and baselines.}
For each category and each of the seeds 0, 1, and 2, the normal training images are sorted, shuffled with \texttt{random.Random(f"\{seed\}:\{category\}")}, and the first eight are kept.
These images are \texttt{train/good} for MVTec AD and MPDD, \texttt{train/ok} for BTAD, the normal training rows of \texttt{1cls.csv} for VisA, and the normal training entries for Real-IAD.
The sets for $k=1,2,4,8$ are nested prefixes, and every method uses the same references.
On HeadCT, BrainMRI, and Br35H, the references come from the normal test images, and the union of the three 8-shot sets is removed once, leaving 78/100, 75/155, and 1,476/1,500 normal/anomalous images for every method, $k$, and seed.
The few-shot baselines use their official code and released encoders. AnomalyDINO and AdaptCLIP fit no parameters to the references.
PromptAD trains per-category prompts for 100 epochs and FoundAD a projector for 2,000 epochs on them, and both use the last epoch.
KAG-prompt and AdaptCLIP use their released adapters under policy B (\cref{sec:supp_baselines}).

\begin{figure}[!htbp]
\centering
\includegraphics[width=\linewidth]{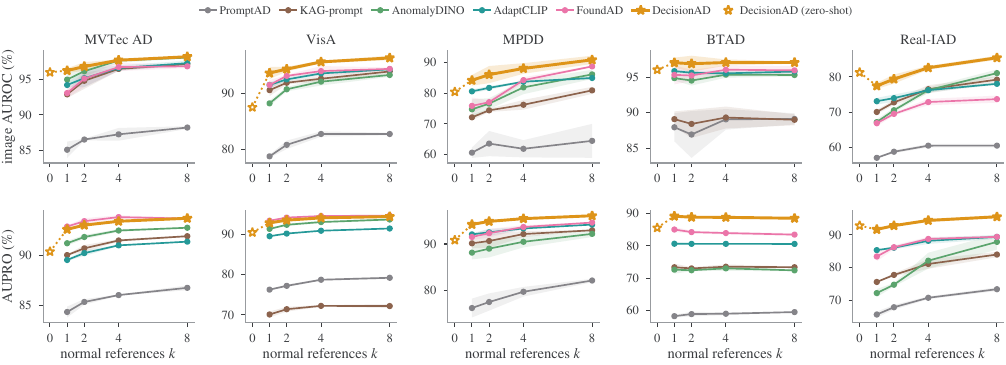}
\caption{\textbf{Few-shot results versus the number of normal references $k$} (five industrial datasets, category macro, mean of three seeds, bands show $\pm$1 standard deviation).
The hollow star at $k=0$ is zero-shot \method.}
\label{fig:supp_fewshot}
\end{figure}

\paragraph{Results.}
The five industrial datasets of the main paper are reported in \cref{fig:supp_fewshot,tab:fewshot}, and the three image-level medical datasets in \cref{tab:fewshot_med} ($k=1,2,4,8$).
In industrial image AUROC, \method ranks first for all $k$ on all five datasets. Its smallest margins are on MVTec AD at $k=4$ and $k=8$, where AnomalyDINO is 0.1\%p below.
On Real-IAD, its image AUROC falls below the zero-shot result with one or two references (77.4 and 79.4 \vs 81.3) and exceeds it from four references on.
Real-IAD shows each part from five camera views, which one or two references cannot cover.
On MPDD, BTAD, and Real-IAD, it also ranks first in pixel AUROC and AUPRO for all $k$. On MVTec AD and VisA, it is second to FoundAD in AUPRO, at most 0.7\%p below, and ties it on MVTec AD at $k=8$.
On average over the five datasets, it ranks first in image AUROC, image AP, and AUPRO for all $k$ and is the fastest method.
In medical image AUROC, it ranks first on HeadCT for all $k$ and on BrainMRI for $k=2,4$, ties AdaptCLIP on BrainMRI for $k=8$ (94.1), and is second to AdaptCLIP in the remaining settings. AdaptCLIP is trained on auxiliary defect data. Without training, \method is within 1.1\%p of it in these settings.
\begin{table}[!htbp]
\centering
\caption{\textbf{Few-shot results on industrial datasets} (category macro, \%, mean of three seeds).
Metrics, Mean, and FPS as in \cref{tab:industrial}. Best and second-best are marked within each $k$.
Train is the data used to train for anomaly detection (-- none, AD auxiliary defect datasets, Ref.\ the $k$ references, on which PromptAD trains per-category prompts and FoundAD a projector).}
\label{tab:fewshot}
\fitbox{\scriptsize
\setlength{\tabcolsep}{1.5pt}
\begin{tabular}{l l c cccc cccc cccc cccc cccc cccc r}
\toprule
\multicolumn{1}{l}{\multirow{3}{*}[-1\dimexpr\aboverulesep+\belowrulesep\relax]{Method}} & \multirow{3}{*}[-1\dimexpr\aboverulesep+\belowrulesep\relax]{Venue} & \multirow{3}{*}[-1\dimexpr\aboverulesep+\belowrulesep\relax]{Train} & \multicolumn{4}{c}{MVTec AD} & \multicolumn{4}{c}{VisA} & \multicolumn{4}{c}{MPDD} & \multicolumn{4}{c}{BTAD} & \multicolumn{4}{c}{Real-IAD} & \multicolumn{4}{c}{Mean} & \multirow{3}{*}[-1\dimexpr\aboverulesep+\belowrulesep\relax]{FPS} \\
\cmidrule(lr){4-7} \cmidrule(lr){8-11} \cmidrule(lr){12-15} \cmidrule(lr){16-19} \cmidrule(lr){20-23} \cmidrule(lr){24-27}
 &  &  & \multicolumn{2}{c}{Image} & \multicolumn{2}{c}{Pixel} & \multicolumn{2}{c}{Image} & \multicolumn{2}{c}{Pixel} & \multicolumn{2}{c}{Image} & \multicolumn{2}{c}{Pixel} & \multicolumn{2}{c}{Image} & \multicolumn{2}{c}{Pixel} & \multicolumn{2}{c}{Image} & \multicolumn{2}{c}{Pixel} & \multicolumn{2}{c}{Image} & \multicolumn{2}{c}{Pixel} &  \\
\cmidrule(lr){4-5} \cmidrule(lr){6-7} \cmidrule(lr){8-9} \cmidrule(lr){10-11} \cmidrule(lr){12-13} \cmidrule(lr){14-15} \cmidrule(lr){16-17} \cmidrule(lr){18-19} \cmidrule(lr){20-21} \cmidrule(lr){22-23} \cmidrule(lr){24-25} \cmidrule(lr){26-27}
 &  &  & AUC & AP & AUC & PRO & AUC & AP & AUC & PRO & AUC & AP & AUC & PRO & AUC & AP & AUC & PRO & AUC & AP & AUC & PRO & AUC & AP & AUC & PRO &  \\
\midrule
DecisionAD, zero-shot ($k{=}0$) & -- & -- & 96.0 & 97.9 & 91.6 & 90.4 & 87.5 & 89.2 & 95.1 & 90.5 & 80.5 & 83.6 & 97.0 & 90.8 & 96.0 & 99.1 & 96.8 & 85.6 & 81.3 & 79.4 & 96.7 & 92.8 & 88.2 & 89.8 & 95.4 & 90.0 & 25.1 \\
\midrule
\multicolumn{28}{l}{\textit{$k=1$}} \\
PromptAD~\cite{li2024promptad} & CVPR'24 & Ref. & 85.1 & 92.7 & 93.0 & 84.3 & 78.8 & 81.4 & 91.8 & 76.3 & 60.7 & 68.0 & 92.6 & 76.2 & 88.0 & 87.5 & 90.2 & 58.2 & 57.0 & 50.6 & 88.9 & 65.6 & 73.9 & 76.1 & 91.3 & 72.1 & \underline{15.4} \\
KAG-prompt~\cite{tao2025kagprompt} & AAAI'25 & AD & 92.9 & 96.7 & 95.8 & 90.0 & 90.6 & 92.2 & 90.3 & 70.1 & 72.2 & 75.0 & 96.8 & 90.2 & 89.1 & 96.1 & 95.8 & 73.4 & 70.0 & 68.2 & 93.9 & 75.6 & 83.0 & 85.7 & 94.5 & 79.8 & 3.0 \\
AnomalyDINO~\cite{damm2025anomalydino} & WACV'25 & -- & \underline{94.9} & 97.2 & \underline{95.9} & 91.2 & 88.2 & 88.9 & \underline{97.4} & 91.4 & 74.8 & 75.6 & 95.6 & 88.1 & 94.8 & 96.8 & 96.5 & 72.6 & 67.2 & 62.0 & 91.2 & 72.2 & 84.0 & 84.1 & 95.3 & 83.1 & 8.9 \\
AdaptCLIP~\cite{gao2025adaptclip} & AAAI'26 & AD & 94.2 & \underline{97.4} & 94.2 & 89.5 & \underline{91.5} & \underline{93.1} & 96.9 & 89.5 & \underline{80.6} & \underline{83.6} & \underline{97.6} & \underline{92.1} & \underline{95.8} & \underline{98.5} & 97.6 & 80.7 & \underline{73.1} & \underline{69.7} & \underline{95.5} & \underline{85.3} & \underline{87.0} & \underline{88.4} & 96.4 & 87.4 & 5.9 \\
FoundAD~\cite{zhai2025foundad} & ICLR'26 & Ref. & 93.0 & 96.3 & \textbf{96.6} & \textbf{92.9} & \underline{91.5} & 90.8 & \textbf{98.2} & \textbf{93.4} & 75.9 & 77.2 & 97.1 & 91.5 & 95.3 & 97.4 & \underline{98.1} & \underline{85.1} & 66.8 & 62.6 & 95.4 & 83.4 & 84.5 & 84.9 & \underline{97.1} & \underline{89.2} & 14.9 \\
\textbf{DecisionAD} & -- & -- & \textbf{96.2} & \textbf{98.3} & 95.0 & \underline{92.6} & \textbf{93.7} & \textbf{94.4} & \underline{97.4} & \underline{92.9} & \textbf{84.2} & \textbf{87.2} & \textbf{97.9} & \textbf{94.2} & \textbf{97.0} & \textbf{99.2} & \textbf{98.3} & \textbf{89.2} & \textbf{77.4} & \textbf{75.8} & \textbf{97.1} & \textbf{91.6} & \textbf{89.7} & \textbf{91.0} & \textbf{97.2} & \textbf{92.1} & \textbf{27.0} \\
\midrule
\multicolumn{28}{l}{\textit{$k=2$}} \\
PromptAD~\cite{li2024promptad} & CVPR'24 & Ref. & 86.5 & 93.6 & 93.5 & 85.3 & 80.8 & 83.2 & 91.9 & 77.2 & 63.6 & 70.0 & 93.0 & 77.5 & 87.0 & 83.2 & 90.5 & 58.9 & 58.7 & 51.9 & 89.9 & 67.8 & 75.3 & 76.4 & 91.8 & 73.3 & \underline{15.2} \\
KAG-prompt~\cite{tao2025kagprompt} & AAAI'25 & AD & 94.8 & 97.7 & 96.2 & 90.7 & 91.9 & 93.4 & 90.7 & 71.4 & 74.4 & 77.3 & 97.0 & 90.7 & 88.4 & 96.7 & 96.0 & 73.0 & 72.8 & \underline{70.6} & 94.7 & 77.7 & 84.5 & 87.2 & 94.9 & 80.7 & 2.6 \\
AnomalyDINO~\cite{damm2025anomalydino} & WACV'25 & -- & \underline{96.1} & \underline{97.9} & \underline{96.4} & 91.8 & 90.8 & 91.5 & \underline{97.8} & 92.4 & 76.6 & 78.3 & 96.3 & 89.0 & 94.5 & 97.0 & 96.8 & 72.5 & 70.5 & 64.7 & 92.1 & 74.7 & 85.7 & 85.9 & 95.9 & 84.1 & 7.7 \\
AdaptCLIP~\cite{gao2025adaptclip} & AAAI'26 & AD & 95.1 & 97.8 & 94.5 & 90.2 & 92.5 & \underline{93.8} & 97.1 & 90.2 & \underline{81.7} & \underline{84.7} & \underline{97.8} & \underline{92.5} & \underline{95.6} & \underline{98.6} & 97.7 & 80.6 & \underline{74.0} & 70.5 & 95.8 & 86.0 & \underline{87.8} & \underline{89.1} & 96.6 & 87.9 & 5.8 \\
FoundAD~\cite{zhai2025foundad} & ICLR'26 & Ref. & 95.1 & 97.5 & \textbf{97.0} & \textbf{93.4} & \underline{93.2} & 92.5 & \textbf{98.5} & \textbf{94.2} & 77.1 & 78.6 & 97.5 & 92.3 & 95.2 & 97.4 & \underline{98.1} & \underline{84.3} & 69.5 & 64.6 & \underline{96.2} & \underline{86.2} & 86.0 & 86.1 & \textbf{97.5} & \underline{90.1} & 14.9 \\
\textbf{DecisionAD} & -- & -- & \textbf{96.8} & \textbf{98.6} & 95.4 & \underline{93.0} & \textbf{94.4} & \textbf{95.0} & 97.6 & \underline{93.5} & \textbf{86.0} & \textbf{88.9} & \textbf{98.2} & \textbf{94.9} & \textbf{96.9} & \textbf{99.2} & \textbf{98.4} & \textbf{88.9} & \textbf{79.4} & \textbf{77.5} & \textbf{97.5} & \textbf{92.8} & \textbf{90.7} & \textbf{91.8} & \underline{97.4} & \textbf{92.6} & \textbf{27.2} \\
\midrule
\multicolumn{28}{l}{\textit{$k=4$}} \\
PromptAD~\cite{li2024promptad} & CVPR'24 & Ref. & 87.3 & 93.9 & 93.9 & 86.0 & 82.8 & 84.8 & 92.6 & 78.7 & 61.9 & 69.0 & 93.5 & 79.7 & 89.1 & 87.9 & 90.5 & 59.0 & 60.5 & 53.9 & 91.4 & 70.7 & 76.3 & 77.9 & 92.4 & 74.8 & 14.0 \\
KAG-prompt~\cite{tao2025kagprompt} & AAAI'25 & AD & 96.4 & 98.6 & 96.5 & 91.5 & 92.7 & 94.0 & 91.0 & 72.2 & 76.2 & 79.1 & 97.4 & 92.2 & 89.3 & 97.3 & 96.1 & 73.5 & \underline{76.5} & \underline{74.2} & 95.8 & 81.1 & 86.2 & 88.6 & 95.4 & 82.1 & 1.8 \\
AnomalyDINO~\cite{damm2025anomalydino} & WACV'25 & -- & \underline{97.6} & \underline{98.8} & \underline{96.7} & 92.5 & 92.1 & 92.7 & \underline{98.0} & 93.1 & 81.9 & 83.3 & 96.6 & 90.5 & 95.3 & 97.3 & 96.9 & 73.0 & 76.2 & 70.0 & 95.1 & 82.1 & 88.6 & 88.4 & 96.7 & 86.2 & 6.8 \\
AdaptCLIP~\cite{gao2025adaptclip} & AAAI'26 & AD & 96.5 & 98.4 & 94.8 & 91.0 & 93.6 & \underline{94.5} & 97.3 & 90.9 & 83.7 & \underline{86.5} & \underline{98.0} & 93.4 & 95.5 & \underline{98.8} & 97.7 & 80.6 & 76.2 & 73.0 & 96.4 & 88.2 & \underline{89.1} & \underline{90.2} & \underline{96.9} & 88.8 & 5.8 \\
FoundAD~\cite{zhai2025foundad} & ICLR'26 & Ref. & 96.7 & 98.5 & \textbf{97.2} & \textbf{93.8} & \underline{94.0} & 93.2 & \textbf{98.5} & \textbf{94.6} & \underline{84.3} & 86.1 & 97.8 & \underline{93.7} & \underline{96.0} & 97.8 & \underline{98.2} & \underline{84.0} & 72.8 & 67.7 & \underline{96.9} & \underline{88.7} & 88.8 & 88.6 & \textbf{97.7} & \underline{91.0} & \underline{14.9} \\
\textbf{DecisionAD} & -- & -- & \textbf{97.7} & \textbf{98.9} & 95.8 & \underline{93.4} & \textbf{95.7} & \textbf{96.0} & 97.8 & \underline{94.1} & \textbf{88.0} & \textbf{91.1} & \textbf{98.4} & \textbf{95.5} & \textbf{97.0} & \textbf{99.3} & \textbf{98.4} & \textbf{88.9} & \textbf{82.6} & \textbf{80.1} & \textbf{98.1} & \textbf{94.4} & \textbf{92.2} & \textbf{93.1} & \textbf{97.7} & \textbf{93.3} & \textbf{23.0} \\
\midrule
\multicolumn{28}{l}{\textit{$k=8$}} \\
PromptAD~\cite{li2024promptad} & CVPR'24 & Ref. & 88.2 & 94.4 & 94.2 & 86.7 & 82.8 & 84.9 & 93.0 & 79.2 & 64.5 & 70.8 & 94.2 & 82.1 & 89.1 & 88.1 & 90.7 & 59.5 & 60.5 & 53.9 & 92.4 & 73.3 & 77.0 & 78.4 & 92.9 & 76.2 & \underline{14.9} \\
KAG-prompt~\cite{tao2025kagprompt} & AAAI'25 & AD & 97.1 & \underline{98.9} & 96.8 & 91.9 & 94.0 & \underline{95.2} & 91.3 & 72.2 & 80.9 & 84.3 & 97.6 & 93.0 & 89.0 & 97.5 & 96.3 & 73.4 & 79.2 & \underline{77.7} & 96.6 & 84.0 & 88.0 & 90.7 & 95.7 & 82.9 & 1.2 \\
AnomalyDINO~\cite{damm2025anomalydino} & WACV'25 & -- & \underline{98.0} & \underline{98.9} & \underline{96.9} & \underline{92.8} & 93.3 & 93.9 & \underline{98.2} & 93.7 & 86.1 & 87.4 & 97.3 & 92.2 & 95.3 & 97.4 & 96.9 & 72.4 & \underline{81.1} & 76.2 & 96.6 & 87.8 & \underline{90.8} & 90.8 & 97.2 & 87.8 & 4.4 \\
AdaptCLIP~\cite{gao2025adaptclip} & AAAI'26 & AD & 97.2 & 98.7 & 95.0 & 91.4 & 94.3 & 95.0 & 97.5 & 91.5 & 85.0 & 88.0 & \underline{98.2} & 94.2 & 95.8 & \underline{98.9} & 97.8 & 80.6 & 78.0 & 75.3 & 96.8 & \underline{89.4} & 90.1 & \underline{91.2} & 97.0 & 89.4 & 5.5 \\
FoundAD~\cite{zhai2025foundad} & ICLR'26 & Ref. & 96.8 & 98.5 & \textbf{97.0} & \textbf{93.7} & \underline{94.4} & 93.7 & \textbf{98.5} & \textbf{94.7} & \underline{88.7} & \underline{91.2} & 98.1 & \underline{94.6} & \underline{95.9} & 98.0 & \underline{98.2} & \underline{83.5} & 73.7 & 68.6 & \underline{97.0} & \underline{89.4} & 89.9 & 90.0 & \underline{97.8} & \underline{91.2} & \underline{14.9} \\
\textbf{DecisionAD} & -- & -- & \textbf{98.1} & \textbf{99.0} & 96.0 & \textbf{93.7} & \textbf{96.4} & \textbf{96.5} & 98.0 & \underline{94.4} & \textbf{90.8} & \textbf{93.1} & \textbf{98.6} & \textbf{96.1} & \textbf{97.0} & \textbf{99.3} & \textbf{98.5} & \textbf{88.6} & \textbf{85.4} & \textbf{83.2} & \textbf{98.4} & \textbf{95.5} & \textbf{93.6} & \textbf{94.2} & \textbf{97.9} & \textbf{93.7} & \textbf{18.8} \\
\bottomrule
\end{tabular}
}
\end{table}
\begin{table}[!htbp]
\centering
\caption{\textbf{Few-shot results on image-level medical datasets} (image AUROC, \%, mean of three seeds).
As the public releases have no training images, references are drawn from the normal test images, and the union of the 8-shot references of the three seeds is removed from evaluation. The zero-shot row uses the same evaluation images and therefore differs from \cref{tab:medical}. Venue and Train as in \cref{tab:fewshot}.}
\label{tab:fewshot_med}
\fitbox{\footnotesize
\setlength{\tabcolsep}{3pt}
\begin{tabular}{l l c cccc cccc cccc}
\toprule
\multicolumn{1}{l}{\multirow{2}{*}[-0.5\dimexpr\aboverulesep+\belowrulesep\relax]{Method}} & \multirow{2}{*}[-0.5\dimexpr\aboverulesep+\belowrulesep\relax]{Venue} & \multirow{2}{*}[-0.5\dimexpr\aboverulesep+\belowrulesep\relax]{Train} & \multicolumn{4}{c}{HeadCT} & \multicolumn{4}{c}{BrainMRI} & \multicolumn{4}{c}{Br35H} \\
\cmidrule(lr){4-7} \cmidrule(lr){8-11} \cmidrule(lr){12-15}
 &  &  & $k{=}1$ & $k{=}2$ & $k{=}4$ & $k{=}8$ & $k{=}1$ & $k{=}2$ & $k{=}4$ & $k{=}8$ & $k{=}1$ & $k{=}2$ & $k{=}4$ & $k{=}8$ \\
\midrule
DecisionAD, zero-shot ($k{=}0$) & -- & -- & \multicolumn{4}{c}{96.0} & \multicolumn{4}{c}{90.2} & \multicolumn{4}{c}{92.7} \\
\midrule
PromptAD~\cite{li2024promptad} & CVPR'24 & Ref. & 65.0 & 62.7 & 66.4 & 66.0 & 76.0 & 71.3 & 72.3 & 72.5 & 67.2 & 68.7 & 74.2 & 74.6 \\
KAG-prompt~\cite{tao2025kagprompt} & AAAI'25 & AD & 87.0 & 90.0 & 91.6 & 92.4 & 85.0 & 87.1 & 88.5 & \underline{90.0} & 92.0 & 92.4 & 92.5 & 92.8 \\
AnomalyDINO~\cite{damm2025anomalydino} & WACV'25 & -- & 68.8 & 73.9 & 81.1 & 84.3 & 62.6 & 68.0 & 71.8 & 75.4 & 56.9 & 64.8 & 73.0 & 77.7 \\
AdaptCLIP~\cite{gao2025adaptclip} & AAAI'26 & AD & \underline{93.4} & \underline{95.5} & \underline{95.9} & \underline{95.8} & \textbf{93.3} & \underline{91.8} & \underline{92.3} & \textbf{94.1} & \textbf{96.0} & \textbf{96.1} & \textbf{96.3} & \textbf{96.8} \\
FoundAD~\cite{zhai2025foundad} & ICLR'26 & Ref. & 83.1 & 82.8 & 86.9 & 87.5 & 75.0 & 82.8 & 81.4 & 81.7 & 83.7 & 86.7 & 86.4 & 87.0 \\
\textbf{DecisionAD} & -- & -- & \textbf{96.6} & \textbf{97.6} & \textbf{97.4} & \textbf{97.8} & \underline{92.9} & \textbf{93.4} & \textbf{93.6} & \textbf{94.1} & \underline{95.2} & \underline{95.5} & \underline{95.5} & \underline{95.7} \\
\bottomrule
\end{tabular}
}
\end{table}

\FloatBarrier
\section{Analysis}
\label{sec:supp_qual}
This section compares the maps of all methods, analyzes where and why the margin over the baselines arises, and examines the failures of \method.

\subsection{Qualitative comparison}
\label{sec:supp_qualfig}
\begin{figure}[!htbp]
\centering
\includegraphics[width=\linewidth]{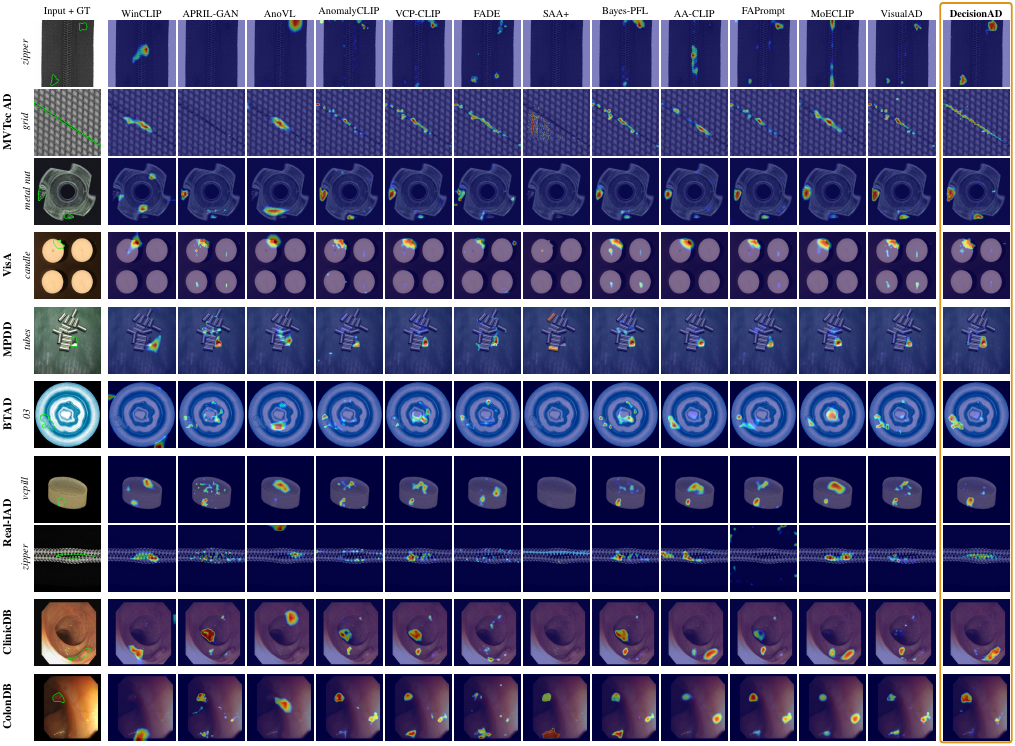}
\caption{\textbf{Qualitative comparison with all 12 baselines.} Rows are grouped by dataset. The first row of each industrial dataset shows the sample of \cref{fig:qual}, and the other rows show more cases. Display as in \cref{fig:qual}.}
\label{fig:supp_qual_rest}
\end{figure}

\Cref{fig:supp_qual_rest} shows the samples of \cref{fig:qual} with all 12 baselines, together with more cases.
The pixel metrics over all images are in \cref{tab:supp_pixel}, and failure cases in \cref{fig:supp_failures}.

\subsection{Results by defect size and image content}
\label{sec:supp_breakdown}
\begin{figure}[!htbp]
\centering
\includegraphics[width=\linewidth]{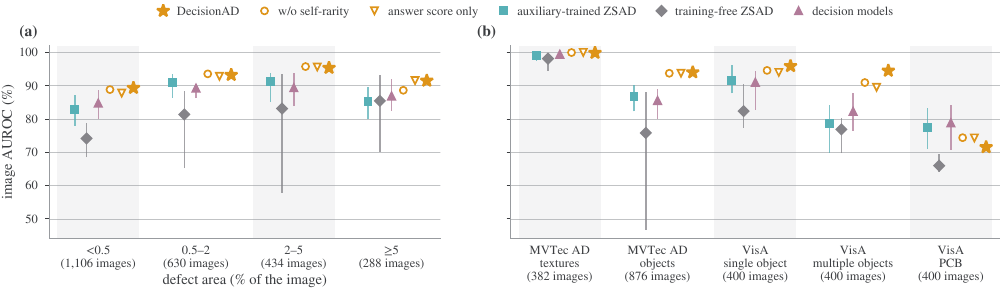}
\caption{\textbf{Image AUROC by defect size and image content} (MVTec AD and VisA, 2,458 defect images).
Markers of the baselines are group means, and bars span the lowest and the highest method of a group.}
\label{fig:supp_breakdown}
\end{figure}

To locate the margin over the baselines, we split the defect images of MVTec AD and VisA in two ways.
The first split is by defect size, measured as the share of the image that the mask covers.
The second is by the content of the image, namely textures and objects in MVTec AD, and a single object, multiple objects, or a printed circuit board (PCB) in VisA.
For each defect type, the image AUROC is computed against the normal test images of its category, and the types are pooled by their number of defect images.
We report \method, four variants from the component ablation, and the mean of each group of baselines (\cref{tab:supp_breakdown}).
\Cref{fig:supp_breakdown} shows \method without self-rarity and with the answer score alone.
\begin{table}[!htbp]
\centering
\caption{\textbf{Image AUROC by defect size and image content} (\%, MVTec AD and VisA).
Area is the median defect area. Without removes one part of the map, and $u$ from uses only the map or only the answer score for the answer.
The values of \method and its variants come from the run of the component ablation.}
\label{tab:supp_breakdown}
\fitbox{\footnotesize
\setlength{\tabcolsep}{4pt}
\begin{tabular}{l r r c cc cc ccc}
\toprule
\multicolumn{1}{l}{\multirow{2}{*}[-0.5\dimexpr\aboverulesep+\belowrulesep\relax]{Split}} & \multirow{2}{*}[-0.5\dimexpr\aboverulesep+\belowrulesep\relax]{Images} & Area & \multirow{2}{*}[-0.5\dimexpr\aboverulesep+\belowrulesep\relax]{\textbf{DecisionAD}} & \multicolumn{2}{c}{Without} & \multicolumn{2}{c}{$u$ from} & \multicolumn{3}{c}{Baselines, mean (best)} \\
\cmidrule(lr){5-6} \cmidrule(lr){7-8} \cmidrule(lr){9-11}
 &  & (\%) &  & self-rarity & answer attribution & Map & Answer & Training-free & Auxiliary-trained & Decision models \\
\midrule
\multicolumn{11}{l}{\textit{Defect area (\% of the image)}} \\
$<$0.5 & 1{,}106 & 0.16 & 89.3 & 88.8 & 89.5 & 85.9 & 87.8 & 74.2 (78.7) & 82.9 (87.3) & 84.7 (88.7) \\
0.5--2 & 630 & 0.89 & 93.2 & 93.6 & 90.9 & 88.3 & 92.8 & 81.4 (88.4) & 90.9 (93.4) & 89.2 (90.4) \\
2--5 & 434 & 2.58 & 95.4 & 95.8 & 95.5 & 93.5 & 95.6 & 83.2 (93.6) & 91.2 (93.8) & 89.4 (93.9) \\
$\geq$5 & 288 & 11.79 & 91.5 & 88.6 & 93.0 & 89.2 & 91.6 & 85.5 (93.2) & 85.2 (89.8) & 86.9 (92.1) \\
\midrule
\multicolumn{11}{l}{\textit{Image content}} \\
MVTec AD, textures & 382 & 1.60 & 99.9 & 100.0 & 99.7 & 99.6 & 100.0 & 98.1 (99.4) & 99.0 (99.7) & 99.4 (99.9) \\
MVTec AD, objects & 876 & 1.53 & 94.0 & 93.7 & 91.3 & 91.1 & 93.8 & 75.8 (88.1) & 86.9 (90.1) & 85.5 (89.1) \\
VisA, single object & 400 & 0.21 & 95.9 & 94.6 & 96.5 & 94.8 & 94.0 & 82.4 (90.6) & 91.6 (96.2) & 91.0 (94.5) \\
VisA, multiple objects & 400 & 0.07 & 94.5 & 91.0 & 95.2 & 92.5 & 89.5 & 76.9 (80.3) & 78.6 (84.2) & 82.3 (87.7) \\
VisA, PCB & 400 & 0.29 & 71.5 & 74.4 & 74.4 & 60.4 & 74.3 & 66.0 (69.6) & 77.5 (83.3) & 78.8 (84.2) \\
\bottomrule
\end{tabular}
}
\end{table}

\paragraph{Defect size.}
Defects under 0.5\% of the image account for 45\% of the defect images.
On them, the training-free methods reach 74.2 on average and the auxiliary-trained methods 82.9, against 89.3 for \method.
The margin over the training-free methods is 15.1\%p for the smallest defects and 6.0\%p for the largest.
The auxiliary-trained methods are within 2.3\%p for defects of 0.5--2\%, but they are 6.3\%p lower for the largest defects.
On average, the decision models are 4.0--6.0\%p lower for all sizes.

\begin{figure}[!htbp]
\centering
\includegraphics[width=\linewidth]{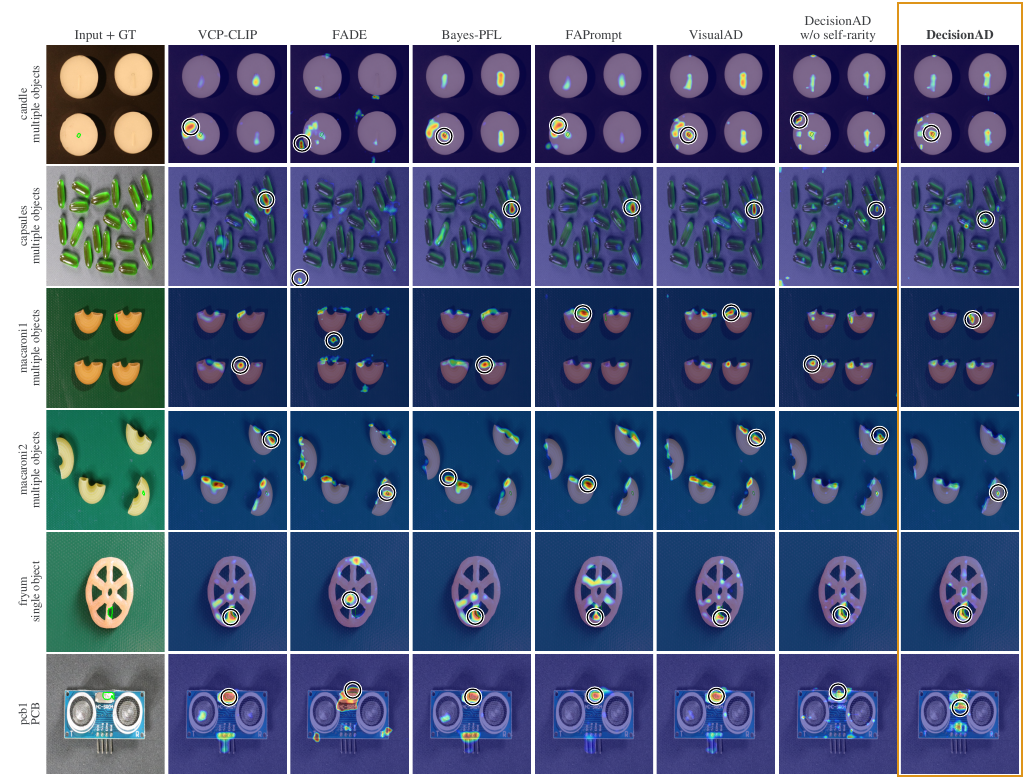}
\caption{\textbf{Maps on the three VisA groups.}
Rows 1--4 show images with multiple objects, row 5 an image with a single object, and row 6 a PCB.
The ring marks the maximum of each map. The column next to \method shows it without self-rarity. The images illustrate the cases of \cref{tab:supp_peak}. Display as in \cref{fig:qual}.}
\label{fig:supp_peak_examples}
\end{figure}

\paragraph{Image content.}
On textures, all groups exceed 98.
The margin over the best method of each group is largest on the four VisA categories with multiple objects per image, namely candle, capsules, macaroni1, and macaroni2.
Their median defect covers 0.07\% of the image.
There, the best auxiliary-trained method, training-free method, and decision model reach 84.2, 80.3, and 87.7, against 94.5 for \method.
\Cref{fig:supp_peak_examples} shows examples of the three VisA groups.

\paragraph{Components.}
On these four categories, \method falls to 91.0 without self-rarity and to 89.5 with the answer score alone, and answer attribution changes it by less than 1\%p (95.2 without it).
Over all categories, the map adds 1.5\%p to the answer score alone for the smallest defects, and the answer score alone matches \method for defects above 2\%.
On the PCB categories, \method (71.5) is above the training-free methods (66.0) and below the auxiliary-trained methods and the decision models (77.5 and 78.8).
\Cref{sec:supp_peak} examines the gain on images with multiple objects and the weakness on the PCB categories.

\FloatBarrier
\subsection{Location of the map maximum}
\label{sec:supp_peak}
\begin{figure}[!htbp]
\centering
\includegraphics[width=\linewidth]{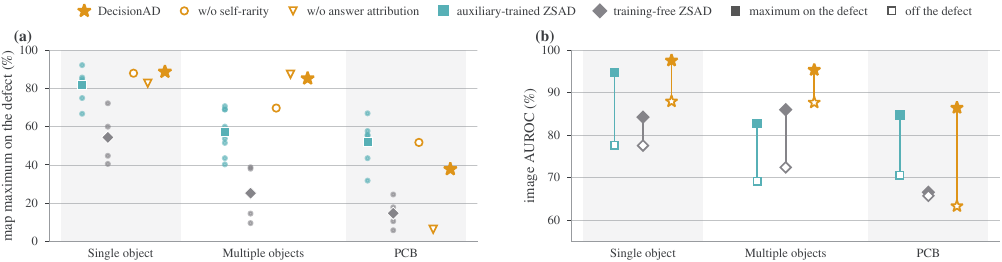}
\caption{\textbf{Location of the map maximum on VisA.}
(a) Share of defect images whose map maximum lies on the defect. Small dots are single methods, and large markers are group means.
(b) Image AUROC of the defect images whose maximum lies on the defect and of those whose maximum lies off it.}
\label{fig:supp_peak}
\end{figure}

An image-level decision depends on the strongest responses of the anomaly map.
We therefore measure, for each defect image, whether the maximum of the map lies on the defect.
The maximum counts as on the defect if it lies on the ground-truth mask or within one patch of it, that is, within 1/37 of the map side.
The analysis uses VisA, whose categories show a single object, multiple objects, or a PCB (\cref{sec:supp_breakdown}).
The maps of the baselines were regenerated with the reported runs, and their image scores reproduce the reported image AUROC.
\Cref{fig:supp_peak,tab:supp_peak} report the share of defect images whose maximum lies on the defect.
\begin{table}[!htbp]
\centering
\caption{\textbf{Location of the map maximum on VisA} (category macro, \%).
On defect is the share of defect images whose map maximum lies on the defect. Image AUC as in \cref{tab:supp_breakdown}.}
\label{tab:supp_peak}
\fitbox{\footnotesize
\setlength{\tabcolsep}{5pt}
\begin{tabular}{l cc cc cc}
\toprule
\multicolumn{1}{l}{\multirow{2}{*}[-0.5\dimexpr\aboverulesep+\belowrulesep\relax]{Method}} & \multicolumn{2}{c}{Single object} & \multicolumn{2}{c}{Multiple objects} & \multicolumn{2}{c}{PCB} \\
\cmidrule(lr){2-3} \cmidrule(lr){4-5} \cmidrule(lr){6-7}
 & On defect & Image AUC & On defect & Image AUC & On defect & Image AUC \\
\midrule
WinCLIP & 44.8 & 82.0 & 9.5 & 80.1 & 17.8 & 64.3 \\
AnoVL & 40.5 & 90.6 & 14.5 & 77.4 & 24.5 & 69.6 \\
FADE & 60.0 & 79.5 & 38.0 & 80.3 & 5.8 & 64.2 \\
SAA+ & 72.2 & 77.4 & 38.8 & 69.7 & 10.5 & 65.9 \\
\textit{Training-free, mean} & 54.4 & 82.4 & 25.2 & 76.9 & 14.6 & 66.0 \\
\midrule
APRIL-GAN & 84.8 & 91.5 & 70.8 & 69.9 & 57.8 & 70.9 \\
AnomalyCLIP & 85.8 & 89.5 & 53.8 & 80.7 & 55.2 & 75.7 \\
VCP-CLIP & 82.8 & 87.7 & 69.0 & 82.7 & 55.8 & 81.1 \\
Bayes-PFL & 83.0 & 93.4 & 60.0 & 84.2 & 51.2 & 83.0 \\
AA-CLIP & 66.8 & 91.0 & 40.2 & 71.8 & 31.8 & 71.6 \\
FAPrompt & 83.8 & 96.2 & 51.5 & 83.0 & 53.5 & 75.8 \\
MoECLIP & 75.0 & 93.4 & 43.5 & 76.5 & 43.5 & 78.6 \\
VisualAD & 92.2 & 90.4 & 69.0 & 80.3 & 67.0 & 83.3 \\
\textit{Auxiliary-trained, mean} & 81.8 & 91.6 & 57.2 & 78.6 & 52.0 & 77.5 \\
\midrule
\textbf{DecisionAD} & 88.8 & 95.9 & 85.2 & 94.5 & 37.8 & 71.5 \\
\quad without self-rarity & 88.0 & 94.6 & 69.8 & 91.0 & 51.8 & 74.4 \\
\quad without answer attribution & 82.7 & 96.5 & 87.5 & 95.2 & 6.2 & 74.4 \\
\bottomrule
\end{tabular}
}
\end{table}

\paragraph{Baselines.}
With a single object, the maximum of the auxiliary-trained methods lies on the defect in 81.8\% of the defect images on average.
With multiple objects, this share decreases to 57.2\%, and it is 25.2\% for the training-free methods.
On these categories, the image AUROC of the auxiliary-trained methods is 82.7 for the images whose maximum lies on the defect and 69.2 for the others (\cref{fig:supp_peak}b).
Their pixel AUROC on these categories remains 97.9, as this metric averages over all pixels and is insensitive to a few strong responses on normal objects.
The weakness of the baselines on these categories therefore lies in the strongest responses of their maps and not in the overall quality of the maps.
Rows 1--4 of \cref{fig:supp_peak_examples} illustrate this case with images on which the maximum of \method lies on the defect and that of most baselines lies on another object or on the background.
Row 5 shows a single object, where only the maximum of the training-free FADE lies off the defect.

\paragraph{\method.}
With multiple objects, the maximum of \method lies on the defect in 85.2\% of the defect images.
Without self-rarity, the share is 69.8\%, close to the highest share among the auxiliary-trained methods (70.8\%), and the image AUROC decreases from 94.5 to 91.0.
Self-rarity compares each patch with the other patches of the image.
The patches of repeated normal objects have similar patches elsewhere in the image and are therefore not rare, whereas a defect is.
A map derived from text prompts has no such reference within the image.
In rows 1--4 of \cref{fig:supp_peak_examples}, the maximum moves to a normal object without self-rarity.

\paragraph{PCB.}
On the PCB categories, the same property works against \method.
Most components of a board appear only once and are rare within the image although they are normal.
Without answer attribution, when the map is self-rarity alone, the maximum lies on the defect in 6.2\% of the defect images.
With the full map, the share is 37.8\%, against 51.8\% without self-rarity and 52.0\% for the auxiliary-trained methods.
The image AUROC is 71.5, against 74.4 when either part of the map is removed (\cref{tab:supp_peak}).
Both parts thus respond to normal components here, as answer attribution also marks components that are present on every board.
Row 6 of \cref{fig:supp_peak_examples} shows such a case, where the maximum of \method lies on a normal component and the baselines place it on the defect.

\FloatBarrier
\subsection{Failure analysis}
\label{sec:supp_failure}
\begin{figure}[!htbp]
\centering
\includegraphics[width=\linewidth]{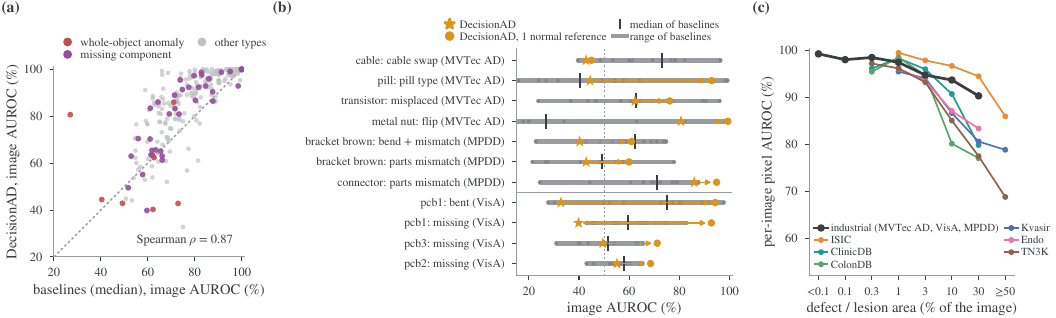}
\caption{\textbf{Failure analysis.}
(a) Image AUROC per defect type (MVTec AD, VisA, MPDD, Real-IAD, at least five images per type) of \method against the median of the 12 reproduced baselines.
(b) Whole-object anomalies and PCB components, with every baseline (gray dots and range, median as a bar), \method (star), and \method with one normal reference image (circle, mean of three seeds).
(c) Per-image pixel AUROC of the map of \method (defect pixels against the rest of the same image) by the area of the defect or lesion. Bins with at least eight images.}
\label{fig:failure}
\end{figure}

\begin{figure}[!htbp]
\centering
\includegraphics[width=\linewidth]{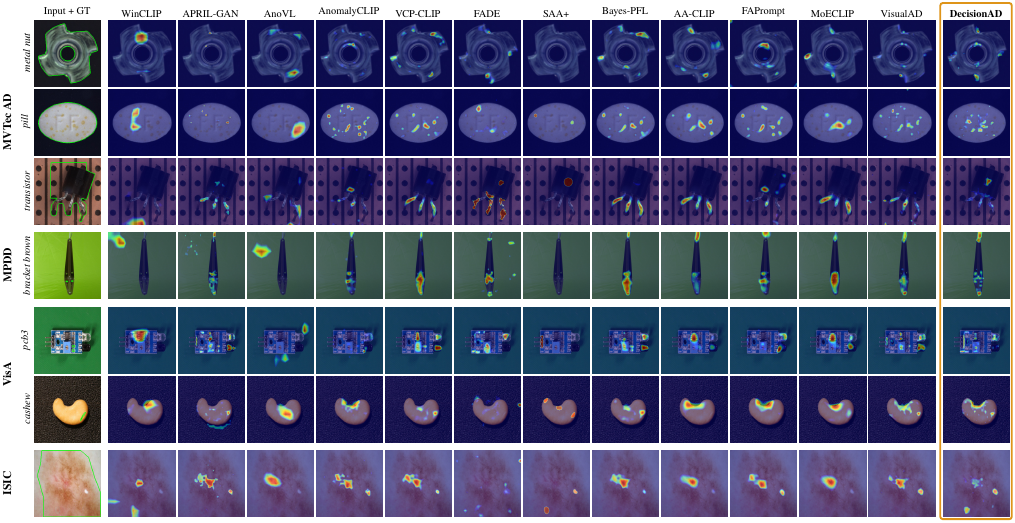}
\caption{\textbf{Failure cases by kind.} Rows 1--4 show whole-object anomalies (flipped nut, pill of another type, misplaced transistor, mismatched part), row 5 a missing component, row 6 a very small defect (0.2\% of the image, about three patches), and row 7 a lesion covering most of the image. Display as in \cref{fig:qual}.}
\label{fig:supp_failures}
\end{figure}

To find where \method fails and whether the failures are specific to it, we compute the image AUROC of every defect type for \method and the 12 reproduced baselines (\cref{fig:failure}a).
We include the 270 types with at least five images on MVTec AD, VisA, MPDD, and Real-IAD, and the negatives of a type are the normal images of the same category.
The difficult types are largely shared.
Across types, the AUROC of \method follows the median of the baselines (Spearman $\rho=0.87$).
\method remains below 70 on 41 types, and the baseline median also remains below 70 on 36 of them.
We describe four kinds of failure, three on industrial images and one that appears mostly on medical images.
\Cref{fig:supp_failures} shows examples of each kind with every baseline.

\paragraph{Whole-object anomalies.}
A flipped metal nut, a pill of another type, a misplaced transistor, swapped cables, and mismatched MPDD parts leave no damaged region on the object.
Every patch is an intact surface, and the anomaly exists only relative to the intended appearance of the product.
Self-rarity then finds no patch rarer than the rest of the object, and the answer score detects no damage.
On pill type, the map marks the object with a per-image pixel AP of 99.1.
It does so no more strongly than on normal pills, and the image AUROC is therefore 44.4.
On these seven types, \method averages 57.0 and the baseline median 55.1 (\cref{fig:failure}b).
No baseline exceeds 70 on all seven types. AnomalyCLIP exceeds it on six, AnoVL on five, and five baselines on none.
What is missing is the normal appearance of the product.
With a single normal reference image (\cref{sec:supp_fewshot}), the mean rises to 75.6.
Pill type rises from 44.4 to 92.9 and flip from 80.6 to 99.3.
Swapped cables remain difficult (45.0), since each cable matches a normal cable and only their arrangement is wrong.

\paragraph{Components on populated boards.}
On missing components of VisA PCB1--PCB3, \method reaches 39.8--55.1 and the baseline median 51.7--59.6.
On bent pins of PCB1, \method reaches 32.8 and the baseline median 75.0.
As a board carries many distinct components, normal components are also rare within the image.
A missing component leaves a plain board area, which is not rare.
On these missing-component images, the maximum of the map never lies on the defect.
One normal reference raises PCB1 to 92.8 on missing components and 94.3 on bent pins, whereas missing components on PCB2 and PCB3 remain below 72.

\paragraph{Very small defects.}
One patch covers 0.07\% of the 592\,px input.
For defects below 0.1\% of the image (about one patch), the pixel AP of \method is 36.1 and the maximum of the map lies on the defect in 36\% of the images, against 63.9 and 63\% for defects of 1--3\%.
These shares count the mask itself, without the tolerance of \cref{sec:supp_peak}.
The image AUROC on defects below 0.5\% is still 89.3, 6.4\%p above the auxiliary-trained methods and 15.1\%p above the training-free methods (\cref{tab:supp_breakdown}).
Four of the five weakest Real-IAD types, scratches and pits on regulators and scratches on switches and USB connectors (44.1--53.9), are also difficult for the baselines. The best of them remains below 72.

\paragraph{Anomalous regions covering much of the image.}
Self-rarity compares each patch with the rest of the image, so a region that covers most of the image leaves few normal patches to compare with.
For defects of 5\% and more on MVTec AD and VisA, self-rarity still adds 2.9\%p of image AUROC (\cref{tab:supp_breakdown}).
The per-image pixel AUROC decreases overall with the defect or lesion area on every dataset (\cref{fig:failure}c).
It falls on industrial images from 99.3 below 0.1\% to 90.3 at 30--50\%, on ISIC from 99.5 at 1--3\% to 86.0 above 50\%, and on TN3K from 97.2 at 0.3--1\% to 68.8 above 50\%.
ISIC lesions cover 28\% of the image on average, and 62 of its 379 images more than half.

\FloatBarrier
\section{Medical images}
\label{sec:supp_medical}
The main paper evaluates industrial inspection.
This section applies \method to nine medical datasets with all settings unchanged except the prompt pair.

\paragraph{Setup.}
Following prior work~\cite{zhou2024anomalyclip,cao2024adaclip}, the medical datasets are HeadCT, BrainMRI, and Br35H for image-level metrics and ISIC, ClinicDB, ColonDB, Kvasir, Endo, and TN3K for pixel-level metrics (\cref{tab:datasets}).
The prompt pair is built as in \cref{sec:answer} with medical wording that names no modality, organ, or disease (\cref{tab:impl}).
The prompt pair is not read from the question.
Like the question, it is chosen once per domain and is the same for all datasets and categories of that domain.

\paragraph{Comparison with ZSAD methods.}
With only the prompt pair changed, \method ranks first in AUPRO on ClinicDB (81.8), ColonDB (80.5), and TN3K (54.7), first in image AP on HeadCT (95.8), and second in image AUROC on HeadCT (\cref{tab:medical}).
On BrainMRI and Br35H, seven and eight of the eight auxiliary-trained methods have a higher image AUROC. Among the training-free methods, \method is first on Br35H and 0.1\%p below AnoVL on BrainMRI.
It is weaker in pixel metrics on ISIC.
ISIC lesions cover 28\% of the image on average, and self-rarity has few normal patches to compare with (\cref{sec:supp_failure}).
\Cref{tab:supp_medical} reports the remaining metrics, \cref{tab:fewshot_med} the results with normal references, and \cref{sec:supp_box} the box answers of Jev 1.14 on the pixel-level datasets.
\begin{table*}[!htbp]
\centering
\caption{\textbf{Comparison on medical datasets} (category macro, \%).
\method uses the medical prompt pair.
Notation as in \cref{tab:industrial}.}
\label{tab:medical}
\fitbox{\scriptsize
\setlength{\tabcolsep}{1.8pt}
\begin{tabular}{l l c cc cc cc cc cc cc cc cc cc}
\toprule
\multicolumn{1}{l}{\multirow{3}{*}[-1\dimexpr\aboverulesep+\belowrulesep\relax]{Method}} & \multirow{3}{*}[-1\dimexpr\aboverulesep+\belowrulesep\relax]{Venue} & \multirow{3}{*}[-1\dimexpr\aboverulesep+\belowrulesep\relax]{Train} & \multicolumn{6}{c}{Image-level} & \multicolumn{12}{c}{Pixel-level} \\
\cmidrule(lr){4-9} \cmidrule(lr){10-21}
 &  &  & \multicolumn{2}{c}{HeadCT} & \multicolumn{2}{c}{BrainMRI} & \multicolumn{2}{c}{Br35H} & \multicolumn{2}{c}{ISIC} & \multicolumn{2}{c}{ClinicDB} & \multicolumn{2}{c}{ColonDB} & \multicolumn{2}{c}{Kvasir} & \multicolumn{2}{c}{Endo} & \multicolumn{2}{c}{TN3K} \\
\cmidrule(lr){4-5} \cmidrule(lr){6-7} \cmidrule(lr){8-9} \cmidrule(lr){10-11} \cmidrule(lr){12-13} \cmidrule(lr){14-15} \cmidrule(lr){16-17} \cmidrule(lr){18-19} \cmidrule(lr){20-21}
 &  &  & AUC & AP & AUC & AP & AUC & AP & AUC & PRO & AUC & PRO & AUC & PRO & AUC & PRO & AUC & PRO & AUC & PRO \\
\midrule
WinCLIP~\cite{jeong2023winclip} & CVPR'23 & -- & 84.1 & 84.1 & 89.9 & 93.5 & 81.6 & 84.0 & 70.7 & 40.5 & 74.3 & 35.0 & 61.2 & 23.5 & 69.6 & 26.6 & 68.5 & 26.9 & 65.7 & 30.2 \\
AnoVL~\cite{deng2023anovl} & arXiv'23 & -- & 81.1 & 83.4 & 90.6 & 93.1 & 87.8 & 88.3 & \textbf{96.9} & \textbf{91.5} & 80.0 & 50.2 & 73.9 & 41.4 & 73.1 & 27.4 & 81.1 & 45.8 & 61.6 & 30.2 \\
FADE~\cite{li2024fade} & BMVC'24 & -- & 84.3 & 83.5 & 88.2 & 92.7 & 78.9 & 81.8 & 83.7 & 49.8 & 71.8 & 39.8 & 72.7 & 43.0 & 60.8 & 20.1 & 68.0 & 26.6 & 68.3 & 34.5 \\
SAA+~\cite{cao2023saa} & TCYB'25 & -- & 48.8 & 46.9 & 49.5 & 58.5 & 59.8 & 52.8 & 77.7 & 59.5 & 64.2 & 23.0 & 72.7 & 49.0 & 84.8 & 25.2 & 82.6 & 65.0 & 66.8 & 23.4 \\
\midrule
APRIL-GAN~\cite{chen2023aprilgan} & arXiv'23 & AD & 86.8 & 87.8 & 92.7 & \underline{94.0} & 93.2 & 93.9 & 90.8 & 79.2 & 80.8 & 61.6 & 78.2 & 65.6 & 74.3 & 37.3 & 82.2 & 55.0 & 73.0 & 38.3 \\
AnomalyCLIP~\cite{zhou2024anomalyclip} & ICLR'24 & AD & 93.1 & 91.1 & 90.0 & 92.0 & 94.3 & \underline{94.3} & 91.5 & 79.5 & 83.0 & 68.3 & 81.9 & 71.9 & 78.3 & 45.9 & 84.4 & 62.5 & \underline{79.9} & 51.2 \\
VCP-CLIP~\cite{qu2024vcpclip} & ECCV'24 & AD & 90.6 & 91.4 & 92.4 & 92.8 & \underline{95.1} & 93.5 & 93.0 & 84.2 & 86.2 & 72.0 & 81.0 & 70.7 & 81.0 & 47.5 & \underline{86.0} & 67.3 & 77.6 & 48.5 \\
Bayes-PFL~\cite{qu2025bayespfl} & CVPR'25 & AD & 87.5 & 84.4 & 91.7 & 92.1 & 93.8 & 92.6 & 90.6 & 83.3 & 84.8 & 67.4 & 81.8 & 70.2 & 80.7 & 50.3 & 84.8 & 63.0 & 77.3 & 49.5 \\
AA-CLIP~\cite{ma2025aaclip} & CVPR'25 & AD & \textbf{97.0} & \underline{95.1} & 92.6 & 91.2 & 94.8 & 92.3 & 95.0 & 87.9 & \textbf{90.3} & \underline{79.2} & 83.7 & \underline{76.1} & \underline{87.0} & \textbf{59.7} & \textbf{90.7} & \textbf{75.2} & 78.2 & 47.1 \\
FAPrompt~\cite{zhu2025faprompt} & ICCV'25 & AD & 94.4 & 91.1 & \textbf{95.1} & \textbf{94.7} & \textbf{97.4} & \textbf{96.8} & 90.7 & 78.5 & 84.9 & 71.0 & \underline{84.5} & 75.0 & 79.2 & 47.2 & 85.4 & 66.1 & \textbf{82.4} & \underline{53.0} \\
MoECLIP~\cite{park2026moeclip} & CVPR'26 & AD & 95.2 & 94.2 & \underline{93.4} & \textbf{94.7} & 93.5 & 93.5 & \underline{95.7} & \underline{89.9} & \underline{89.8} & 77.4 & \textbf{84.9} & \underline{76.1} & \textbf{88.0} & \underline{57.9} & \textbf{90.7} & \textbf{75.2} & 76.4 & 43.9 \\
VisualAD~\cite{hou2026visualad} & CVPR'26 & AD & 89.4 & 84.9 & 92.6 & 91.0 & 94.4 & 91.6 & 86.7 & 72.2 & 84.8 & 71.5 & 76.8 & 72.9 & 79.8 & 45.4 & 84.7 & 66.0 & 65.4 & 41.2 \\
\midrule
\textbf{DecisionAD} & -- & -- & \underline{96.3} & \textbf{95.8} & 90.5 & 91.0 & 92.7 & 89.8 & 90.0 & 76.2 & 89.6 & \textbf{81.8} & 83.3 & \textbf{80.5} & 83.9 & 54.7 & 85.8 & \underline{68.0} & 76.9 & \textbf{54.7} \\
\bottomrule
\end{tabular}
}
\end{table*}

\begin{table}[!htbp]
\centering
\caption{\textbf{Remaining metrics on medical datasets} (category macro, image F1-max, pixel AP and F1-max, \%).}
\label{tab:supp_medical}
\fitbox{\scriptsize
\setlength{\tabcolsep}{1.8pt}
\begin{tabular}{l l c c c c cc cc cc cc cc cc}
\toprule
\multicolumn{1}{l}{\multirow{3}{*}[-1\dimexpr\aboverulesep+\belowrulesep\relax]{Method}} & \multirow{3}{*}[-1\dimexpr\aboverulesep+\belowrulesep\relax]{Venue} & \multirow{3}{*}[-1\dimexpr\aboverulesep+\belowrulesep\relax]{Train} & \multicolumn{3}{c}{Image-level} & \multicolumn{12}{c}{Pixel-level} \\
\cmidrule(lr){4-6} \cmidrule(lr){7-18}
 &  &  & \multicolumn{1}{c}{HeadCT} & \multicolumn{1}{c}{BrainMRI} & \multicolumn{1}{c}{Br35H} & \multicolumn{2}{c}{ISIC} & \multicolumn{2}{c}{ClinicDB} & \multicolumn{2}{c}{ColonDB} & \multicolumn{2}{c}{Kvasir} & \multicolumn{2}{c}{Endo} & \multicolumn{2}{c}{TN3K} \\
\cmidrule(lr){4-4} \cmidrule(lr){5-5} \cmidrule(lr){6-6} \cmidrule(lr){7-8} \cmidrule(lr){9-10} \cmidrule(lr){11-12} \cmidrule(lr){13-14} \cmidrule(lr){15-16} \cmidrule(lr){17-18}
 &  &  & F1 & F1 & F1 & AP & F1 & AP & F1 & AP & F1 & AP & F1 & AP & F1 & AP & F1 \\
\midrule
WinCLIP~\cite{jeong2023winclip} & CVPR'23 & -- & 79.8 & 86.9 & 74.4 & 32.6 & 40.0 & 23.5 & 30.7 & 13.3 & 19.6 & 28.3 & 36.4 & 25.1 & 33.9 & 20.1 & 29.0 \\
AnoVL~\cite{deng2023anovl} & arXiv'23 & -- & 76.4 & 88.4 & 80.3 & \textbf{89.2} & \textbf{81.2} & 29.0 & 36.4 & 15.6 & 23.8 & 32.5 & 38.8 & 40.3 & 45.8 & 19.6 & 26.1 \\
FADE~\cite{li2024fade} & BMVC'24 & -- & 78.2 & 85.6 & 72.1 & 40.3 & 55.5 & 16.7 & 26.6 & 16.0 & 25.0 & 18.6 & 32.5 & 20.9 & 35.0 & 23.1 & 31.1 \\
SAA+~\cite{cao2023saa} & TCYB'25 & -- & 66.9 & 77.8 & 69.0 & 49.4 & 61.3 & 12.5 & 23.0 & 27.5 & 40.0 & \textbf{62.9} & \textbf{64.3} & 58.8 & 59.4 & 20.9 & 34.0 \\
\midrule
APRIL-GAN~\cite{chen2023aprilgan} & arXiv'23 & AD & 81.2 & 91.0 & 85.6 & 73.2 & 66.1 & 29.1 & 39.3 & 21.4 & 29.4 & 32.2 & 39.9 & 39.9 & 46.0 & 32.4 & 35.2 \\
AnomalyCLIP~\cite{zhou2024anomalyclip} & ICLR'24 & AD & 88.4 & 86.9 & 86.9 & 72.6 & 67.3 & 38.2 & 42.5 & 31.7 & 37.5 & 39.5 & 45.7 & 47.2 & 51.3 & \underline{44.4} & \underline{46.5} \\
VCP-CLIP~\cite{qu2024vcpclip} & ECCV'24 & AD & 83.1 & \underline{92.2} & \underline{91.2} & 76.0 & 71.0 & 38.1 & 45.8 & 24.6 & 34.0 & 41.6 & 47.8 & 48.3 & 52.2 & 39.0 & 41.9 \\
Bayes-PFL~\cite{qu2025bayespfl} & CVPR'25 & AD & 82.3 & 89.7 & 86.8 & 77.8 & 70.6 & 33.2 & 42.3 & 24.1 & 33.8 & 40.1 & 47.8 & 44.0 & 50.3 & 38.9 & 42.2 \\
AA-CLIP~\cite{ma2025aaclip} & CVPR'25 & AD & \textbf{93.8} & 91.6 & 89.9 & 81.7 & 74.5 & \textbf{55.1} & \underline{55.0} & 32.9 & 39.1 & 56.4 & 57.1 & \textbf{61.0} & \textbf{61.1} & 37.0 & 40.3 \\
FAPrompt~\cite{zhu2025faprompt} & ICCV'25 & AD & 90.5 & \textbf{94.1} & \textbf{93.5} & 71.9 & 66.2 & 40.9 & 44.9 & 35.0 & 40.2 & 41.6 & 46.8 & 49.9 & 52.8 & \textbf{47.1} & \textbf{48.2} \\
MoECLIP~\cite{park2026moeclip} & CVPR'26 & AD & 90.7 & 91.0 & 87.2 & \underline{85.8} & \underline{77.8} & 51.2 & 52.7 & \underline{35.4} & \textbf{41.9} & \underline{58.9} & \underline{58.8} & \underline{60.9} & \underline{61.0} & 34.4 & 38.7 \\
VisualAD~\cite{hou2026visualad} & CVPR'26 & AD & 84.5 & 90.3 & 87.9 & 64.4 & 62.5 & 37.7 & 45.6 & 27.1 & 35.6 & 41.8 & 47.7 & 46.0 & 52.4 & 29.1 & 31.3 \\
\midrule
\textbf{DecisionAD} & -- & -- & \underline{91.1} & 90.0 & 87.4 & 65.3 & 64.6 & \underline{53.5} & \textbf{56.5} & \textbf{36.8} & \underline{40.5} & 42.2 & 53.4 & 41.6 & 53.9 & 37.8 & 43.3 \\
\bottomrule
\end{tabular}
}
\end{table}
\paragraph{Comparison with decision models.}
The decision models are asked the medical questions of \cref{tab:dm_questions}.
Their mean image AUROC on HeadCT, BrainMRI, and Br35H is 95.7--98.5, above that of all reproduced ZSAD methods (at most 95.6) and of \method (93.1, \cref{tab:supp_decision}).
Their probabilities are also more consistent than on industrial images, with a polarity inconsistency of 0.02--0.17 on average against 0.04--0.38 (\cref{tab:supp_gap}).

\paragraph{Role of the answer score.}
\Cref{tab:supp_ablation_medical} repeats \cref{tab:ablation} with BrainMRI.
Without the answer score in the map and in $u$, the image AUROC on BrainMRI is at chance, 40.4\%p below the full method (90.5).
The decision on these images thus rests on the answer score, which reflects the whole image, rather than on a local region of the map.
\begin{table}[!htbp]
\centering
\caption{\textbf{Component ablation with BrainMRI.} Notation as in \cref{tab:ablation}. BrainMRI has image-level labels only and no multi-scale pooling variant (--).}
\label{tab:supp_ablation_medical}
\fitbox{\scriptsize
\setlength{\tabcolsep}{2.2pt}
\begin{tabular}{l ccccccc}
\toprule
\multicolumn{1}{l}{\multirow{3}{*}[-1\dimexpr\aboverulesep+\belowrulesep\relax]{Setting}} & \multicolumn{3}{c}{MVTec AD} & \multicolumn{3}{c}{VisA} & BrainMRI \\
\cmidrule(lr){2-4} \cmidrule(lr){5-7} \cmidrule(lr){8-8}
 & Image & \multicolumn{2}{c}{Pixel} & Image & \multicolumn{2}{c}{Pixel} & Image \\
\cmidrule(lr){2-2} \cmidrule(lr){3-4} \cmidrule(lr){5-5} \cmidrule(lr){6-7} \cmidrule(lr){8-8}
 & AUC & AP & PRO & AUC & AP & PRO & AUC \\
\midrule
\rowcolor{oursrow}\textbf{DecisionAD} & 96.0 & 45.8 & 90.4 & 87.5 & 23.9 & 90.5 & 90.5 \\
\midrule
\multicolumn{8}{l}{\textit{(a) Answer score removed from}} \\
\quad map and $u$ (image only) & \cellcolor{red!40}$-$8.7 & \cellcolor{red!31}$-$4.4 & \cellcolor{red!29}$-$4.0 & \cellcolor{red!40}$-$7.1 & \cellcolor{red!21}$-$2.7 & \cellcolor{red!21}$-$2.6 & \cellcolor{red!40}$-$40.4 \\
\quad map ($Z{=}z_R$) & \cellcolor{red!14}$-$1.5 & \cellcolor{red!31}$-$4.4 & \cellcolor{red!29}$-$4.0 & \cellcolor{blue!12}$+$1.2 & \cellcolor{red!21}$-$2.7 & \cellcolor{red!21}$-$2.6 & \cellcolor{blue!14}$+$1.5 \\
\quad $u$ ($u{=}z_L$) & \cellcolor{red!16}$-$1.9 & $\cdot$ & $\cdot$ & \cellcolor{red!33}$-$4.6 & $\cdot$ & $\cdot$ & \cellcolor{red!40}$-$42.1 \\
\midrule
\multicolumn{8}{l}{\textit{(b) Image side removed}} \\
\quad map from $u$ ($u{=}z_s$) & $-$0.3 & $\cdot$ & $\cdot$ & \cellcolor{red!14}$-$1.5 & $\cdot$ & $\cdot$ & \cellcolor{blue!27}$+$3.6 \\
\quad self-rarity ($Z{=}z_A$) & \cellcolor{red!9}$-$0.6 & \cellcolor{red!40}$-$7.1 & \cellcolor{red!20}$-$2.5 & \cellcolor{red!9}$-$0.6 & \cellcolor{red!27}$-$3.7 & \cellcolor{red!30}$-$4.1 & \cellcolor{red!9}$-$0.6 \\
\quad refinement & $-$0.1 & \cellcolor{red!18}$-$2.2 & \cellcolor{red!8}$-$0.5 & \cellcolor{red!9}$-$0.7 & \cellcolor{red!14}$-$1.5 & 0.0 & \cellcolor{red!9}$-$0.6 \\
\quad multi-scale pooling & \cellcolor{red!17}$-$2.0 & $\cdot$ & $\cdot$ & \cellcolor{red!20}$-$2.5 & $\cdot$ & $\cdot$ & -- \\
\bottomrule
\end{tabular}
}
\end{table}

\paragraph{Why the decision models are stronger here.}
The training data of the decision models are undisclosed. Three observations narrow down the cause.
\begin{enumerate}[label=(\roman*),leftmargin=*]
\item \textbf{The task is whole-image recognition.} Each of the three datasets asks for one named finding, a hemorrhage or a brain tumor, that is visible at the scale of the image. It does not ask for a small deviation from the normal appearance of a product. \method shows the same pattern. On BrainMRI, the answer score alone is 3.6\%p above the full method in image AUROC, and the map alone is at chance, whereas on MVTec AD and VisA the map raises the image AUROC (\cref{tab:supp_ablation_medical}). Whole-image recognition is what vision-language models are mostly trained and evaluated on, whereas MMAD~\cite{jiang2025mmad} finds them short of industrial requirements.
\item \textbf{Medical findings are common in public image--text data.} The zero-shot accuracy of multimodal models on a concept grows with the frequency of that concept in their pretraining data~\cite{udandarao2024nozeroshot}. BiomedCLIP~\cite{zhang2023biomedclip} alone collects 15 million figure--caption pairs from biomedical articles. Captioned images of the defects of a specific product are rare by comparison.
\item \textbf{The datasets are public with their labels.} HeadCT, BrainMRI, and Br35H have been public on Kaggle with their labels for several years. Benchmark contamination is documented for closed-source language models~\cite{balloccu2024leak} and for multimodal models, where it is stronger for proprietary models and older benchmarks~\cite{chen2024mmstar,song2025leaked}. However, as the industrial benchmarks are public as well, availability alone does not explain why the decision models are stronger on medical than on industrial images.
\end{enumerate}
We therefore read these results as evidence that the three datasets reward whole-image recognition. The probabilities of the decision models remain inconsistent on these datasets, although less than on industrial images, and only Jev 1.14 returns a location, as one box (\cref{sec:supp_box}).

\end{document}